\documentclass[journal]{IEEEtran}

\usepackage{cite}

\ifCLASSINFOpdf
  \usepackage[pdftex]{graphicx}
\graphicspath{{./images/}{./figures}}
\DeclareGraphicsExtensions{.pdf,.jpeg,.png}
\else
\usepackage[dvips]{graphicx}
\graphicspath{{./figures/}}
\DeclareGraphicsExtensions{.eps}
\fi

\usepackage{amsmath}
\usepackage{amssymb}
\usepackage{amsthm}
\usepackage{physics}
\usepackage{mathtools}
\usepackage{bm}
\usepackage{enumitem}
\DeclareMathOperator*{\argmin}{arg\,min}
\DeclareMathOperator*{\argmax}{arg\,max}
\DeclareMathOperator{\sign}{sign}

\DeclarePairedDelimiterX\setc[2]{\{}{\}}{#1 \,\delimsize\vert\, #2}
\DeclarePairedDelimiterX\sets[2]{\{}{\}}{\, #1 : #2 \,}

\makeatletter
\newcommand{\@giventhatstar}[2]{\left(#1\!\:\middle|\!\:#2\right)}
\newcommand{\@giventhatnostar}[3][]{#1(#2\!\:#1|\!\:#3#1)}
\newcommand{\giventhat}{\@ifstar\@giventhatstar\@giventhatnostar}
\makeatother

\usepackage{algorithm}
\usepackage{algorithmic}

\usepackage{array}

\ifCLASSOPTIONcompsoc
 \usepackage[caption=false,font=normalsize,labelfont=sf,textfont=sf]{subfig}
\else
 \usepackage[caption=false,font=footnotesize]{subfig}
\fi

\usepackage{stfloats}
\usepackage{url}
\usepackage[numbers]{natbib}

\usepackage{footnote}

\usepackage{amsmath}
\usepackage[amssymb]{SIunits}
\usepackage{amssymb}
\usepackage{amsthm}
\usepackage{wasysym}
\usepackage{physics}
\usepackage{mathtools}
\usepackage{bm}

\usepackage{xfrac}
\usepackage{nicefrac}

\newcommand*\textfrac[2]{
  \text{#1}/\text{#2}
}
\newcommand{\centered}[1]{\begin{tabular}{l} #1 \end{tabular}}

\makeatletter
\newcommand{\removelatexerror}{\let\@latex@error\@gobble}
\makeatother

\usepackage{array}
\newcolumntype{C}[1]{>{\centering\arraybackslash}p{#1}}
\newcolumntype{L}[1]{>{\raggedright\arraybackslash}p{#1}}
\newcolumntype{R}[1]{>{\raggedleft\arraybackslash}p{#1}}
\usepackage{color}
\usepackage{colortbl}
\usepackage{xcolor}
\usepackage{multirow}

\usepackage{diagbox}

\usepackage{comment}

\usepackage{hyperref}
\usepackage{cleveref}
\crefname{alg}{alg.}{algs.}
\Crefname{alg}{Algorithm}{Algorithms}
\crefname{ALC@unique}{line}{lines}
\Crefname{ALC@unique}{Line}{Lines}
\Crefname{appsec}{Appendix}{Appendices}
\crefformat{footnote}{#2\footnotemark[#1]#3}

\begin{document}
\bstctlcite{IEEEexample:BSTcontrol}

\title{PDPGD: Primal-Dual Proximal Gradient Descent Adversarial Attack}

\author{Alexander~Matyasko,~\IEEEmembership{Member,~IEEE}
  and~Lap-Pui~Chau,~\IEEEmembership{Fellow,~IEEE,}\thanks{The authors are with the Department of Electrical and Electronic
    Engineering, Nanyang Technological University, Singapore, 639798 SG (e-mail:
    aliaksan001@e.ntu.edu.sg; elpchau@ntu.edu.sg).}}

\maketitle

\begin{abstract}
  State-of-the-art deep neural networks are sensitive to small input perturbations. Since the discovery of this intriguing vulnerability, many defence methods have been proposed that attempt to improve robustness to adversarial noise. Fast and accurate attacks are required to compare various defence methods. However, evaluating adversarial robustness has proven to be extremely challenging. Existing norm minimisation adversarial attacks require thousands of iterations (e.g. Carlini \& Wagner attack), are limited to the specific norms (e.g. Fast Adaptive Boundary), or produce sub-optimal results (e.g. Brendel \& Bethge attack). On the other hand, PGD attack, which is fast, general and accurate, ignores the norm minimisation penalty and solves a simpler perturbation-constrained problem. In this work, we introduce a fast, general and accurate adversarial attack that optimises the original non-convex constrained minimisation problem. We interpret optimising the Lagrangian of the adversarial attack optimisation problem as a two-player game: the first player minimises the Lagrangian wrt the adversarial noise; the second player maximises the Lagrangian wrt the regularisation penalty. Our attack algorithm simultaneously optimises primal and dual variables to find the minimal adversarial perturbation. In addition, for non-smooth $l_p$-norm minimisation, such as $l_{\infty}$-, $l_1$-, and $l_0$-norms, we introduce primal-dual proximal gradient descent attack. We show in the experiments that our attack outperforms current state-of-the-art $l_{\infty}$-, $l_2$-, $l_1$-, and $l_0$-attacks on {MNIST}, {CIFAR-10} and {Restricted ImageNet} datasets against unregularised and adversarially trained models.
\end{abstract}


\section{Introduction}\label{sec:norm-pgd-introduction}

\IEEEPARstart{D}{eep} neural networks~(DNNs) have been remarkably successful for a wide range of perceptual problems: image classification~\cite{he2015deep}, object detection~\cite{shaoqing2015fasterrcnn}, speech recognition~\cite{hinton2012deep}, machine translation~\cite{sutskever2014sequence}. Despite their excellent performance in an extensive range of practical applications, DNNs are sensitive to small, imperceptible perturbations in the input data. This intriguing vulnerability was discovered by~\citet{szegedy2014intriguing}. It was found that it is possible to perturb any given image in such a way that deep neural network misclassifies it with high confidence, but the image remains visually indistinguishable from the original image to a human observer. The lack of robustness is not specific to convolutional neural networks~(CNNs) for image recognition problems. Subsequently, it was found that recurrent neural networks~(RNNs) are susceptible to perturbations in text for language understanding task~\cite{liang2018deeptext} and audio for speech recognition task~\cite{carlini2018audio}.

The lack of robustness to small, imperceptible perturbations is counter-intuitive. Unlike state-of-the-art deep neural networks models, human vision is remarkably robust to variations in the input, e.g.~changes in the lighting condition or changes in the object shape or pose. For example, if someone wants to conceal his or her identity from the police, the person will have to wear a mask or undergo cosmetic surgery. In comparison, an adversary needs to change only a few pixels in the image to fool a state-of-the-art facial recognition system~\cite{su2019onepixel}. Additionally, the existence of adversarial examples poses a serious concern for the application of deep neural networks in safety and security-critical applications~\cite{papernot2016limitations}. For example, recent studies showed that the attacks could be applied in the physical world~\cite{kurakin2016adversarialexamples,eykholt2018robust}.

The problem of adversarial examples has spurred significant interest in the research of deep neural networks. The field of research in the area of robust deep learning can be broadly divided into the research on attacks~\cite{dezfooli2015deepfool,carlini2016towards,croce2019minimally} and the research on the defences~\cite{goodfellow2014explaining,madry2017towards,matyasko2017margin}. Like in an arms race, these two sides compete with each other: new defences introduced to protect against existing attacks; new attacks introduced to counteract existing defences. Since this vulnerability was discovered, dozens of techniques have been proposed to improve robustness to adversarial noise. To no avail, the problem of training a robust deep neural network remains unsolved. It's now even speculated that adversarial examples could fool time-limited human observers~\cite{elsayed2018fool}.

The goal of the adversarial attack is to find a minimal $l_p$-norm perturbation that changes the model's prediction. Solving this non-convex constrained minimisation problem has proven to be challenging~\cite{carlini2019evaluating}. Existing attacks, such as C\&W~\cite{carlini2016towards} and EAD~\cite{chen2018ead}, in place of the original constrained problem, solve a sequence of unconstrained problems for multiple values of the regularisation weight $\lambda$ selected using line search or binary search. The attack's optimisation search needs to be restarted multiple times for each value $\lambda$, which increase the computational cost of the attack. Other methods, such as B\&B~\cite{brendel2019accurate} and FAB~\cite{croce2019minimally} attacks, are either limited to the specific norms or produce sub-optimal results. Fast, general and accurate attack, such as PGD~\cite{kurakin2016adversarialexamples,madry2017towards}, reformulates the original norm minimisation problem with non-convex error constraint as a surrogate loss minimisation with convex $l_{p}$-norm perturbation constraint. For this ``simpler'' problem, projected gradient descent attack (PGD) is an optimal first-order adversary~\cite{madry2017towards}. However, PGD attack does not explicitly minimise the perturbation $l_p$-norm. Instead, PGD attack minimises the model's accuracy at the threshold $\epsilon$. PGD attack needs to be restarted multiple times to evaluate the robust accuracy at multiple thresholds.

In this work, we introduce a fast and efficient adversarial attack. Our attack directly optimises the original non-convex constrained norm minimisation problem without intensive optimisation restarts. We interpret optimising the Lagrangian of the attack as playing a two-player game. The first player minimises the Lagrangian wrt the adversarial noise, while the second player maximises the Lagrangian wrt the regularisation penalty, which penalises the first player for the violation of the misclassification constraint. Then, we apply primal-dual gradient descent to simultaneously update primal and dual variables to find the minimal adversarial perturbation. In addition, we introduce primal-dual proximal gradient descent attack for non-smooth $l_p$-norm minimisation, such as $l_{\infty}$-, $l_1$- and $l_0$-norms. For RGB images, we propose a group $l_{0}^G$-norm proximal operator, which we use to minimise the number of perturbed pixels. We demonstrate in the experiments that our attack outperforms current state-of-the-art $l_{\infty}$-, $l_2$-, $l_1$- and $l_0$-attacks on {MNIST}, {CIFAR-10} and {Restricted ImageNet} datasets against unregularised and adversarially trained models. The source code to reproduce all our experiments is available at~\url{https://github.com/aam-at/cpgd}. We hope that our attack will be considered as a benchmark for a future comparison between different defence methods.

\section{Related work}\label{sec:related-work}

A plethora of adversarial attacks and defences against adversarial attacks have been proposed since the discovery of the vulnerability of DNNs to adversarial examples. In this section, we attempt to review the landscape of the research on adversarial attacks. Due to the space constraints, only selected relevant works are discussed. For a detailed overview of this diverse and active field, interested readers may refer to~\cite{yuan2019adversarial}.

Adversarial attacks can be broadly categorized based on the attacker's knowledge about the model~(white-box versus black-box attacks), the attack's specificity~(targeted versus non-targeted attacks), and the perturbation measurement~($l_{\infty}$-, $l_2$-, $l_1$-, and $l_0$-norm attacks). White-box attacks have full knowledge of the neural network model, including training data, model architecture, model weights, and model hyperparameters. Adversarial examples are generated using the model's gradients. Black-box attacks have access to the model's outputs but do not know the training data and the model architecture. This assumption is valid for attacking online ML services. Targeted attacks aim to produce a targeted misclassification, whereas the adversarial label for an untargeted attack can be arbitrary except the original one.

\subsection{White-box Adversarial Attacks}

\cite{szegedy2014intriguing} first discovered the phenomenon of adversarial examples and introduced a targeted gradient-based adversarial attack against DNNs known as {LBFGS-B} method. {LBFGS-B} method is the basis of many attack algorithms. Starting with an input $\mathbf{x}$ with a label $y$, the authors minimised the norm of the perturbation $\mathbf{r}$ subject to the constraint that the neural network misclassifies the adversarial example $\mathbf{x} + \mathbf{r}$ as some target $y_t \neq y$. A surrogate loss function is introduced in place of the original non-differentiable error constraint. Then, an unconstrained optimisation problem is solved using {LBFGS-B} optimiser for multiple values of the constraint regularisation weight $\lambda$, which is selected using a line search. However, this attack is impractical against large models because 1) it uses computationally intensive {LBFGS-B} method; 2) it requires full optimisation for each value of the regularisation weight $\lambda$.

\subsubsection{\textbf{$l_{\infty}$-norm attacks}}

\cite{goodfellow2014explaining} reformulated the attack's problem as minimising the misclassification surrogate loss subject to the $l_p$-norm perturbation constraint. Then, they noticed that after applying the first-order approximation to the surrogate loss, the normalised loss gradient wrt inputs is the adversarial direction. In particular, the adversarial perturbation for the $l_{\infty}$-norm constraint is the scaled sign of the gradient. This method, known as fast gradient method~(FGM), is inaccurate but extremely fast. \cite{kurakin2016adversarialexamples,madry2017towards} proposed an iterative version of FGM, which is known as basic iterative method~(BIM)~\cite{kurakin2016adversarialexamples} or projected gradient descent~(PGD)~\cite{madry2017towards}. PGD iteratively takes a step in the direction of FGM attack and constrains the perturbation after each update. \cite{madry2017towards} argued that PGD is an optimal first-order adversary. PGD attack is a recommended starting attack for $l_{\infty}$-norm distortions~\cite{carlini2019evaluating}. \cite{dong2018iterative} integrated the momentum into PGD iterative process. Distributionally adversarial attack~(DAA)~\cite{zheng2018distributionally} finds an adversarial data distribution that maximises the generalisation risk. The optimal data distribution is described by PDE, which they solve using particle optimisation. \cite{pooladian2020principled} introduced a proximal log-barrier attack (ProxLogBarrier), which, similar to our work, uses proximal optimisation for non-smooth norms. In comparison with our work, ProxLogBarrier attack: 1) requires an adversarial starting point, while any starting point can be used for our attack; 2) optimises the log-barrier loss, while our primal-dual attack solves for the original error constraint.

\subsubsection{\textbf{$l_{2}$-norm attacks}}

In seminal work, \citet{carlini2016towards} formally defined the problem of finding adversarial perturbation. They introduced C\&W attack that solves a sequence of unconstrained optimisation problems, similar to~\cite{szegedy2014intriguing}. They investigated how the choice of the optimiser, the surrogate loss function, and handling of the box constraints affect the attack's success. C\&W is the recommended attack for the assessment of DNNs robustness to $l_2$-norm perturbations~\cite{carlini2019evaluating}. C\&W, like {LBFGS-B} attack, requires full optimisation for each value of the regularisation weight $\lambda$, which increases the attack's computational cost. DeepFool~\cite{dezfooli2015deepfool} finds the closest class boundary and takes a step in that direction. DeepFool attack does not enforce box constraints and does not explicitly minimise the norm of the perturbation. The optimisation process stops as soon as adversarial perturbation is found. Fast Adaptive Boundary attack~(FBA)~\cite{croce2019minimally} solves the box-constrained $l_p$-norm projection on the decision hyperplane exactly and introduces a biased backward step to minimise the perturbation norm. Decoupling direction and norm $l_2$-attack~(DDN)~\cite{rony2019decoupling} proposes to adjust the radius of $l_2$-norm ball used for the $l_2$-norm projection. If the perturbation is adversarial/not adversarial, the radius of $l_2$-norm projection ball can be decreased/increased, respectively.

\subsubsection{\textbf{$l_{1}$-norm attacks}}

\cite{chen2018ead} introduced $l_1$-norm attack known as Elastic-net~(EAD). EAD attack, similar to C\&W, solves a sequence of unconstrained optimisation problems. To minimise non-smooth, subdifferentiable $l_1$-norm, they proposed to use fast iterative shrinkage-thresholding algorithm~(FISTA)~\cite{beck2009thresholding}. EAD attack is recommended $l_1$-norm attack~\cite{carlini2019evaluating}. SparseFool~(SF)~\cite{modas2019sparsefool} is a geometry-inspired $l_1$-norm attack that uses DeepFool attack~\cite{dezfooli2015deepfool} as subprocedure to estimate the local curvature of the decision boundary. They developed an efficient algorithm to compute $l_1$-projection of the perturbation on the decision boundary subject to the box constraints. Sparse $l_1$ descent attack~(SLIDE)~\cite{tramer2019adversarial} is a variant of PGD attack for $l_1$-norm. SLIDE iteratively takes a small step in the direction of the $q$-th percentile of the loss gradient and applies $l_1$-norm projection, which can be computed efficiently in $\mathcal{O}\left(n \log{n} \right)$ time~\cite{duchi2008efficient}.

\subsubsection{\textbf{$l_{0}$-norm attacks}}

$l_{\infty}$ and $l_2$ adversarial attacks often produce perturbations that change a large number of pixels in the image. Several attack methods have been proposed to minimise non-convex $l_0$-quasinorm. \cite{papernot2016limitations} proposed a targeted adversarial attack, known as Jacobian saliency map attack~(JSMA), which minimises $l_0$-norm perturbation constraint. JSMA uses the Jacobian matrix to select and modify a pair of the most salient pixels in the image. This process is repeated until the adversarial perturbation is found. \cite{su2019onepixel} found that it is possible to change the prediction of the model by modifying a single pixel. To generate adversarial examples, they applied differential evolution~(DE) algorithm on the population of vectors that change a single pixel in the image. \cite{croce2019sparse} introduced $l_0$-norm variant of PGD attack and black-box, score-based $l_0$-norm attack known as CornerSearch~(CS) attack. PGD-$l_0$ attack iteratively takes a small step in the direction of the loss gradient and applies $l_0$-norm projection. CS attack creates a probability distribution of the most salient pixels from which the adversarial perturbation is sampled.

\subsection{Black-box adversarial attacks}

A black-box adversary has limited knowledge about the model, e.g.~model prediction scores or outputs. Black-box attacks are more difficult to perform than white-box attacks because we do not know the model's gradient. \cite{papernot2016practical}~introduced a practical black-box adversarial attack based on the property that adversarial examples can transfer between models. They trained a substitute model on the model's task. Then, adversarial examples generated for the substitute model are used to attack the target model. \citet{brendel2018decisionbased} introduced a decision-based attack that estimates the decision boundary using rejection sampling. Starting at some adversarial input, they randomly draw a random perturbation from the candidate distribution and minimise the distance to the original input. \cite{guo2019simpleblackbox} sampled a random perturbation from an orthonormal basis of discrete cosine transform~(DCT), which improves query-efficiency of the decision-based attack. Gradient-based attacks should be almost always more precise than gradient-free attacks. However, gradient masking~\cite{papernot2016practical} can fool gradient-based attacks and give a false sense of security~\cite{athalye2018obfuscated}. If the defence obfuscates the gradients, gradient-free attacks often perform better than white-box attacks. \cite{carlini2019evaluating} suggested that defences should be tested on both white-box and black-box adversaries. If the model is more robust to white-box adversaries, then the model obfuscates the gradients.

\section{Adversarial Attacks on DNNs}\label{sec:advers-attacks-dnns}

In this section, we introduce a general formulation of the attack against deep neural networks in white-box settings, where the attacker has full knowledge about the model. Let $f\left(\cdot \right)$ be the mapping from the space of input pixels to the unnormalised predictive distribution on discrete label output space $f : \mathbb{R}^N \to \mathbb{R}^k$ where $k$ is the number of classes. The network prediction is the label with the highest score $\hat{k}(\mathbf{x})=\argmax f(\mathbf{x})$. For a given input image $\mathbf{x}$ with the label $y$, an adversary aims to find a minimal adversarial perturbation $\mathbf{r}$ wrt some norm $\norm{\cdot}$, such that after adding the perturbation to the original image $\mathbf{x}$ it changes the network prediction $\hat{k}(\mathbf{x} + \mathbf{r}) \neq y$. We can formulate the attack as the following optimisation problem:
\begin{equation}
  \begin{aligned} & \underset{\mathbf{r}}{\text{min}} & & \norm{\mathbf{r}} \\ &
\text{s.t.} & & \hat{k}(\mathbf{x} + \mathbf{r}) \neq y \\ & & & \mathbf{x} +
\mathbf{r} \in \mathbb{C}
  \end{aligned}\label{eq:general_perturbation}
\end{equation}
where $\mathbf{r}$ and $\mathbf{x} + \mathbf{r}$ is the \textit{adversarial} noise and the \textit{adversarial example} respectively; $\mathbb{C}$ is the input domain, e.g. $\left[0, 1 \right]^{N}$ box constraints for the normalised image. The problem above is an example of an untargeted adversarial attack. A targeted adversarial attack searches for the perturbation which changes the network prediction to the specific target: $\hat{k}(\mathbf{x} + \mathbf{r}) = y_{t}$.

The optimisation problem in~\Cref{eq:general_perturbation} is a non-convex constrained norm minimisation problem. Solving it is a challenging and non-trivial task because 1) the misclassification constraint is non-convex and non-differentiable; 2) the norm of the perturbation is non-differentiable, e.g. $l_0$- and $l_1$-, and $l_{\infty}$-norms. A plethora of attack methods have been proposed to solve the adversarial attack problem. \citet{szegedy2014intriguing} first introduced a blueprint for a targeted white-box adversarial attack, {LBFGS-B} method, against DNNs in~{\protect\NoHyper\citeyear{szegedy2014intriguing}\protect\endNoHyper}. The authors optimised the following unconstrained minimisation problem:
\begin{equation}
  \begin{aligned} & \underset{\mathbf{r}}{\text{min}} & & \norm{\mathbf{r}} +
\lambda \mathcal{L}\left(\mathbf{x} + \mathbf{r}; y\right) \\ & \text{s.t.} & &
\mathbf{x} + \mathbf{r} \in \mathbb{C}
  \end{aligned}\label{eq:szegedy_unconstrained_problem}
\end{equation}
where $\lambda$ is the regularisation penalty which penalises the violation of the misclassification constraint; $\mathcal{L}$ is a surrogate loss function, which is minimised when $\hat{k}(\mathbf{x} + \mathbf{r}) \neq y$. The solution of the problem in~\Cref{eq:szegedy_unconstrained_problem} for a fixed $\lambda$ can be found using an off-shelf optimiser, e.g. {LBFGS-B} in~\cite{szegedy2014intriguing} or ADAM in~\cite{carlini2016towards}. Finally, a line search or a binary search is performed to find the optimal regularisation weight $\lambda^{*}$, which minimises $\norm{\mathbf{r}}$. Optimisation of~\Cref{eq:szegedy_unconstrained_problem} needs to be restarted for each $\lambda$, which significantly increases the attack cost. Besides that, the optimisation search above requires gradient, so this procedure is not applicable for non-smooth or non-differentiable function minimisation, e.g. $l_{\infty}$- or $l_0$-norms.

\citet{goodfellow2014explaining} reformulated the original non-convex constrained minimisation problem as the problem of minimising the differentiable surrogate loss function subject to the convex $l_p$-norm perturbation constraint:
\begin{equation}
  \begin{aligned} & \underset{\mathbf{r}}{\text{min}} & &
\mathcal{L}\left(\mathbf{x} + \mathbf{r}; y\right) \\ & \text{s.t.} & &
\norm{\mathbf{r}} \leq \epsilon \\ & & & \mathbf{x} + \mathbf{r} \in \mathbb{C}
  \end{aligned}\label{eq:pgd_attack}
\end{equation}
They observed that after applying the first-order approximation to the surrogate loss $\mathcal{L}$, the normalised gradient of the loss is the solution of the $l_p$-norm constrained minimisation problem. In particular, the adversarial direction for the $l_{\infty}$-norm constraint is the sign of the loss gradient wrt inputs: $\mathbf{r} = \sign \nabla_{\mathbf{x}} \mathcal{L}(\mathbf{x}; y)$. This method, known as fast gradient sign method~(FGSM), is inaccurate but has dramatically increased the speed of generating adversarial noise. \cite{kurakin2016adversarialexamples,madry2017towards} proposed an iterative version of FGSM attack, known as projected gradient descent~(PGD). PGD iteratively updates the perturbation $\mathbf{r}$ by taking a small step in the direction of the adversarial target and constraints the total perturbation to~$\epsilon$ after each iteration. PGD is a simple, fast and accurate attack. This attack has also been extended to other norms, including non-differentiable $l_{1}$- and $l_0$-norms~\cite{tramer2019adversarial, croce2019sparse}. However, PGD does not explicitly minimise the $l_p$-norm of the perturbation $\mathbf{r}$. PGD attack needs to be restarted $N$ times to evaluate robustness at $N$ distinct thresholds $\epsilon$, which increases the attack's computational cost.

\section{Primal-Dual Gradient Descent Attack}

Let us revisit the Lagrangian of the original non-convex constrained $l_p$-norm minimisation problem in~\Cref{eq:general_perturbation}:
\begin{equation}
  \mathbb{L}(\mathbf{r}, \lambda) = \norm{\mathbf{r}} + \lambda \mathrm{I}\left[ \hat{k}\left(\mathbf{x} + \mathbf{r}\right) \neq y \right]~\label{eq:lagrangian_general_attack}
\end{equation}
where $\lambda$ is a dual variable, which controls the weight of the misclassification constraint; $\mathrm{I}$ is an indicator function. For brevity, we omit the domain constraints $\mathbf{x} + \mathbf{r} \in \mathbb{C}$ as we can enforce it easily for typical inputs, e.g.~$\left[0, 1\right]$ box-projection $\mathrm{\Pi}_{\mathbb{C}}$ for natural images. As $\lambda \to \infty$, the solution of the problem in~\Cref{eq:lagrangian_general_attack} converges to the feasible solution of the original non-convex constrained norm minimisation problem.

Optimising the Lagrangian in~\Cref{eq:lagrangian_general_attack} can be interpreted as playing a two-player game: the $\mathbf{r}$-player seeks to minimise the Lagrangian wrt primal variable $\mathbf{r}$; the $\lambda$-player wishes to maximise the Lagrangian wrt dual variable $\lambda$. The dual variable $\lambda$ penalises the $\mathbf{r}$-player for the violation of the misclassification constraint. C\&W attack~\cite{carlini2016towards} uses a binary search to find an optimal value of the dual variable $\lambda^{*}$. For a fixed $\lambda$, each iteration of binary search requires full optimisation of the Lagrangian to find an optimal value of the primal variable $\mathbf{r}^{*}$, which increases the attack's running time.

We propose a primal-dual gradient algorithm to simultaneously optimise primal and dual variables. Unfortunately, we cannot calculate the constraint gradients to optimise the Lagrangian $\mathcal{L}(\mathbf{r}, \lambda)$ using the first-order algorithm because the misclassification constraint is non-differentiable. In line with the previous research, we express the error constraint in terms of the prediction margin. We define the prediction margin $m_y(\mathbf{x})$ for the input $\mathbf{x}$ with label $y$ as follows:
\begin{equation}
  m_y(\mathbf{x}) = f(\mathbf{x})_y - \underset{i \neq y}{\max} \ f(\mathbf{x})_i
  \label{eq:prediction_margin}
\end{equation}
The input $\mathbf{x}$ with the label $y$ is misclassified if and only if $m_y(\mathbf{x}) < 0$. Some popular smooth relaxations of $\textfrac{0}{1}$-indicator function are squared loss, hinge loss, and logistic loss. We adopt logistic loss $\log\left(1 + e^{m_y\left(\mathbf{x}\right)}\right)$ instead of hinge loss $\max\left(0, 1 + m_y\left(\mathbf{x}\right)\right)$ as in~\cite{carlini2016towards} because logistic loss is differentiable everywhere unlike hinge loss.

Using our surrogate loss $\mathcal{L}$ for the misclassification constraint, we introduce two proxy-Lagrangians for the $\mathbf{r}$-player and the $\mathbf{\lambda}$-player as follows:
\begin{equation}
  \begin{split}
    \mathbb{L}_{\mathbf{r}}(\mathbf{r}, \mathbf{\lambda}) &= \lambda_1 \norm{\mathbf{r}} + \lambda_2 \mathcal{L}\left(\mathbf{x} +\mathbf{r}, y\right) \\
    \mathbb{L}_{\lambda}(\mathbf{r}, \mathbf{\lambda}) &= \lambda_2 \mathrm{I}\left[ \hat{k}\left(\mathbf{x} + \mathbf{r}\right) \neq y \right]
  \end{split}
  \label{eq:proxy-lagrangian}
\end{equation}
where $(\lambda_{1}, \lambda_2) \in \Lambda$ and $\Lambda \subseteq \mathbb{R}_+^2$ is a 2-dimensional simplex. In the formulation above, we represent $\lambda \in \mathbb{R}_+$ as point in the 2-dimensional simplex $\lambda = \sfrac{\lambda_2}{\lambda_1}$. Note that only the $\mathbf{r}$-player uses surrogate loss in place of the misclassification constraint, while the $\lambda$-player uses original non-differentiable misclassification constraint. The $\lambda$-player chooses how much the $\mathbf{r}$-player should penalize surrogate misclassification constraint, but does so in a way as to satisfy original non-differentiable constraint.

The proxy-Lagrangians formulation in~\Cref{eq:proxy-lagrangian} avoids the issue of the non-differentiable error constraint. The $\mathbf{r}$-player wishes to find perturbation $\mathbf{r}$ which minimises $\mathbb{L}_{\mathbf{r}}(\mathbf{r}, \lambda)$, while the $\lambda$-player wants to maximise $\mathbb{L}_{\lambda}(\mathbf{r}, \lambda)$. Unfortunately, the proxy-Lagrangian formulation corresponds to a non-zero-sum game because two players optimise two different functions. Fortunately, this proxy-Lagrangian formulation admits a weaker type of equilibrium, $\Phi$-correlated equilibrium~(see~\cite[Theorem 2]{cotter2019twoplayer} for the details).

\begin{algorithm}[!b]
  \caption{Primal-Dual Gradient Descent Attack~\label{alg:primal_dual_gradient}}
  \begin{algorithmic}[1]
    \REQUIRE Image $\mathbf{x}$, label $y$, initial perturbation $\mathbf{r}^{(0)}$, the total number of iterations $T$, learning rate $\theta_{\mathbf{r}}$ and $\theta_{\lambda}$.
    \ENSURE Adversarial perturbation $\mathbf{r}$. \STATE $\mathbf{r} \gets
    \mathbf{0}$
    \FOR{$k \gets 1$ to $T$}
    \STATE Let $\nabla_{\mathbf{r}}^{(k)}$ be a gradient of $\mathbb{L}_{\mathbf{r}}(\mathbf{r}^{(k)}, \lambda^{(k)})$
    \STATE Let $\nabla_{\lambda}^{(k)}$ be a gradient of $\mathbb{L}_{\lambda}(\mathbf{r}^{(k)}, \lambda^{(k)})$
    \STATE Update $\mathbf{r}^{(k+1)} = \mathrm{\Pi}_{\mathbb{C}}^{\mathbf{x}}\left(\mathbf{r}^{(k)} - \theta_{\mathbf{r}} \nabla_{\mathbf{r}}^{(k)} \right)$\label{lst:update-primal}
    \STATE Update $\lambda^{(k+1)} = \mathrm{\Pi}_{\Lambda} \left(\lambda^{(k)} + \theta_{\lambda} \nabla_{\lambda}^{(k)} \right)$\label{lst:update-dual}
    \IF{$\hat{k}(\mathbf{x} + \mathbf{r}^{(k + 1)}) \neq y$ \AND $\norm{\mathbf{r}^{(k+1)}} \leq \norm{\mathbf{r}}$}\label{lst:save-best-1}
    \STATE $\mathbf{r}\label{lst:save-best-2}
    \gets \mathbf{r}^{(k+1)}$
    \ENDIF\label{lst:save-best-3}
    \ENDFOR
  \end{algorithmic}
\end{algorithm}

Next, we describe our primal-dual gradient descent attack~(PDGD) in~\Cref{alg:primal_dual_gradient}. The $\mathbf{r}$-player minimises the external regret, while the $\lambda$-player minimises the swap regret. We perform gradient descent on primal variables using Adam~\cite{kingma2014adam}. Adam optimiser produced the smallest perturbation in our experiments. In~\cref{lst:update-primal}, after each iteration, we project the perturbation $\mathbf{r}$ on the domain constraints $\mathbb{C}$ using operator $\mathrm{\Pi}_{\mathbb{C}}^{\mathbf{x}}$, which we define as follows: $\mathrm{\Pi}_{\mathbb{C}}^{\mathbf{x}}(\mathbf{r}) = \mathrm{\Pi}_{\mathbb{C}}(\mathbf{x} + \mathbf{r}) - \mathbf{x}$. For the $\lambda$-player, we perform gradient ascent in the log domain. Gradient updates in the log domain are equivalent to multiplicative updates, which guarantee that dual variables remain positive. In~\cref{lst:update-dual}, after each update, we project $\lambda$ onto a 2-dimensional simplex. Intuitively, if at an iteration $k$, the misclassification constraint is not satisfied, we can increase the penalty $\lambda_2$ for the $\mathbf{r}$-player. If the constraint is satisfied, we can reduce the penalty weight $\lambda_2$. Finally, we record and store the best perturbation found during the optimisation in~\crefrange{lst:save-best-1}{lst:save-best-3}.

Our primal-dual gradient descent attack has two shortcomings. First, we use gradient descent to minimise the external regret of the $\mathbf{r}$-player. Gradient descent for smooth convex functions has a convergence rate of $\mathcal{O}\left(\sfrac{1}{T}\right)$~\cite{boyd2004convex}, where $T$ is the number of gradient iterations. For the non-smooth functions, e.g. $l_1$-norm, the convergence rate of subgradient descent is $\mathcal{O}\left(\sfrac{1}{\sqrt{T}}\right)$, which is considerably slower than sublinear convergence of gradient descent for smooth functions. Secondly, our optimisation algorithm requires gradient, so it cannot be used to minimise non-differentiable functions, such as $l_0$-quasinorm. In the next section, we introduce a proximal formulation of our attack suitable for minimising any $l_p$-norm or function with a closed-form proximity operator, including non-differentiable functions.

\section{Primal-Dual Proximal Gradient Descent}

In this section, we introduce a proximal formulation of PDGD attack introduced in the previous section. Our primal-dual proximal gradient attack~(PDPGD) can be used to directly minimise any norm or function for which the proximal operator can be computed easily, including but not limited $l_{\infty}$-, $l_2$-, and $l_1$-norms, and $l_0$-quasinorm.

First, we review some basics of proximal algorithms before introducing our attack. A detailed overview of proximal algorithms can be found in~\cite{parikh2014proximal}. We define the proximal operator of the scaled function $\lambda f$ where $\lambda > 0$ as follows:
\begin{equation}
  \mathrm{prox}_{\lambda f}(x) = \underset{\mathbf{u}}{\argmin} \left( f(u) + \frac{1}{2\lambda}\norm{u - x}_2^2\right)
  \label{eq:proximity_operator}
\end{equation}

The following useful relation holds true for any proximal operator of the proper closed function $f$:
\begin{equation}
  \mathrm{prox}_{\lambda f}(x) + \lambda \mathrm{prox}_{\lambda^{-1}f^{*}}(\nicefrac{x}{\lambda}) = x 
  \label{eq:moreau_decomposition}
\end{equation}
where $f^{*}$ is the convex conjugate of $f$. The equation above is known as Moreau decomposition. Moreau decomposition is useful for deriving the proximal operators of $l_p$-norm functions. In particular, it implies that for any norm $\norm{\cdot}$:
\begin{equation}
  \mathrm{prox}_{\lambda \norm{\cdot}}(x) + \lambda \mathrm{\Pi}_{\mathcal{B}}(\nicefrac{x}{\lambda}) = x
  \label{eq:moreau_decomposition_norm}
\end{equation}
where $\mathrm{\Pi}_{\mathcal{B}}$ is a projection operator on the unit $l_p$ ball $\mathcal{B}$.

Let us revisit the $\mathbf{r}$-player proxy-Lagrangian:
\begin{equation}
  \mathbb{L}_{\mathbf{r}}(\mathbf{r}, \mathbf{\lambda}) = \lambda_1 \norm{\mathbf{r}} + \lambda_2 \mathcal{L}\left(\mathbf{x} +\mathbf{r}, y\right)
  \label{eq:r_player_proxy_lagrangian}
\end{equation}
where $\mathcal{L}$ is the surrogate loss for the non-differentiable misclassification constraint. The goal of the $\mathrm{r}$-player is to minimise the proxy-Lagrangian function. However, if the norm $\norm{\cdot}$ is non-smooth, first-order subgradient descent needs $\mathcal{O}(\sfrac{1}{\epsilon^2})$ iterations to find $\epsilon$-error local minimum. Moreover, we cannot use first-order algorithms for $l_0$-norm minimisation because the gradient of $l_0$-norm is $\emptyset$ almost everywhere. We can address the above limitations in the framework of the proximal optimisation.

First, we rewrite the proxy-Lagrangian for the $\mathbf{r}$-player as follows:
\begin{equation}
  \mathbb{L}_{\mathbf{r}}(\mathbf{r}, \lambda) = \lambda \norm{r} + \mathcal{L}(\mathbf{x} + \mathbf{r}, y)
  \label{eq:r-player-proximal-lagrangian}
\end{equation}
where $\lambda \in \mathbb{R}_{+}$ and is equal to $\nicefrac{\lambda_1}{\lambda_2}$; $\mathcal{L}$ is the surrogate loss, e.g.~logistic or hinge loss.

Consider a quadratic approximation of the $\mathbf{r}$-player proxy-Lagrangian at iteration $k$ and point $\mathbf{u}$:
\begin{equation}
  \begin{split}
    \hat{\mathbb{L}}_{\mathbf{r}}^{(k)}(\mathbf{u}, &\lambda) = \lambda \norm{\mathbf{u}} + \mathcal{L}(\hat{\mathbf{x}}^{(k)} , y) \\
    &+ \nabla \mathcal{L}(\hat{\mathbf{x}}^{(k)}, y)^T \left(\mathbf{u} - \mathbf{r}^{(k)}\right) + \frac{1}{2t}\norm{\mathbf{u} - \mathbf{r}^{(k)}}^{2}_2
  \end{split}
  \label{eq:proximal_gradient_derivation}
\end{equation}
where $\mathbf{r}^{(k)}$ and $\hat{\mathbf{x}}^{(k)} = \mathbf{x} + \mathbf{r}^{(k)}$ are the adversarial perturbation and the adversarial example at iteration $k$, respectively. Note that we ignore not necessarily differentiable $l_p$-norm penalty when applying a quadratic approximation.

We can find the perturbation at iteration $k+1$ by minimising the quadratic approximation above:
\begin{equation}
  \begin{split}
    \mathbf{r}^{(k+1)} &= \underset{\mathbf{u}}{\argmin} \, \hat{\mathbb{L}}_{\mathbf{r}}^{(k)}(\mathbf{u}, \lambda) \\
    =& \underset{\mathbf{u}}{\argmin} \, \lambda \norm{\mathbf{u}} + \frac{1}{2t} \norm{\mathbf{u} - \left(\mathbf{r}^{(k)} - t \mathcal{L}(\hat{\mathbf{x}}^{(k)}, y)\right)}_2^2 \\
  \end{split}
\end{equation}
which is by the definition of the proximal operator in~\Cref{eq:proximity_operator} equivalent to:
\begin{equation}
  \mathbf{r}^{(k+1)} = \mathrm{prox}_{\lambda \norm{},t} \left(\mathbf{r}^{(k)} - t \nabla \mathcal{L}(\hat{\mathbf{x}}^{(k)}) \right)
\end{equation}
where $t$ is a step size or a learning rate. The algorithm above is known as proximal gradient. Proximal gradient descent has a convergence rate of $\mathcal{O}\left(\sfrac{1}{T}\right)$ for non-smooth functions minimisation~\cite{parikh2014proximal}, which is faster than subgradient descent with a convergence rate of $\mathcal{O}\left(\sfrac{1}{\sqrt{T}}\right)$.

We list our primal-dual proximal gradient~(PDPGD) attack in~\Cref{alg:primal_dual_proximal_gradient}. Compare to PDGD attack in~\Cref{alg:primal_dual_gradient}, PDPGD differs: 1) in~\cref{lst:primal-gradient}, it ignores non-smooth and not necessarily differentiable norm when computing gradient wrt primal variable $\mathbf{r}$; 2) in~\cref{lst:proximal-gradient}, it uses proximal gradient update instead of gradient update when updating primal variable $\mathbf{r}$. PDPGD attack can be used to minimise any function with a closed-form proximal operator. In this paper, we limit our discussion to the minimisation of $l_p$-norm perturbations. Next, we derive and list $l_{\infty}$-, $l_2$-, $l_1$-, and $l_0$-norm proximal operators used for our proximal attack.

\begin{algorithm}[!b]
  \setcounter{ALC@unique}{0}
  \caption{Primal-Dual Proximal Gradient Descent Attack~\label{alg:primal_dual_proximal_gradient}}
  \begin{algorithmic}[1]
    \REQUIRE Image $\mathbf{x}$, label $y$, initial perturbation $\mathbf{r}^{(0)}$, the total number of iterations $T$, learning rate $\theta_{\mathbf{r}}$ and $\theta_{\lambda}$.
    \ENSURE Adversarial perturbation $\mathbf{r}$.
    \STATE $\mathbf{r} \gets \mathbf{0}$
    \FOR{$k \gets 1$ to $T$}
    \STATE Let $\nabla_{\mathbf{r}}^{(k)}$ be a gradient of $\mathcal{L}(\mathbf{x} +
    \mathbf{r}^{(k)}, y)$ where $\mathcal{L}$ is the surrogate loss, e.g. logistic loss\label{lst:primal-gradient}
    \STATE Let $\nabla_{\lambda}^{(k)}$ be a gradient of $\mathbb{L}_{\lambda}(\mathbf{r}^{(k)}, \lambda^{(k)})$
    \STATE Update $\mathbf{r}^{(k+1)} = \mathrm{\Pi}_C^{\mathbf{x}} \left( \mathrm{prox}_{\lambda
        \norm{\cdot}, \theta_{\mathbf{r}}}\left(\mathbf{r}^{(k)} - \theta_{\mathbf{r}} \nabla_{\mathbf{r}}^{(k)} \right)\right)$\label{lst:proximal-gradient}
    \STATE Update $\lambda^{(k+1)} = \mathrm{\Pi}_{\Lambda} \left(\lambda^{(k)} + \theta_{\lambda} \nabla_{\lambda}^{(k)} \right)$
    \IF{$\hat{k}(\mathbf{x} + \mathbf{r}^{(k+1)}) \neq y$ \AND
      $\norm{\mathbf{r}^{(k+1)}} \leq \norm{\mathbf{r}}$} \STATE $\mathbf{r}
    \gets \mathbf{r}^{(k+1)}$ \ENDIF \ENDFOR
  \end{algorithmic}
\end{algorithm}

\subsection{$l_\infty$-attack}

$l_{\infty}$-norm of the vector $\mathbf{x}$ returns the largest absolute element of the vector $\mathbf{x}$: $l_{\infty}(\mathbf{x}) = \max{\lvert \mathbf{\mathbf{x}}\rvert}$. Using Moreau decomposition in~\Cref{eq:moreau_decomposition_norm}, we can show that:
\begin{equation}
  \boxed{\mathrm{prox}_{\lambda \norm{\cdot}_{\infty}}(\mathbf{x}) = \mathbf{x} - \lambda \mathrm{proj}_{\{\|\cdot\|_1 \leq 1\}}(\sfrac{\mathbf{x}}{\lambda})} 
  \label{eq:linf_proximity_operator}
\end{equation}
where $\mathrm{proj}$ is a projection operator. $l_1$-norm projection can be
computed efficiently in $\mathcal{O}\left(n \log{n} \right)$
time~\cite{duchi2008efficient}.

\subsection{$l_2$-attack}

Using Moreau decomposition in~\Cref{eq:moreau_decomposition_norm}, we can derive the proximal operator of the $l_2$-norm as follows:
\begin{equation}
  \boxed{\mathrm{prox}_{\lambda \norm{\cdot}_2}(\mathbf{x}) = (1 - \sfrac{\lambda}{\norm{\mathbf{x}}_2})_+ \mathbf{x}}
  \label{eq:l2_proximity_operator}
\end{equation}
This operator is known as block soft thresholding operator.

\subsection{$l_1$-attack}

$l_1$-norm proximal operator is well-known in signal processing~\cite{beck2009thresholding}. It is defined as follows:
\begin{equation}
  \boxed{\mathrm{prox}_{\lambda \norm{\cdot}_1}(\mathbf{x}) = \mathcal{T}_{\lambda}(\mathbf{x})}
  \label{eq:l1_proximity_operator}
\end{equation}
where $\mathcal{T}_{\lambda}(\mathbf{x}) = \mathrm{sign}(\mathbf{x}) (\mathbf{x} - \lambda)_+$ is soft-thresholding.

\subsection{$l_0$-attack}\label{sec:primal_dual_l_0_prox_oper}

$l_0$-norm is non-convex quasinorm. It measures the cardinality of the vector $\mathbf{x}$. The proximal operator of $l_0$-norm minimises the total number of non-zero elements in the vector $\mathbf{x}$, and it is defined as follows:
\begin{equation}
  \boxed{\mathrm{prox}_{\lambda \norm{\cdot}_0}(x) = \mathcal{H}_{\sqrt{2\lambda}}(x)}
  \label{eq:l0_proximity_operator}
\end{equation}
where $\mathcal{H}_{\lambda}(x) = \mathrm{I}\left[x - \lambda\right]x$ is a hard-thresholding operator.

The goal of the attack for the multichannel images is to minimise the number of non-zero pixels. We define group $l_{0,G}$-norm of the vector $x$ for the groups $\mathcal{G} = \left(g_1, g_2, \ldots, g_G \right)$ as the number of groups for which at least one element of the group is non-zero:
\begin{equation}
  \norm{x}_{0}^G = \sum_{i=1}^{G} \mathrm{I}\left[ \max \abs{x}_{g_{i}} > 0 \right]
  \label{eq:group_l0_norm}
\end{equation}
For RGB images, the group partition $\mathcal{G}$ naturally corresponds to the pixels in the image. Then, we can derive the proximal operator of $l_{0,G}$-quasinorm as follows:
\begin{equation}
  \boxed{\mathrm{prox}_{\lambda \norm{\cdot}_0}(x) = \mathcal{H}_{\sqrt{2\lambda}}^{G}(x)}
  \label{eq:group_l0_proximity_operator}
\end{equation}
where $\mathcal{H}_{\lambda}^{G}(x) = \mathrm{I}\left[ \max{\abs{x_{g}}} - \lambda\right] x$ is a group hard-thresholding operator which sets all elements in the group to 0 if the maximal element in the group is less than $\lambda$.

Minimising $l_0$-norm is NP-hard problem. We also examine $l_p$-norm relaxation of the original $l_0$-norm minimisation problem where $0 < p \leq 1$. We consider $l_{\nicefrac{1}{2}}$-, $l_{\nicefrac{2}{3}}$-, and $l_1$-norm relaxation of $l_0$-norm because 1) it promotes sparsity of the solution; 2) its proximal operators can be computed in a closed-form~(see~\cref{eq:l1_proximity_operator} and~\cite{chen2016computing}). For RGB images, we apply $l_p$-norm proximal operator to the pixel with the maximal value and set other channels to 0 if the maximal pixel is 0 after applying the proximal operator.

\section{Experiments}

\textbf{Models}: We compare our attack to state-of-the-art attacks on MNIST, {CIFAR-10} and Restricted ImageNet~(R-ImageNet) datasets~\cite{tsipras2018robustness}. For each dataset, we consider a naturally trained model (\textit{plain}) and $l_{\infty}$~($l_{\infty}$-AT) and $l_2$~($l_2$-AT) adversarially trained models as in~\cite{madry2017towards}. The models for {MNIST} and {CIFAR-10} dataset are available at~\url{https://github.com/fra31/fab-attack}, while on R-ImageNet dataset we use models from~\cite{tsipras2018robustness}, which can be downloaded from~\url{https://github.com/MadryLab/robust-features-code}.

The models on MNIST achieve the following clean accuracy on the test dataset (first 1000 test images): \textit{plain} 99.17\% (98.7\%), $l_{\infty}$-AT 98.53\% (98.5\%), and $l_2$-AT 98.95\% (98.7\%). The models on {CIFAR-10} achieve the following clean test accuracy (first 1000 test images): \textit{plain} 88.38\% (89.4\%), $l_{\infty}$-AT 79.9\% (80.4\%), and $l_2$-AT 80.44\% (80.7\%). The models on R-ImageNet achieve the following clean validation accuracy (first 1000 images of the validation set): \textit{plain} 94.5\% (94.9\%), $l_{\infty}$-AT 91.62\% (91.5\%), and $l_2$-AT 91.68\% (91.9\%).

\textbf{Attacks}: We test the robustness of each model wrt to $l_{\infty}$-, $l_{2}$-, $l_1$-, and $l_0$-norm adversaries. We compare the performance of our attacks to attacks representing state-of-the-art for each norm: Brendel \& Bethge attack~(B\&B, $l_{\infty}$-, $l_2$-, $l_1$-, $l_0$-norms)~\cite{brendel2019accurate}; Carlini-Wagner $l_2$-attack~(C\&W, $l_2$-norm)~\cite{carlini2016towards}; CornerSearch $l_0$-attack~(CS, $l_0$-norm)~\cite{croce2019sparse}; Decoupled Direction and Norm $l_2$-attack~(DDN, $l_2$-norm)~\cite{rony2019decoupling}; DeepFool~(DF, $l_{\infty}$-, and $l_2$-norms)~\cite{dezfooli2015deepfool}; Distributionally Adversarial Attack~(DAA, $l_{\infty}$-norm)~\cite{zheng2018distributionally}; Elastic-net attack~(EAD, $l_1$-norm)~\cite{chen2018ead}; Fast Adaptive Boundary Attack~(FAB, $l_{\infty}$-, $l_2$-, $l_1$-norms)~\cite{croce2019minimally}; Jacobian-based Saliency Map attack~(JSMA, $l_0$-norm)~\cite{papernot2016limitations}; One Pixel attack~(Pixel, $l_0$-norm)~\cite{su2019onepixel}; Projected Gradient Descent~(PGD, $l_{\infty}$-, $l_2$-, $l_1$-, $l_0$-norms)~\cite{kurakin2016adversarialexamples, madry2017towards, tramer2019adversarial, croce2019sparse}; Sparsefool~(SF, $l_1$- and $l_0$-norms)~\cite{modas2019sparsefool}. We use B\&B, C\&W, DDN, and EAD attacks from Foolbox~\cite{rauber2017foolbox}; Pixel and JSMA attacks from ART~\cite{nicolae2018art}; PGD $l_{\infty}$-, $l_2$-, and $l_1$-norm attacks as in Cleverhans~\cite{papernot2018cleverhans}; CS, DF, FAB, and SF attacks with the code from the original papers, while we reimplemented DAA and PGD $l_0$-norm attacks.

We conduct all our experiments using Tensorflow~\cite{ttdt2016tensorflow}. The code for the experiments to reproduce all our results is available at~\url{https://github.com/aam-at/cpgd}. For a fair comparison, we perform a hyperparameter search for each attack, model and dataset. We report the results for the best configuration of parameters. For all attacks with multiple restarts, we find optimal parameters using 1 random restart. The parameters optimal for the attack with 1 random restart are used for the attack's experiments with multiple random restarts. Next, we present details about the parameters of all attacks.

\textbf{Attack parameters:}

\begin{itemize}
\item B\&B~\cite{brendel2019accurate} with 1000 iterations on MNIST and {CIFAR-10} and 100\footnote{\label{first}We reduce the number of iterations on R-ImageNet due to the attack's high computational cost.} on R-ImageNet; initial learning rate selected from $\{1.0, 0.1, 0.01\}$; learning rate decay selected from every $\{20, 100\}$ steps.
\item C\&W~\cite{carlini2016towards} with 9 binary search steps; 10000 iterations on {MNIST} and {CIFAR-10}, and 1000\cref{first} on R-ImageNet; learning rate 0.01; initial const 0.01, and no early stopping.
\item CS~\cite{croce2019sparse} with 1000 iterations; top-100 candidates for sampling; 784 and 1024 maximum sparsity on {MNIST} and {CIFAR-10}. We are unable to run CS on R-ImageNet due to the attack's high memory usage.
\item DDN~\cite{rony2019decoupling} with 10000 iterations on MNIST, and 1000 on CIFAR-10 and R-ImageNet; initial epsilon selected from $\{1.0, 0.1\}$; gamma selected from $\{0.1, 0.05, 0.01\}$.
\item DF~\cite{dezfooli2015deepfool} with 100 iterations and 0.02 overshoot.
\item DAA~\cite{zheng2018distributionally} with Lagrangian Blob method; 500 iterations; epsilon step selected from $\{\epsilon, \sfrac{\epsilon}{2}, \sfrac{\epsilon}{5}, \sfrac{\epsilon}{10}, \sfrac{\epsilon}{25}, \sfrac{\epsilon}{50}, \sfrac{\epsilon}{100} \}$ for every epsilon $\epsilon$; surrogate loss selected from cross-entropy and hinge losses.
\item EAD~\cite{chen2018ead} with 9 binary search steps; 10000 iterations on {MNIST} and {CIFAR-10}, and 1000\cref{first} on R-ImageNet; learning rate 0.01; initial const 0.01; beta 0.05; $l_1$ decision rule, and no early stopping.
\item FAB~\cite{croce2019minimally} with 100 iterations. The remaining parameters are set to the values recommended in~\cite{croce2019minimally}.
\item JSMA~\cite{papernot2016limitations} with gamma set to 1.0 and theta selected from $\{\pm 0.1, \pm 1.0\}$. JSMA requires selecting the target. We attack all targets on {MNIST} and {CIFAR-10}. On R-ImageNet, we attack only the second-highest class due to the attack's high computational cost.
\item Pixel~\cite{su2019onepixel} with differential evolution strategy; 400 population size, and 100 iterations. We are unable to provide the results for Pixel attack on R-ImageNet due to its high computational cost.
\item PGD~\cite{kurakin2016adversarialexamples,madry2017towards, croce2019sparse} with 500 iterations; epsilon step selected from $\{\epsilon, \sfrac{\epsilon}{2}, \sfrac{\epsilon}{5}, \sfrac{\epsilon}{10}, \sfrac{\epsilon}{25}, \sfrac{\epsilon}{50}, \sfrac{\epsilon}{100} \}$ for every epsilon $\epsilon$; surrogate loss selected from cross-entropy and hinge losses; optimal sparsity levels selected from $\{10\%, 5\%, 1\%\}$ (PGD-$l_{1}$ only).
\item SF~\cite{modas2019sparsefool} with 20 iterations; epsilon 0.02; lambda incremented from 1 to 5, so the attack always succeeds.
\end{itemize}

\textbf{Parameters for our attack}: we set the number of iterations to 500, so the computational cost of our attack is similar to PGD attack with 500 iterations. Yet, the overall complexity of our attack is lower than PGD since PGD attack needs to be restarted for each threshold. For the primal variable $\mathbf{r}$, we use Adam~\cite{kingma2014adam} and Proximal Adam~\cite{melchior2019proximal} in PDGD and PDPGD attacks, respectively. For the dual variable $\lambda$, we perform gradient ascent in the log domain to guarantee that it remains positive. We apply an exponential moving average to smooth the value of the dual variable during optimisation. We select the learning rate for primal variables from $\{1.0, 0.1, 0.01\}$ using 1 random restart. The optimal learning rate for the attack with 1 random restart is used in the experiments with 10 and 100 random restarts. Learning rate and initial value of the dual variable is set to $0.1$ in all experiments. We exponentially and linearly decay the learning rate for the primal and dual variables to $0.01$ and $0.1$ of its initial value. We sample the initial perturbation from a uniform distribution $\mathcal{U} = [-\epsilon, \epsilon]$ with $\epsilon$ set to 0.5 on MNIST, 0.25 on {CIFAR-10}, and 0.1 on {R-ImageNet} datasets. We finetune the perturbation found after $N$-restarts for an additional 500 iterations.

\textbf{Evaluation metrics}: We define the \textit{robust accuracy} of the model at a threshold $\epsilon$ as the classification accuracy of the model if the adversary is allowed to perturb the input with the perturbation of $l_p$-norm smaller than the threshold $\epsilon$ in order to change the model prediction. Given a perturbation budget $\epsilon$, an adversarial attack aims to maximise the reduction of the model's accuracy. We fix five thresholds per model and per dataset and calculate the robust accuracy of each attack at five selected thresholds. We compare the attacks using the following statistics for each dataset: i) \textbf{avg. rob. accuracy}: the mean of the robust accuracies achieved by the attack over all models and thresholds (lower is better); ii) \textbf{\# best}: how many times the attack achieves the lowest robust accuracy (it is the most effective); iii) \textbf{avg. difference to best}: for each model/threshold we compute the difference between the robust accuracy of the attack and the best one across all the attacks, then we average over all models/thresholds; iv) \textbf{max difference to best}: as "avg. difference to best", but with the maximum difference instead of the average one. In addition, we compare the average norm of the perturbations if the adversary is allowed to perturb the input without any perturbation bound (perturb only correctly classified inputs). Unbounded adversarial attack aims to minimise the perturbation budget while also achieving a high attack success rate. Please note that the comparison using the average norm of the adversarial perturbation is only available for the attacks that minimise the perturbation norm and excludes PGD and DAA attacks.

We compare the attack methods based on their computational complexity in~\Cref{sec:runtime-comparison}. We test the effectiveness of the attacks on MNIST, {CIFAR-10}, and Restricted ImageNet datasets in~\Cref{sec:main-results}. We summarise our main results in~\Cref{tab:all-perf,tab:all-average}. We provide the complete results for all our experiments in supplementary materials, including detailed comparison of the proposed attack with PGD and FAB.

\subsection{Runtime Comparison}\label{sec:runtime-comparison}

It is difficult to compare the speed of various attack methods due to the differences in the per iteration runtime complexity, the number of iterations required for the attack to converge, and the attacks' implementation details. We perform a two-fold comparison of various attack methods based on the theoretical per iteration runtime complexity and the attack's actual running time on the equivalent hardware.

\subsubsection{Runtime Complexity}

We measure the per iteration runtime complexity as the number of forward and backward passes per attack's iteration. Yet, counting only the number of the forward and backward passes is insufficient as attacks at each iteration may perform additional non-trivial operations. For example, B\&B solves a second-order cone program at each iteration, significantly increasing the attack's running time. Our attack requires computing $l_p$-norm proximity operator at an additional cost of $O(d \log d)$ for $l_{\infty}$-norm, $O(d)$ for $l_1$- and $l_0$-norms proximity operators, where $d$ is the input's dimensionality. We summarise the results of the comparison in~\Cref{tab:runtime-complexity}. Because the number of the model's parameters significantly larger than the input's dimensionality, the cost of computing proximity operator is negligible. So, the overall complexity of our attack is similar to PGD attack.

\begin{table}[!t]
  \captionsetup[table]{position=t}
  \renewcommand*{\arraystretch}{1.2}
  \setlength\tabcolsep{3.8pt} \centering
  \caption{Runtime complexity comparison of adversarial attacks.~\label{tab:runtime-complexity}}
\vspace{-0.2cm}
  \begin{tabular}{L{0.9cm}|c|c|L{5.5cm}}
    Attack & \# FW & \# BP & Extra cost                                                                                                \\
    \hline
    B\&B   & $1$   & $1$   & Solve SOCP                                                                                                \\
    CS     & $1$   & $-$   & $-$                                                                                                       \\
    C\&W   & $1$   & $1$   & $-$                                                                                                       \\
    DAA    & $1$   & $1$   & Compute pairwise distance matrix $O(d^2)$                                                                  \\
    DDN    & $1$   & $1$   & $l_2$-ball projection $O(d)$                                                                               \\
    DF     & $1$   & $k$   & $l_{p}$-ball projection $O(d)$                                                                             \\
    EAD    & $1$   & $1$   & $l_1$-norm proximity operator $O(d)$                                                                      \\
    FAB    & $2$   & $k$   & $l_p$-ball box projection $O(d \log d)$                                                                   \\
    JSMA   & $1$   & $k$   & $-$                                                                                                       \\
    Pixel  & $1$   & $-$   & $-$                                                                                                       \\
    PGD    & $1$   & $1$   & $l_p$-ball projection: $O(d)$ for $l_{\infty}$- and $l_2$-norms; $O(d \log d)$ for $l_1$- and $l_0$-norms \\
    SF     & $1$   & $-$   & Compute DF and $l_1$-norm projection onto hyperplane                                                      \\
    \hline
    Our    & 1     & 1     & $l_p$-norm proximity operator: $O(d \log d)$ for $l_{\infty}$-norm, $O(d)$ for $l_1$- and $l_0$-norms
  \end{tabular}
\end{table}

\subsubsection{Running Time}

We report the running time in seconds on Nvidia Titan V averaged across all models for 1000 points on MNIST, {CIFAR-10} and Restricted ImageNet. Unless otherwise stated, the running time includes all the restarts. For PGD, DAA and Pixel attacks, this is the time for evaluating robust accuracy at five thresholds. Note that when measuring the running time for these attacks, we exploit the fact that the inputs non-robust at the threshold $\epsilon$ are also non-robust at thresholds larger than $\epsilon$. For the other attacks, a single attack is sufficient to compute the robust accuracy at all thresholds. \textbf{MNIST}: B\&B 328s; CS 2611s; C\&W 3758s; DAA-100 4680s; DDN 254s; DF 12s; EAD 4812s; FAB-100 2974s; JSMA 275s; Pixel 41126s; PGD-100 2135s for $l_{\infty}/l_2$ and 2583 for $l_1/l_0$; SF 1301s; Our-10/Our-100 217s/1805s for $l_{\infty}$, 122s/965s for $l_2$ and 176s/1484s for $l_1/l_0$. \textbf{\mbox{CIFAR-10}}: B\&B 1327s; CS 23603s; C\&W 15230s; DAA-100 29030s; DDN 165s; DF 13s; EAD 16222s; FAB-100 20590s; JSMA 1333s; Pixel 22928s; PGD-100 12267s for $l_{\infty}/l_2$ and 12361s for $l_1/l_0$; SF 790s; Our-10/Our-100 1358s/12060s for $l_{\infty}$, 1157s/10054s for $l_2$ and 1316s/11620s for $l_1/l_0$. \textbf{Restricted ImageNet}: B\&B 10942s; C\&W 36451s; DAA-10 87194s; DDN 3703s; DF 173s; EAD 46762s; FAB-10 33757s; JSMA 19551s; PGD-10 32671s for $l_{\infty}/l_2$ and 53858s for $l_1/l_0$; SF 16367s; Our-1/Our-10 6818s/38719s for $l_{\infty}$, 5127s/29335s for $l_2$ and 5644s/30586s for $l_1/l_0$.

The running time depends upon the number of attack's iterations. DF is the fastest attack requiring few iterations to succeed, but it is also the least accurate attack as its goal is to find adversarial perturbation as fast as possible without minimising its norm. The running time of our attack is comparable to existing attacks. PGD attack is fast and accurate with the complexity similar to our attack, but it needs to be restarted at each threshold, so the total running time of our attack is lower than PGD. FAB converges faster and requires relatively small number of iterations (100 in our experiments), but the complexity of each iteration is higher than our attack as it requires to compute $k$ gradients at each step. We also report the running time of our lower complexity attack with a reduced number of restarts, which is the fastest when excluding less accurate DF and SF attacks. As we show in the section, our lower complexity attack often outperforms state-of-the-art attacks.

\subsection{Main Results}\label{sec:main-results}

\begin{figure}[!b]
  \captionsetup[table]{position=b}
  \captionsetup[subfloat]{captionskip=1pt}
  \caption{$l_{\infty}$-, $l_2$-, $l_1$-, and $l_0$-norm adversarial examples for a naturally trained, $l_{\infty}$-, and $l_2$- adversarially trained models on MNIST.~\label{fig:mnist_adversarial_examples}}
  \setlength\tabcolsep{0pt} \subfloat[][Original images~\label{tab:original_examples_mnist}]{
    \begin{tabular}{C{6mm} @{\hspace{3pt}} l}
      \\[-0.7cm]
      &  \raisebox{-.5\height}{\includegraphics[width=0.90\columnwidth]{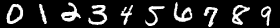}} \\
    \end{tabular}
  }\\[-2.4ex]
  \subfloat[][Adversarial examples for naturally trained model~\label{tab:natural_examples_mnist}]{
    \begin{tabular}{C{6mm} @{\hspace{3pt}} l}
      \centered{$l_{\infty}$} &  \raisebox{-.5\height}{\includegraphics[width=0.90\columnwidth]{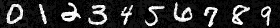}} \\
      \centered{$l_{2}$} &  \raisebox{-.5\height}{\includegraphics[width=0.90\columnwidth]{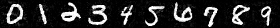}} \\
      \centered{$l_{1}$} &  \raisebox{-.5\height}{\includegraphics[width=0.90\columnwidth]{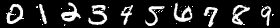}} \\
      \centered{$l_{0}$} &  \raisebox{-.5\height}{\includegraphics[width=0.90\columnwidth]{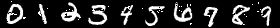}} \\
    \end{tabular}
    \hfill
  }\\[-2.4ex]
  \subfloat[][Adversarial examples for $l_{\infty}$-AT model~\label{tab:linf_examples_mnist}]{
    \begin{tabular}{C{6mm} @{\hspace{3pt}} l}
      \centered{$l_{\infty}$} &  \raisebox{-.5\height}{\includegraphics[width=0.90\columnwidth]{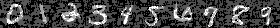}} \\
      \centered{$l_{2}$} &  \raisebox{-.5\height}{\includegraphics[width=0.90\columnwidth]{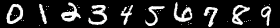}} \\
      \centered{$l_{1}$} &  \raisebox{-.5\height}{\includegraphics[width=0.90\columnwidth]{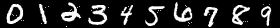}} \\
      \centered{$l_{0}$} &  \raisebox{-.5\height}{\includegraphics[width=0.90\columnwidth]{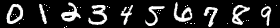}} \\
    \end{tabular}
  }\\[-2.4ex]
  \subfloat[][Adversarial examples for $l_2$-AT model~\label{tab:l2_examples_mnist}]{
    \begin{tabular}{C{6mm} @{\hspace{3pt}} l}
      \centered{$l_{\infty}$} &  \raisebox{-.5\height}{\includegraphics[width=0.90\columnwidth]{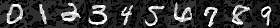}} \\
      \centered{$l_{2}$} &  \raisebox{-.5\height}{\includegraphics[width=0.90\columnwidth]{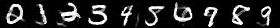}} \\
      \centered{$l_{1}$} &  \raisebox{-.5\height}{\includegraphics[width=0.90\columnwidth]{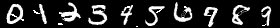}} \\
      \centered{$l_{0}$} &  \raisebox{-.5\height}{\includegraphics[width=0.90\columnwidth]{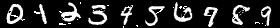}} \\
    \end{tabular}
  }
\end{figure}

We compare the attacks on the first 1000 images of {MNIST} and {CIFAR-10} test sets and 1000 images of {Restricted ImageNet} validation set. For each dataset, we evaluate robust accuracy at five thresholds for \textit{plain}, $l_{\infty}$~($l_{\infty}$-AT) and $l_2$~($l_2$-AT) adversarially trained models. We use the same five thresholds that were selected in~\cite{croce2019minimally}. We show adversarial examples generated by our attack for 10 randomly selected MNIST test images in~\Cref{fig:mnist_adversarial_examples}. We report aggregated results for all datasets, models, norms and thresholds in~\Cref{tab:all-perf}, while we provide complete results at each threshold in supplementary materials. Note that we are unable to run CornerSearch and Pixel attacks on R-ImageNet due to high memory usage and high computational cost of the attacks.

\begin{table*}[!t]
  \centering
  \captionsetup[table]{position=t}
  \captionsetup[subfloat]{captionskip=1pt,justification=centering}
  \renewcommand*{\arraystretch}{1}
  \caption{Performance summary (aggregated) of all attacks on MNIST, CIFAR-10 and Restricted Imagenet in~\Cref{tab:mnist-perf}, \Cref{tab:cifar10-perf} and \Cref{tab:imagenet-perf}, respectively. *Note that for Our-10 the "\# best" is computed excluding the results of Our-100.~\label{tab:all-perf}}
  \vspace{-0.5cm}
  \subfloat[][MNIST results~\label{tab:mnist-perf}]{
    \begin{tabular}{L{25mm}*{8}{|C{14mm} }}
      \cellcolor{gray!20} \textbf{$l_\infty$-norm} & & DeepFool & B\&B & DAA-100 & PGD-100 & FAB-100 & Our-10 & Our-100 \\
      \hline
      avg. rob. acc.     & & 78.46 & 57.51 & 55.93 & 56.24 & 59.47 & 56.00 & \textbf{55.15} \\
      \# best            & & 0     & 2     & 5     & 3     & 1     & 8$^*$ & \textbf{14}    \\
      avg. diff. to best & & 23.33 & 2.38  & 0.80  & 1.10  & 4.34  & 0.87  & \textbf{0.02}  \\
      max diff. to best  & & 81.6  & 7.6   & 4.0   & 4.2   & 24.6  & 3.9   & \textbf{0.3}   \\
      \hline
      \multicolumn{9}{c}{}\\[-0.05cm]
      \cellcolor{gray!20}\textbf{$l_2$-norm} & DeepFool & C\&W & DDN & B\&B & PGD-100 & FAB-100 & Our-10 & Our-100 \\
      \hline
      avg. rob. acc.     & 67.4  & 48.13 & 42.81 & 39.57 & 45.65 & 34.66 & \textbf{33.91} & \textbf{33.05} \\
      \# best            & 0     & 1     & 2     & 2     & 1     & 4     & 12$^*$         & \textbf{14}    \\
      avg. diff. to best & 34.35 & 15.09 & 9.76  & 6.52  & 12.6  & 1.61  & 0.87           & \textbf{0.01}  \\
      max diff. to best  & 91.9  & 63.9  & 53.3  & 30.9  & 66.3  & 5.0   & 4.1            & \textbf{0.1}   \\
      \hline
      \multicolumn{9}{c}{}\\[-0.05cm]
      \cellcolor{gray!20}\textbf{$l_1$-norm} & & SparseFool & EAD & B\&B & PGD-100 & FAB-100 & Our-10 & Our-100 \\
      \hline
      avg. rob. acc.     & & 74.33 & 45.36 & 48.85 & 58.59 & 37.75 & \textbf{34.19} & \textbf{32.15} \\
      \# best            & & 0     & 1     & 0     & 0     & 0     & 12$^*$         & \textbf{15}    \\
      avg. diff. to best & & 42.18 & 13.21 & 16.69 & 26.44 & 5.6   & 2.03           & \textbf{0.0}   \\
      max diff. to best  & & 91.9  & 50.8  & 52.6  & 80.3  & 19.0  & 8.2            & \textbf{0.0}   \\
      \hline
      \multicolumn{9}{c}{}\\[-0.05cm]
      \cellcolor{gray!20}\textbf{$l_0$-norm} & SparseFool &  JSMA & Pixel & B\&B & PGD-100 & CornerSearch &  Our-10 & Our-100 \\
      \hline
      avg. rob. acc.     & 80.43 & 84.98 & 77.03 & 58.37 & 56.09 & 49.99 &  \textbf{43.15} & \textbf{40.57} \\
      \# best            & 0     & 0     & 0     & 0     & 0     & 2     &  6$^*$          & \textbf{14}    \\
      avg. diff. to best & 38.83 & 44.55 & 36.61 & 17.94 & 15.66 & 9.57  &  2.73           & \textbf{0.15}  \\
      max diff. to best  & 92.5  & 96.1  & 94.3  & 57.0  & 65.9  & 36.0  &  11.9           & \textbf{2.2}   \\
      \hline
    \end{tabular}
  }\\
  \vspace{-0.2cm}
  \subfloat[][CIFAR-10 results~\label{tab:cifar10-perf}]{
    \raggedright  
    \begin{tabular}{L{25mm}*{8}{|C{14mm} }}
      \cellcolor{gray!20} \textbf{$l_\infty$-norm}&  & DeepFool  & B\&B  &   DAA-100  &   PGD-100  &   FAB-100  & Our-10  & Our-100 \\
      \hline
      avg. rob. acc.     & & 40.63 & 32.75 & 31.72 & 31.65 & 31.61 & \textbf{31.24} & \textbf{30.83} \\
      \# best            & & 0     & 0     & 0     & 2     & 0     & 13$^*$         & \textbf{14}    \\
      avg. diff. to best & & 9.81  & 1.93  & 0.89  & 0.83  & 0.78  & 0.41           & \textbf{0.01}  \\
      max diff. to best  & & 17.4  & 3.0   & 1.7   & 1.7   & 1.4   & 1.1            & \textbf{0.1}   \\
      \hline
      \multicolumn{9}{c}{}\\[-0.05cm]
      \cellcolor{gray!20}\textbf{$l_2$-norm}& DeepFool  & C\&W  & DDN  & B\&B  & PGD-100  & FAB-100  & Our-10  & Our-100 \\
      \hline
      avg. rob. acc.     & 44.81 & 37.49 & 37.77 & 38.49 & 36.90 & 36.63 & \textbf{36.09} & \textbf{35.89} \\
      \# best            & 0     & 0     & 1     & 0     & 2     & 1     & 14$^*$         & \textbf{14}    \\
      avg. diff. to best & 8.92  & 1.60  & 1.89  & 2.60  & 1.01  & 0.75  & 0.20           & \textbf{0.01}  \\
      max diff. to best  & 14.8  & 2.8   & 4.7   & 4.5   & 2.2   & 1.4   & 0.6            & \textbf{0.1}   \\
      \hline
      \multicolumn{9}{c}{}\\[-0.05cm]
      \cellcolor{gray!20}\textbf{$l_1$-norm} &  & SparseFool  &    EAD  & B\&B  &  PGD-100  &   FAB-100  & Our-10  & Our-100 \\
      \hline
      avg. rob. acc.     & & 40.53 & 20.76 & 22.79 & 26.46 & 21.49 & \textbf{19.59} & \textbf{18.36} \\
      \# best            & & 0     &  0    & 0     & 0     & 0     & 12$^*$         & \textbf{15}    \\
      avg. diff. to best & & 22.17 & 2.40  & 4.43  & 8.10  & 3.13  & 1.23           & \textbf{0.00}  \\
      max diff. to best  & & 35.9  & 4.9   & 9.1   & 13.8  & 6.5   & 2.5            & \textbf{0.0}   \\
      \hline
      \multicolumn{9}{c}{}\\[-0.05cm]
      \cellcolor{gray!20}\textbf{$l_0$-norm} & SparseFool &  JSMA & Pixel & B\&B & PGD-100 & CornerSearch & Our-10 & Our-100 \\
      \hline
      avg. rob. acc.     & 48.45 & 69.15 & 51.66 & 47.86 & 35.23 & 30.51 & \textbf{27.33} & \textbf{25.00} \\
      \# best            & 0     & 0     & 0     & 0     & 0     & 4     & 10$^*$         & \textbf{11}    \\
      avg. diff. to best & 25.91 & 46.61 & 29.12 & 25.32 & 12.69 & 7.97  & 4.79           & \textbf{2.46}  \\
      max diff. to best  & 45.7  & 68.6  & 53.3  & 39.7  & 16.4  & 20.6  & 13.0           & \textbf{11.5}  \\
      \hline
    \end{tabular}
  }\\
  \vspace{-0.2cm}
  \subfloat[][Restricted ImageNet results~\label{tab:imagenet-perf}]{
    \raggedleft
    \begin{tabular}{L{25mm}*{8}{|C{14mm} }}
      \cellcolor{gray!20} \textbf{$l_\infty$-norm}& & DeepFool & B\&B & DAA-10 & PGD-10 & FAB-10 & Our-1 & Our-10 \\
      \hline
      avg. rob. acc.     & & 36.99 & 29.26 & 27.99 & \textbf{27.39} & 28.50 & 28.05 & 27.54 \\
      \# best            & & 0     & 0     & 2     & 7              & 0     & 5     & \textbf{9}    \\
      avg. diff. to best & & 9.72  & 1.99  & 0.72  & \textbf{0.11}  & 1.23  & 0.78  & 0.27 \\
      max diff. to best  & & 18.5  & 4.5   & 1.8   & \textbf{0.5}   & 2.6   & 2.9   & 2.1 \\
      \hline
      \multicolumn{8}{c}{}\\[-0.05cm]
      \cellcolor{gray!20}\textbf{$l_2$-norm} & DeepFool & C\&W & DDN & B\&B & PGD-10 & FAB-10 & Our-1 & Our-10 \\
      \hline
      avg. rob. acc.     & 45.80 & 42.25 & 32.87 & 36.17 & 32.70 & 34.57 & \textbf{32.09} & \textbf{31.21} \\
      \# best            & 0     & 0     & 0     & 0     & 1     & 0     & 10             & \textbf{14}    \\
      avg. diff. to best & 14.67 & 11.12 & 1.74  & 5.03  & 1.57  & 3.44  & 0.95           & \textbf{0.08} \\
      max diff. to best  & 28.4  & 43.9  & 5.1   & 10.0  & 5.0   & 8.0   & 2.9            & \textbf{1.2} \\
      \hline
      \multicolumn{8}{c}{}\\[-0.05cm]
      \cellcolor{gray!20}\textbf{$l_1$-norm} & & SparseFool & EAD & B\&B & PGD-10 & FAB-10 & Our-1 & Our-10 \\
      \hline
      avg. rob. acc.     & & 67.51 & 36.05 & 41.46 & 58.89 & 41.79 & 38.19 & \textbf{35.21} \\
      \# best            & & 0     &  6    & 0     & 0     & 0     & 2     & \textbf{10}    \\
      avg. diff. to best & & 32.83 & 1.37  & 6.78  & 24.21 & 7.11  & 3.51  & \textbf{0.53}  \\
      max diff. to best  & & 61.6  & 6.4   & 10.3  & 54.5  & 16.6  & 8.2   & \textbf{3.1}   \\
      \hline
      \multicolumn{8}{c}{}\\[-0.05cm]
      \cellcolor{gray!20}\textbf{$l_0$-norm} & SparseFool & JSMA & Pixel & B\&B & PGD-10 & CornerSearch & Our-1 & Our-10 \\
      \hline
      avg. rob. acc.     & 49.18 & 79.82 & - & 46.28 & 46.56 & - & \textbf{33.40} & \textbf{28.75} \\
      \# best            & 0     & 0     & - & 0     & 0     & - & 14             & \textbf{15}    \\
      avg. diff. to best & 20.43 & 51.07 & - & 17.53 & 17.81 & -  & 4.65           & \textbf{0.00} \\
      max diff. to best  & 40.1  & 74.2  & - & 25.9  & 36.0  & - & 9.0            & \textbf{0.0} \\
      \hline
    \end{tabular}
    \hspace{1.74cm}
  }
\end{table*}

Our attack is the strongest as it most substantially reduces the robust accuracy compared with other attacks~(see~\cref{tab:all-perf}). In particular, it outperforms other attacks in 11 out of 12 cases in terms of the average robust accuracy. The second best attack varies depending upon the dataset and norm, which shows that our attack is accurate and general. PGD $l_{\infty}$-norm attack on Restricted ImageNet is the only attack to outperform our method. The improvement over the state-of-the-art attacks is most significant for our $l_1$- and $l_0$-norm attacks. For example, our $l_1$- and $l_0$-norm attack reduces the average robust accuracy by 17.4\% and 23.2\% on MNIST against FAB-100 and CornerSearch attacks, respectively. Our fast lower complexity attack with 10 random restarts on MNIST and {CIFAR-10} and 1 random restart on Restricted ImageNet outperforms other attacks in 9 out of 12 cases.

\begin{table}[!t]
  \captionsetup[table]{position=t}
  \captionsetup[subfloat]{captionskip=1pt}
  \renewcommand*{\arraystretch}{1.05}
  \caption{Average norm of the perturbation found by the attacks (when successful, excluding the already misclassified points) for every model on MNIST, CIFAR-10 and Restricted ImageNet.~\label{tab:all-average}}
  \addtocounter{table}{-1}
  \setlength\tabcolsep{3.8pt} \centering
  \subfloat[][Average norm of the perturbation on MNIST]{
    \begin{tabular}{@{\hspace{2pt}} c @{\hspace{2pt}} | @{\hspace{4pt}} *{7}{r}}
      \\[-0.5cm]
      \cellcolor{gray!20}$l_{\infty}\times 10^{-1}$  &  &  & DF  & B\&B  & FAB-100  & Our-10  & Our-100 \\
      \hline
      plain         & & & 0.82 & 0.65 & 0.66 & \textbf{0.63} & \textbf{0.63} \\
      $l_\infty$-at & & & 5.28 & 3.2  & 3.27 & \textbf{3.18} & \textbf{3.16} \\
      $l_2$-at      & & & 2.59 & 1.72 & 1.7  & \textbf{1.67} & \textbf{1.66} \\
      \hline
      \multicolumn{7}{c}{}\\[-1.2ex]
      \cellcolor{gray!20}$l_2$  & DF  & C\&W  & DDN  & B\&B  & FAB-100  & Our-10  & Our-100 \\
      \hline
      plain         & 1.13 & 1.01 & 1.0  & 1.01 & 0.99 & \textbf{0.98} & \textbf{0.98} \\
      $l_\infty$-at & 5.03 & 2.08 & 1.71 & 1.4  & 1.12 & \textbf{1.11} & \textbf{1.06} \\
      $l_2$-at      & 3.08 & 2.34 & 2.29 & 2.35 & 2.25 & \textbf{2.20} & \textbf{2.18} \\
      \hline
      \multicolumn{7}{c}{}\\[-1.2ex]
      \cellcolor{gray!20}$l_1$  & \multicolumn{2}{r}{Sparsefool}  & EAD & B\&B & FAB-100 & Our-10 & Our-100 \\
      \hline
      plain           & \multicolumn{2}{r}{8.71}   & 6.21  & 6.59  & 6.04  & \textbf{6.03}  & \textbf{5.87} \\
      $l_{\infty}$-at & \multicolumn{2}{r}{207.70} & 6.73  & 6.32  & 3.48  & \textbf{2.59}  & \textbf{2.32} \\
      $l_2$-at        & \multicolumn{2}{r}{16.52}  & 11.96 & 13.97 & 12.15 & \textbf{11.40} & \textbf{10.93} \\
      \hline
      \multicolumn{7}{c}{}\\[-1.2ex]
      \cellcolor{gray!20}$l_0$  & \multicolumn{2}{r}{Sparsefool}  & JSMA  & B\&B  & CS  & Our-10  & Our-100 \\
      \hline
      plain           & \multicolumn{2}{r}{12.45}  & 13.76 & 7.66  & 8.74  & \textbf{7.01}  & \textbf{6.76} \\
      $l_{\infty}$-at & \multicolumn{2}{r}{249.06} & 59.30 & 11.29 & 3.85  & 4.27  & \textbf{3.82} \\
      $l_2$-at        & \multicolumn{2}{r}{21.35}  & 26.61 & 15.14 & 17.28 & \textbf{12.28} & \textbf{11.72} \\
      \hline
    \end{tabular}
  }\\
  \subfloat[][Average norm of the perturbation on CIFAR-10]{
    \begin{tabular}{@{\hspace{2pt}} c @{\hspace{2pt}} | @{\hspace{4pt}} *{7}{r}}
      \\[-0.55cm]
      \cellcolor{gray!20}$l_{\infty}\times 10^{-2}$  &  &  & DF  & B\&B  & FAB-100  & Our-10  & Our-100 \\
      \hline
      plain         & & & 0.77 & 0.57 & 0.56 & \textbf{0.55} & \textbf{0.54} \\
      $l_\infty$-at & & & 3.13 & 2.5  & 2.37 & \textbf{2.34} & \textbf{2.32} \\
      $l_2$-at      & & & 2.59 & 1.98 & 1.94 & \textbf{1.90} & \textbf{1.88} \\
      \hline
      \multicolumn{7}{c}{}\\[-1.2ex]
      \cellcolor{gray!20}$l_2 \times 10^{-1}$  & DF  & C\&W  & DDN  & B\&B  & FAB-100  & Our-10  & Our-100 \\
      \hline
      plain         & 2.72 & 2.13 & 2.11  & 2.15 & 2.06 & \textbf{2.04} & \textbf{2.02} \\
      $l_\infty$-at & 9.42 & 7.28 & 7.51 & 7.62  & 7.15 & \textbf{6.99} & \textbf{6.95} \\
      $l_2$-at      & 9.06 & 7.12 & 7.38 & 7.22 & 6.98 & \textbf{6.85} & \textbf{6.82} \\
      \hline
      \multicolumn{7}{c}{}\\[-1.2ex]
      \cellcolor{gray!20}$l_1$  & \multicolumn{2}{r}{Sparsefool}  & EAD & B\&B & FAB-100 & Our-10 & Our-100 \\
      \hline
      plain           & \multicolumn{2}{r}{6.99}  & 2.90 & 2.83 & 2.86 & \textbf{2.67} & \textbf{2.58} \\
      $l_{\infty}$-at & \multicolumn{2}{r}{10.74} & 5.63 & 6.53 & 6.07 & \textbf{5.38} & \textbf{5.09} \\
      $l_2$-at        & \multicolumn{2}{r}{13.77} & 7.79 & 8.78 & 8.02 & \textbf{7.50} & \textbf{7.16} \\
      \hline
      \multicolumn{7}{c}{}\\[-1.2ex]
      \cellcolor{gray!20}$l_0$  & \multicolumn{2}{r}{Sparsefool}  & JSMA  & B\&B  & CS  & Our-10  & Our-100 \\
      \hline
      plain           & \multicolumn{2}{r}{17.73} & 21.29 & 6.80 & 3.98 & \textbf{3.16} & \textbf{2.86} \\
      $l_{\infty}$-at & \multicolumn{2}{r}{9.09}  & 25.03 & 9.60 & 5.76 & \textbf{4.38} & \textbf{3.99} \\
      $l_2$-at        & \multicolumn{2}{r}{27.46} & 26.13 & 9.78 & 6.50 & \textbf{5.07} & \textbf{4.60} \\
      \hline
    \end{tabular}
  }\\
  \subfloat[][Average norm of the perturbation on Restricted ImageNet]{
    \begin{tabular}{@{\hspace{2pt}} c @{\hspace{2pt}} | @{\hspace{4pt}} *{7}{r}}
      \\[-0.55cm]
      \cellcolor{gray!20}$l_{\infty}\times 10^{-2}$ &  &  & DF   & B\&B & FAB-10 & Our-1 & Our-10 \\
      \hline
      plain                                         &  &  & 0.25 & 0.19 & 0.19   & \textbf{0.19}  & \textbf{0.18}  \\
      $l_\infty$-at                                 &  &  & 2.16 & 1.75 & 1.71   & \textbf{1.68}  & \textbf{1.67} \\
      $l_2$-at                                      &  &  & 1.87 & 1.58 & 1.54   & \textbf{1.53}  & \textbf{1.51} \\
      \hline
      \multicolumn{8}{c}{}\\[-1.2ex]
      \cellcolor{gray!20}$l_2$ & DF   & C\&W & DDN  & B\&B & FAB-10 & Our-1 & Our-10  \\
      \hline
      plain                    & 0.54 & 0.75 & 0.40 & 0.42 & 0.41   & \textbf{0.39}  & \textbf{0.38} \\
      $l_\infty$-at            & 3.43 & 2.34 & 2.28 & 2.41 & 2.35   & \textbf{2.17}  & \textbf{2.10} \\
      $l_2$-at                 & 4.53 & 3.65 & 3.56 & 3.78 & 3.67   & \textbf{3.56}  & \textbf{3.54} \\
      \hline
      \multicolumn{8}{c}{}\\[-1.2ex]
      \cellcolor{gray!20}$l_1$ & \multicolumn{2}{r}{Sparsefool} & EAD            & B\&B   & FAB-10 & Our-1 & Our-10 \\
      \hline
      plain                    & \multicolumn{2}{r}{87.56}      & 15.61          & 16.67  & 21.25  & 15.65 & \textbf{13.48} \\
      $l_{\infty}$-at          & \multicolumn{2}{r}{138.35}     & \textbf{40.42} & 52.48  & 52.78  &  53.14     & 47.87 \\
      $l_2$-at                 & \multicolumn{2}{r}{374.02}     & 166.27         & 188.17 & 179.35 & 169.74      & \textbf{165.71} \\
      \hline
      \multicolumn{8}{c}{}\\[-1.2ex]
      \cellcolor{gray!20}$l_0$ & \multicolumn{2}{r}{Sparsefool} & JSMA   & B\&B   & CS & Our-1  & Our-10 \\
      \hline
      plain                    & \multicolumn{2}{r}{52.06}      & 211.77 & 33.60  & -  & \textbf{22.96}  & \textbf{19.21} \\
      $l_{\infty}$-at          & \multicolumn{2}{r}{80.33}      & 372.33 & 62.14  & - &  \textbf{53.66} & \textbf{41.84} \\
      $l_2$-at                 & \multicolumn{2}{r}{131.27}     & 629.18 & 191.86 & - & \textbf{123.90} & \textbf{117.80} \\
      \hline
    \end{tabular}\hspace{1.3cm}
  }
\end{table}
\addtocounter{table}{1}

Robust accuracy measures the model's robustness at a specific threshold $\epsilon$. Robust norm gives a full picture of the model's robustness as a continuous function of the perturbation's size $\epsilon$. In~\Cref{tab:all-average}, we report the average $l_p$-norm of the adversarial perturbations found by the attacks~(when successful) for every dataset, model and norm. We exclude the points that the models already misclassify. All attacks, except DF attack on MNIST dataset against $l_{\infty}$-AT model, have a 100\% success rate, so we do not include the attack's success rate in the table results.

Our attack finds the smallest norm adversarial perturbation in 35 out of 36 cases~(see~\cref{tab:all-average}). The improvements for $l_1$- and $l_0$-norm minimisation are particularly significant. For example, our $l_1$-norm attack reduces the average robust norm of $l_{\infty}$-AT model on MNIST by 50\% compared to the second-best attack FAB-100 (a reduction from 3.48 to 2.32). Our $l_0$-norm attack reduces the average robust norm of $l_2$-AT model on MNIST by 29.2\% compared to the second-best attack B\&B (reduction from 15.14 to 11.72). EAD $l_1$-attack against $l_{\infty}$-AT model on Restricted ImageNet is the only attack to outperform our method when comparing the average robust $l_1$-norm (an increase from 40.42 to 47.87). We also report the results of our attack with the reduced number of random restarts. Our lower complexity attack with 10 random restarts outperforms all other attacks in 23 out of 24 cases on MNIST and {CIFAR-10} datasets. Our-1 outperforms all other attacks in 9 out of 12 cases on R-ImageNet dataset.

\textbf{To summarize, our main results are:}
\begin{itemize}
\item Our $l_{\infty}$-norm attack reduces the average robust norm~/~robust accuracy by $\nicefrac{1.4\%}{1.5\%}$ on MNIST, $\nicefrac{2.2\%}{2.5\%}$ on CIFAR-10 and $\nicefrac{2.2\%}{-0.5\%}$ on R-ImageNet compared to the second-best attack. {PGD-10} on R-ImageNet is the only attack to outperform our attack in terms of the average robust accuracy.
\item Our $l_2$-norm attack reduces the average robust norm / robust accuracy by $\nicefrac{3.8\%}{4.9\%}$ on MNIST, $\nicefrac{2.5\%}{2.0\%}$ on {CIFAR-10} and $\nicefrac{4.2\%}{4.8\%}$ on R-ImageNet compared to the second-best attack.
\item Our $l_1$-norm attack reduces the average robust norm~/ robust accuracy by $\nicefrac{21.7\%}{17.4\%}$ on MNIST, $\nicefrac{9.7\%}{13\%}$ on {CIFAR-10} and $\nicefrac{0.2\%}{2.4\%}$ on R-ImageNet compared to the second-best attack. EAD on R-ImageNet against $l_{\infty}$-AT is the only attack to outperform our attack in terms of the average robust norm.
\item Our $l_0$-norm attack reduces the average robust norm / robust accuracy by $\nicefrac{14.5\%}{23.2\%}$ on {MNIST}, $\nicefrac{41.6\%}{22.0\%}$ on {CIFAR-10} and $\nicefrac{45.0\%}{61.0\%}$ on R-ImageNet compared to the second-best attack.
\end{itemize}

Overall, our attack is the best attack to reduce the robust accuracy and the robust norm compared to state-of-the-art attacks with a similar computational budget. Our attack is fast, accurate and general as it works for all $l_p$-norms in $p \in \{0, 1, 2, \infty\}$. It outperforms all algorithms, including even one that is specialised in individual norms. Our lower complexity attack with the reduced number of restarts is the second-best attack and requires a fraction of the running time / computational cost.

\section{Conclusion}

Fast and accurate estimation of the robust norm and robust accuracy of deep neural networks is crucial for comparing models. However, evaluating the DNNs robustness has proven to be challenging. The original non-convex constrained norm minimisation problem is difficult to solve. In this work, we introduce an adversarial attack that efficiently solves the original attack's problem. We interpret optimising the Lagrangian of the adversarial attack as playing a two-player game. The first player minimises the Lagrangian wrt the adversarial noise; the second player maximises the Lagrangian wrt the regularisation penalty, which penalises the first player for violating the misclassification constraint. We apply a primal-dual gradient descent algorithm to simultaneously update primal and dual variables to find the minimal optimal adversarial perturbation. For non-smooth $l_p$-norm minimisation, such as $l_{\infty}$-, $l_1$-, and $l_0$-norms, we introduce primal-dual proximal gradient descent attack. We also derive group $l_{0,G}$-norm proximal operator, which we use to minimise the number of perturbed pixels. Our method is fast, accurate and general. In the experiments on MNIST, CIFAR-10 and Restricted ImageNet, we show that our attack outperforms state-of-the-art $l_{\infty}$-, $l_2$-, $l_1$- and $l_0$-norm attacks in terms of robust norm and robust accuracy in 35 out 36 and 11 out of 12 cases, respectively. In future work, we plan to extend the proposed attack to multiple norm perturbations and combine our attack with adversarial training defence.

\ifCLASSOPTIONcaptionsoff
\newpage
\fi

\renewcommand\tablename{Table}
\renewcommand\appendixname{}
\makeatletter
\long\def\@makecaption#1#2{\ifx\@captype\@IEEEtablestring \footnotesize\bgroup\par\centering\@IEEEtabletopskipstrut{\normalfont\footnotesize
  #1.} {\normalfont\footnotesize\scshape #2}\par\addvspace{0.5\baselineskip}\egroup \@IEEEtablecaptionsepspace
\else
\@IEEEfigurecaptionsepspace
\setbox\@tempboxa\hbox{\normalfont\footnotesize {#1.}\nobreakspace\nobreakspace #2}\ifdim \wd\@tempboxa >\hsize \setbox\@tempboxa\hbox{\normalfont\footnotesize {#1.}\nobreakspace\nobreakspace}\parbox[t]{\hsize}{\normalfont\footnotesize\noindent\unhbox\@tempboxa#2}\else \ifCLASSOPTIONconference \hbox to\hsize{\normalfont\footnotesize\hfil\box\@tempboxa\hfil}\else \hbox to\hsize{\normalfont\footnotesize\box\@tempboxa\hfil}\fi\fi\fi}
\makeatother

\appendices
\onecolumn
\section{Detailed Experimental Results}

\begin{table*}[!bp]
\captionsetup[table]{position=t}
\captionsetup[subfloat]{captionskip=1pt}
\renewcommand*{\arraystretch}{1}
\vspace*{-26pt}
\caption{Comparison of $l_{\infty}$, $l_2$, $l_1$, and $l_0$-attacks on a naturally trained, $l_{\infty}$-, and $l_2$- adversarially trained models on MNIST.\label{tab:mnist_results}}
\addtocounter{table}{-1}
\vspace{-8mm}
\subfloat[][$l_{\infty}$-attack~\label{tab:mnist_linf}]{
\setlength\tabcolsep{3.95pt} \begin{tabular}{@{\hspace{4pt}}C{10mm}@{\hspace{4pt}} | @{\hspace{3pt}} C{6mm} @{\hspace{3pt}} | @{\hspace{4pt}} *{14}{r}}
\multicolumn{16}{c}{}\\
\diagbox[width=13mm,height=5mm]{\tiny model}{\tiny $l_{\infty}$} & $\epsilon$ & DF & B\&B & DAA-1 & DAA-10 & DAA-100 & PGD-1 & PGD-10 & PGD-100 & FAB-1 & FAB-10 & FAB-100 & Our-1 & Our-10 & Our-100\\
\hline
\multirow{5}{*}{plain}
& 0.03 & 93.6 & \textbf{91.8} & 91.9 & 91.9 & 91.9 & 92.0 & 91.9 & 91.9 & 92.0 & 92.0 & 92.0 & 91.9          & 91.9          & \textbf{91.8} \\
& 0.05 & 84.4 & 74.6          & 75.6 & 74.7 & 74.4 & 75.6 & 74.8 & 74.4 & 77.1 & 76.9 & 76.1 & \textbf{73.9} & \textbf{73.6} & \textbf{73.2} \\
& 0.07 & 65.0 & 40.3          & 42.8 & 41.1 & 40.1 & 42.5 & 40.3 & 39.7 & 44.8 & 43.3 & 42.7 & \textbf{39.4} & \textbf{38.1} & \textbf{37.9} \\
& 0.09 & 38.5 & 13.1          & 15.0 & 13.3 & 12.4 & 15.1 & 13.1 & 12.2 & 16.0 & 15.0 & 14.6 & 12.5          & \textbf{11.1} & \textbf{10.5} \\
& 0.11 & 16.7 & 2.0           & 3.0  & 2.4  & 2.2  & 2.7  & 2.2  & 2.0  & 3.4  & 3.1  & 2.4  & 2.1           & \textbf{1.4}  & \textbf{1.3} \\
\hline
\multirow{5}{*}{$l_{\infty}$-at}
& 0.2   & 95.2 & 94.2 & 94.1 & 94.0 & \textbf{93.7} & 94.7 & 93.9 & \textbf{93.7} & 94.6 & 94.3 & 93.9 & 94.2 & \textbf{93.7} & \textbf{93.7} \\
& 0.25  & 94.7 & 91.8 & 92.2 & 91.2 & \textbf{91.1} & 93.1 & 91.9 & 91.2          & 93.4 & 91.9 & 91.7 & 92.0 & 91.5          & \textbf{91.1} \\
& 0.3   & 93.9 & 89.1 & 88.7 & 87.4 & \textbf{87.0} & 91.4 & 88.5 & 87.6          & 91.3 & 89.5 & 88.8 & 89.1 & 88.4          & 87.3 \\
& 0.325 & 92.1 & 63.3 & 63.6 & 58.9 & 57.4          & 73.4 & 62.9 & 59.0          & 86.5 & 83.4 & 81.3 & 65.3 & 60.0          & \textbf{56.7} \\
& 0.35  & 89.5 & 14.6 & 13.4 & 9.4  & \textbf{7.9}  & 26.8 & 14.0 & 10.8          & 50.2 & 30.9 & 24.7 & 17.0 & 11.8          & \textbf{7.9} \\
\hline
\multirow{5}{*}{$l_2$-at}
& 0.05 & 96.7 & \textbf{96.3} & 96.4 & \textbf{96.3} & \textbf{96.3} & \textbf{96.3} & \textbf{96.3} & \textbf{96.3} & 96.4 & \textbf{96.3} & \textbf{96.3} & 96.4 & \textbf{96.3} & \textbf{96.3} \\
& 0.1  & 93.8 & 90.3          & 90.7 & 90.1          & 90.0          & 90.5          & 90.2          & 90.0          & 90.8 & 90.5          & 90.4          & 90.2 & \textbf{90.0} & \textbf{89.7} \\
& 0.15 & 86.9 & 72.6          & 73.0 & 71.4          & 70.7          & 73.0          & 71.6          & 70.8          & 74.0 & 72.4          & 71.7          & 71.8 & 70.9          & \textbf{70.2} \\
& 0.2  & 76.0 & 26.9          & 29.8 & 25.2          & 23.3          & 29.0          & 25.9          & 23.5          & 33.6 & 27.7          & 24.6          & 24.8 & \textbf{20.8} & \textbf{19.3} \\
& 0.25 & 59.9 & 1.8           & 0.9  & 0.9           & 0.6           & 1.5           & 0.6           & \textbf{0.4}  & 1.7  & 1.0           & 0.9           & 1.0  & 0.5           & \textbf{0.4} \\
\hline
\end{tabular}
}\\[-3.5ex]
\subfloat[][$l_2$-attack~\label{tab:mnist_l2}]{
\setlength\tabcolsep{6.0pt} \begin{tabular}{@{\hspace{4pt}}C{10mm}@{\hspace{4pt}} | @{\hspace{3pt}} C{6mm} @{\hspace{3pt}} | @{\hspace{4pt}} *{13}{r}}
\multicolumn{15}{c}{}\\
\diagbox[width=13mm,height=5mm]{\tiny model}{\tiny $l_2$} & $\epsilon$ & DF & C\&W & DDN & B\&B & PGD-1 & PGD-10 & PGD-100 & FAB-1 & FAB-10 & FAB-100 & Our-1 & Our-10 & Our-100\\
\hline
\multirow{5}{*}{plain}
& 0.5 & 92.1 & 91.1         & \textbf{90.8} & 91.1         & 91.1         & 91.1         & 91.1         & 91.1         & 91.1         & 91.1         & \textbf{90.8} & \textbf{90.8} & \textbf{90.8} \\
& 1   & 60.1 & 48.4         & 46.8          & 49.0         & 49.8         & 47.8         & 47.6         & 48.4         & 47.8         & 47.2         & 47.7          & \textbf{45.3} & \textbf{45.2} \\
& 1.5 & 18.6 & 9.5          & 8.0           & 8.4          & 10.0         & 8.2          & 7.8          & 8.4          & 8.2          & 7.8          & 8.9           & \textbf{6.4}  & \textbf{6.0} \\
& 2   & 1.9  & 0.9          & 0.8           & 0.6          & 0.9          & 0.7          & 0.6          & 0.7          & 0.6          & 0.6          & 0.8           & \textbf{0.3}  & \textbf{0.3} \\
& 2.5 & 0.1  & \textbf{0.0} & \textbf{0.0}  & \textbf{0.0} & \textbf{0.0} & \textbf{0.0} & \textbf{0.0} & \textbf{0.0} & \textbf{0.0} & \textbf{0.0} & \textbf{0.0}  & \textbf{0.0}  & \textbf{0.0} \\
\hline
\multirow{5}{*}{$l_{\infty}$-at}
& 1   & 94.4 & 87.6 & 87.0 & 78.3         & 90.9 & 90.1 & 88.3 & 83.7 & 71.7         & 64.7         & 72.6         & \textbf{63.8} & \textbf{59.7} \\
& 1.5 & 93.3 & 72.9 & 62.3 & 39.9         & 85.4 & 80.2 & 75.3 & 45.9 & 19.6         & 12.3         & 27.7         & 13.0          & \textbf{9.0} \\
& 2   & 92.1 & 54.5 & 31.6 & 11.3         & 74.6 & 63.2 & 51.1 & 15.3 & 2.4          & 0.7          & 3.3          & 0.8           & \textbf{0.2} \\
& 2.5 & 90.0 & 32.4 & 10.4 & 2.3          & 59.0 & 36.9 & 24.3 & 4.0  & 0.1          & \textbf{0.0} & 0.2          & \textbf{0.0}  & \textbf{0.0} \\
& 3   & 87.2 & 14.3 & 1.8  & \textbf{0.0} & 40.8 & 16.9 & 6.8  & 2.0  & \textbf{0.0} & \textbf{0.0} & \textbf{0.0} & \textbf{0.0}  & \textbf{0.0} \\
\hline
\multirow{5}{*}{$l_2$-at}
& 1   & 93.5 & 92.3 & 92.4 & 92.9 & 92.3 & 92.3 & 92.3 & 92.3 & 92.3 & \textbf{92.2} & 92.7 & 92.3          & 92.3 \\
& 1.5 & 87.2 & 84.5 & 83.6 & 85.4 & 84.0 & 83.6 & 83.6 & 84.2 & 83.8 & 83.7          & 84.1 & \textbf{83.6} & \textbf{83.3} \\
& 2   & 79.2 & 70.4 & 68.4 & 71.1 & 68.7 & 68.0 & 67.6 & 70.2 & 69.1 & 68.2          & 68.6 & \textbf{67.2} & \textbf{66.0} \\
& 2.5 & 67.5 & 45.9 & 42.7 & 46.6 & 44.2 & 39.9 & 38.0 & 46.5 & 41.7 & 39.6          & 40.7 & \textbf{36.6} & \textbf{34.6} \\
& 3   & 53.8 & 17.3 & 15.5 & 16.6 & 15.6 & 11.3 & 10.3 & 19.0 & 14.2 & 11.8          & 11.5 & \textbf{8.6}  & \textbf{8.4} \\
\hline
\end{tabular}
}\\[-3.5ex]
\subfloat[][$l_1$-attack~\label{tab:mnist_l1}]{
\setlength\tabcolsep{6pt} \begin{tabular}{@{\hspace{4pt}}C{10mm}@{\hspace{4pt}} | @{\hspace{3pt}} C{6mm} @{\hspace{3pt}} | @{\hspace{5pt}} *{12}{r}}
\multicolumn{13}{c}{}\\
\diagbox[width=13mm,height=5mm]{\tiny model}{\tiny $l_1$} & $\epsilon$ & SparseFool & EAD & B\&B & PGD-1 & PGD-10 & PGD-100 & FAB-1 & FAB-10 & FAB-100 & Our-1 & Our-10 & Our-100 \\
\hline
\multirow{5}{*}{plain}
& 2  & 95.0 & \textbf{93.3} & 94.2 & 94.3 & 94.1 & 94.0 & 94.2 & 93.8 & 93.6 & 94.2 & 93.5          & \textbf{93.3} \\
& 4  & 86.8 & 75.5          & 79.7 & 81.0 & 78.0 & 76.9 & 79.6 & 76.1 & 75.0 & 79.7 & 75.4          & \textbf{73.7} \\
& 6  & 70.6 & 45.9          & 51.8 & 58.4 & 50.6 & 48.1 & 53.4 & 47.4 & 43.5 & 55.2 & 43.9          & \textbf{41.3} \\
& 8  & 50.4 & 25.0          & 28.6 & 36.1 & 30.2 & 28.0 & 31.6 & 24.5 & 22.4 & 32.7 & \textbf{22.3} & \textbf{20.5} \\
& 10 & 32.7 & 9.7           & 12.6 & 16.0 & 13.2 & 12.3 & 15.3 & 10.6 & 8.3  & 16.6 & \textbf{8.3}  & \textbf{6.8} \\
\hline
\multirow{5}{*}{$l_{\infty}$-at}
& 2.5  & 94.5 & 88.0 & 84.2 & 93.4 & 93.1 & 92.8 & 85.2 & 69.9 & 56.8 & 65.3          & \textbf{46.0} & \textbf{37.8} \\
& 5    & 92.5 & 54.4 & 56.2 & 87.6 & 85.8 & 83.9 & 57.6 & 32.4 & 19.4 & \textbf{18.2} & \textbf{6.2}  & \textbf{3.6} \\
& 7.5  & 92.0 & 31.2 & 29.7 & 81.6 & 75.6 & 71.6 & 47.0 & 15.6 & 7.3  & \textbf{4.9}  & \textbf{0.5}  & \textbf{0.1} \\
& 10   & 91.4 & 17.0 & 14.1 & 77.3 & 67.4 & 59.8 & 39.3 & 8.0  & 3.0  & \textbf{0.9}  & \textbf{0.0}  & \textbf{0.0} \\
& 12.5 & 90.8 & 10.3 & 6.7  & 71.3 & 56.9 & 46.8 & 33.7 & 5.6  & 0.9  & \textbf{0.1}  & \textbf{0.0}  & \textbf{0.0} \\
\hline
\multirow{5}{*}{$l_2$-at}
& 5     & 91.7 & 89.8 & 92.7 & 90.7 & 90.3 & 90.2 & 90.5 & 90.3 & 89.9 & 90.5 & \textbf{88.9} & \textbf{88.5} \\
& 8.75  & 80.2 & 71.3 & 78.9 & 75.5 & 73.8 & 72.8 & 75.6 & 73.6 & 72.4 & 74.1 & \textbf{69.2} & \textbf{66.7} \\
& 12.5  & 65.3 & 42.4 & 58.4 & 60.5 & 56.8 & 54.4 & 55.3 & 48.8 & 45.8 & 50.9 & \textbf{38.4} & \textbf{33.7} \\
& 16.25 & 49.3 & 19.1 & 30.6 & 46.4 & 39.0 & 32.3 & 32.7 & 25.0 & 20.3 & 24.8 & \textbf{15.3} & \textbf{12.7} \\
& 20    & 31.8 & 7.5  & 14.3 & 23.7 & 18.3 & 14.9 & 14.9 & 9.5  & 7.7  & 10.0 & \textbf{4.9}  & \textbf{3.6} \\
\hline
\end{tabular}
}\\[-3.5ex]
\subfloat[][$l_0$-attack~\label{tab:mnist_l0}]{
\setlength\tabcolsep{6pt} \begin{tabular}{@{\hspace{4pt}}C{10mm}@{\hspace{4pt}} | @{\hspace{3pt}} C{6mm} @{\hspace{3pt}} | @{\hspace{5pt}} *{11}{r}}
\multicolumn{11}{c}{}\\
\diagbox[width=13mm,height=5mm]{\tiny model}{\tiny $l_{\infty}$} & $\epsilon$ & SparseFool & JSMA & Pixel & B\&B & PGD-1 & PGD-10 & PGD-100 & CornerSearch & Our-1 & Our-10 & Our-100 \\
\hline
\multirow{5}{*}{plain}
& 1  & 98.4 & 98.7 & 97.7 & 97.7 & 97.8 & 97.6 & 97.3 & 97.1 &  97.5 & \textbf{97.1} & \textbf{97.0} \\
& 3  & 95.3 & 96.6 & 91.6 & 90.9 & 92.4 & 90.1 & 89.4 & 89.0 &  90.3 & \textbf{88.7} & \textbf{88.2} \\
& 5  & 89.0 & 92.9 & 88.9 & 72.0 & 77.6 & 68.2 & 65.0 & 72.0 &  72.3 & 65.2          & \textbf{62.7} \\
& 8  & 73.9 & 73.4 & 87.2 & 36.5 & 47.9 & 31.9 & 27.9 & 44.5 &  38.1 & 28.1          & \textbf{25.4} \\
& 12 & 45.3 & 48.2 & 85.2 & 7.2  & 17.1 & 7.9  & 5.2  & 19.9 &  10.9 & \textbf{4.9}  & \textbf{4.0} \\
\hline
\multirow{5}{*}{$l_{\infty}$-at}
& 1  & 97.7 & 98.5 & 92.5 & 95.2 & 98.2 & 97.3 & 95.0 & \textbf{90.9} & 93.8 & 91.0         & \textbf{90.9} \\
& 3  & 96.5 & 97.7 & 61.8 & 85.5 & 97.6 & 93.8 & 89.1 & \textbf{46.5} & 75.2 & 58.4         & 48.7 \\
& 5  & 95.5 & 97.1 & 29.8 & 72.8 & 96.1 & 89.8 & 81.7 & 16.4          & 47.7 & 24.3         & \textbf{15.8} \\
& 8  & 93.8 & 96.9 & 21.4 & 49.1 & 94.2 & 83.6 & 65.5 & 3.6           & 16.9 & 4.7          & \textbf{2.9} \\
& 12 & 92.7 & 96.2 & 17.9 & 31.7 & 89.9 & 68.0 & 42.1 & 0.3           & 3.7  & \textbf{0.2} & \textbf{0.1} \\
\hline
\multirow{5}{*}{$l_2$-at}
& 5  & 94.8 & 96.3 & 97.4 & 93.6 & 92.3 & 91.1 & 89.9 & 92.1 &  91.2 & 90.0          & \textbf{89.2} \\
& 10 & 83.4 & 87.9 & 96.6 & 75.5 & 75.1 & 66.8 & 61.8 & 77.1 &  69.9 & \textbf{61.6} & \textbf{58.5} \\
& 15 & 68.7 & 80.4 & 96.4 & 45.0 & 47.1 & 29.2 & 23.3 & 56.1 &  35.7 & 24.5          & \textbf{20.1} \\
& 20 & 50.5 & 64.1 & 96.0 & 16.8 & 21.2 & 9.9  & 6.8  & 29.7 &  13.7 & \textbf{6.8}  & \textbf{4.3} \\
& 25 & 31.0 & 49.8 & 95.1 & 6.0  & 8.5  & 3.1  & 1.3  & 14.7 &  4.0  & 1.8           & \textbf{0.8} \\
\hline
\end{tabular}
}
\end{table*}
\addtocounter{table}{1}

\begin{table*}[!bp]
\captionsetup[table]{position=t}
\captionsetup[subfloat]{captionskip=1pt}
\renewcommand*{\arraystretch}{1}
\caption{Comparison of $l_{\infty}$-, $l_2$-, $l_1$-, and $l_0$-attacks on a naturally trained, $l_{\infty}$-, and $l_2$- adversarially trained models on {CIFAR-10}.~\label{tab:cifar10_results}}
\addtocounter{table}{-1}
\vspace{-5mm}
\subfloat[][$l_{\infty}$-attack~\label{tab:cifar10_linf}]{
\setlength\tabcolsep{3.90pt} \begin{tabular}{@{\hspace{4pt}}C{10mm}@{\hspace{4pt}} | @{\hspace{3pt}} C{6mm} @{\hspace{3pt}} | @{\hspace{4pt}} *{14}{r}}
\multicolumn{16}{c}{}\\
\diagbox[width=13mm,height=5mm]{\tiny model}{\tiny $l_{\infty}$} & $\epsilon$ & DF & B\&B & DAA-1 & DAA-10 & DAA-100 & PGD-1 & PGD-10 & PGD-100 & FAB-1 & FAB-10 & FAB-100 & Our-1 & Our-10 & Our-100\\
\hline
\multirow{5}{*}{plain}
& \sfrac{1}{255}   & 63.0 & 57.1 & 56.9 & 56.0 & 56.0 & 56.1 & 55.6 & \textbf{55.5} & 56.5 & 55.9 & 55.8 & 56.4 & 56.1          & 55.6          \\
& \sfrac{1.5}{255} & 49.7 & 37.9 & 39.0 & 37.7 & 37.5 & 39.0 & 37.8 & 37.4          & 38.7 & 37.7 & 37.2 & 37.7 & 37.5          & \textbf{36.6} \\
& \sfrac{2}{255}   & 37.8 & 23.0 & 23.3 & 22.5 & 22.1 & 23.3 & 22.4 & 22.0          & 22.8 & 22.4 & 21.4 & 22.0 & \textbf{21.5} & \textbf{20.4} \\
& \sfrac{2.5}{255} & 26.9 & 12.6 & 13.3 & 12.8 & 12.3 & 13.5 & 12.7 & 12.3          & 13.0 & 12.3 & 11.8 & 12.0 & \textbf{11.3} & \textbf{10.6} \\
& \sfrac{3}{255}   & 19.2 & 6.0  & 7.2  & 6.5  & 5.8  & 7.1  & 6.2  & 6.1           & 6.7  & 6.0  & 5.5  & 5.9  & \textbf{5.1}  & \textbf{4.6}  \\
\hline
\multirow{5}{*}{$l_{\infty}$-at}
& \sfrac{2}{255}  & 66.8 & 66.5 & 65.6 & 65.6 & 65.5 & 65.5 & 65.5 & \textbf{65.4} & 65.8 & 65.7 & 65.7 & 65.5          & \textbf{65.4} & \textbf{65.4} \\
& \sfrac{4}{255}  & 53.0 & 50.3 & 49.5 & 49.1 & 48.9 & 49.4 & 49.2 & 48.9          & 49.2 & 49.1 & 48.9 & 49.4          & \textbf{48.9} & \textbf{48.5} \\
& \sfrac{6}{255}  & 42.7 & 36.9 & 35.4 & 34.8 & 34.6 & 35.2 & 34.8 & 34.6          & 35.3 & 34.8 & 34.6 & \textbf{34.6} & \textbf{34.0} & \textbf{33.9} \\
& \sfrac{8}{255}  & 32.6 & 25.3 & 24.1 & 24.0 & 23.7 & 24.2 & 23.8 & 23.6          & 23.8 & 23.5 & 23.4 & 23.7          & \textbf{23.0} & \textbf{22.6} \\
& \sfrac{10}{255} & 24.3 & 16.1 & 14.7 & 14.4 & 13.9 & 14.7 & 14.4 & 14.0          & 15.2 & 14.6 & 14.5 & 14.3          & \textbf{13.7} & \textbf{13.2} \\
\hline
\multirow{5}{*}{$l_2$-at}
& \sfrac{2}{255}  & 64.3 & 63.0 & 62.7 & 62.5 & 62.4 & 62.6 & 62.6 & 62.4 & 62.7 & 62.6 & 62.6 & 62.5          & \textbf{62.4} & \textbf{62.3} \\
& \sfrac{4}{255}  & 49.1 & 45.1 & 44.6 & 44.4 & 44.4 & 44.6 & 44.4 & 44.3 & 44.4 & 44.2 & 44.2 & \textbf{44.1} & \textbf{43.7} & \textbf{43.7} \\
& \sfrac{6}{255}  & 36.8 & 28.1 & 27.9 & 27.6 & 27.2 & 27.9 & 27.5 & 27.2 & 27.1 & 26.8 & 26.8 & 27.2          & \textbf{26.2} & \textbf{25.9} \\
& \sfrac{8}{255}  & 25.8 & 15.0 & 14.7 & 14.3 & 14.2 & 14.6 & 14.2 & 13.9 & 14.9 & 14.1 & 13.8 & 13.9          & \textbf{13.2} & \textbf{12.7} \\
& \sfrac{10}{255} & 17.5 & 8.4  & 8.5  & 7.6  & 7.3  & 8.7  & 7.5  & 7.2  & 8.6  & 8.1  & 7.9  & 7.4           & \textbf{6.6}  & \textbf{6.5} \\
\hline
\end{tabular}
}\\[-3.5ex]
\subfloat[][$l_2$-attack~\label{tab:cifar10_l2}]{
\setlength\tabcolsep{6.0pt} \begin{tabular}{@{\hspace{4pt}}C{10mm}@{\hspace{4pt}} | @{\hspace{3pt}} C{6mm} @{\hspace{3pt}} | @{\hspace{4pt}} *{13}{r}}
\multicolumn{15}{c}{}\\
\diagbox[width=13mm,height=5mm]{\tiny model}{\tiny $l_2$} & $\epsilon$ & DF & C\&W & DDN & B\&B & PGD-1 & PGD-10 & PGD-100 & FAB-1 & FAB-10 & FAB-100 & Our-1 & Our-10 & Our-100 \\
\hline
\multirow{5}{*}{plain}
& 0.1  & 71.7 & 68.9 & 67.5 & 68.5 & 67.9 & 67.6 & 67.5 & 68.1 & 68.1 & 67.9 & \textbf{67.5} & \textbf{67.5} & \textbf{67.4} \\
& 0.15 & 61.8 & 56.4 & 55.0 & 56.3 & 55.8 & 55.1 & 54.8 & 55.7 & 55.2 & 55.0 & \textbf{54.6} & \textbf{53.7} & \textbf{53.6} \\
& 0.2  & 51.8 & 44.0 & 43.5 & 44.0 & 43.8 & 43.2 & 43.0 & 43.4 & 42.8 & 42.4 & 42.6          & \textbf{41.9} & \textbf{41.3} \\
& 0.3  & 34.1 & 22.1 & 21.5 & 23.1 & 22.5 & 21.9 & 21.5 & 21.2 & 20.1 & 20.0 & 20.5          & \textbf{19.8} & \textbf{19.3} \\
& 0.4  & 20.1 & 9.4  & 9.7  & 9.8  & 10.2 & 9.5  & 9.0  & 9.7  & 8.5  & 8.3  & 8.6           & \textbf{7.9}  & \textbf{7.4}  \\
\hline
\multirow{5}{*}{$l_{\infty}$-at}
& 0.25 & 67.2 & 65.5 & 64.8 & 65.8 & \textbf{64.7} & \textbf{64.7} & \textbf{64.7} & 64.9 & 64.8 & \textbf{64.7} & \textbf{64.7} & \textbf{64.7} & \textbf{64.7} \\
& 0.5  & 53.3 & 49.2 & 49.0 & 51.1 & 49.2          & 49.1          & 48.9          & 49.3 & 49.0 & 49.0          & \textbf{48.8} & \textbf{48.6} & \textbf{48.5} \\
& 0.75 & 42.3 & 33.8 & 34.7 & 35.4 & 34.0          & 33.8          & 33.5          & 34.0 & 33.4 & 33.3          & \textbf{33.1} & \textbf{32.8} & \textbf{32.7} \\
& 1    & 32.0 & 22.2 & 23.3 & 24.1 & 22.2          & 21.6          & 21.2          & 21.8 & 21.2 & 20.8          & \textbf{20.2} & \textbf{19.9} & \textbf{19.6} \\
& 1.25 & 23.2 & 11.6 & 13.9 & 14.0 & 12.3          & 11.6          & 11.5          & 12.3 & 11.5 & 11.2          & \textbf{10.6} & \textbf{10.1} & \textbf{9.9}  \\
\hline
\multirow{5}{*}{$l_2$-at}
& 0.25 & 66.5 & 65.8 & \textbf{65.3} & 66.1 & 65.4 & 65.4 & \textbf{65.3} & 65.4 & 65.4 & 65.4 & 65.5          & 65.4          & 65.4          \\
& 0.5  & 54.3 & 49.2 & 49.3          & 50.4 & 49.4 & 49.1 & 49.1          & 49.6 & 49.2 & 48.9 & \textbf{48.8} & \textbf{48.5} & \textbf{48.5} \\
& 0.75 & 41.9 & 33.1 & 33.2          & 34.0 & 33.4 & 33.1 & 32.7          & 33.1 & 32.8 & 32.4 & \textbf{32.3} & \textbf{31.9} & \textbf{31.8} \\
& 1    & 29.9 & 20.1 & 21.7          & 22.1 & 20.6 & 20.3 & 20.1          & 20.7 & 20.0 & 19.7 & \textbf{19.1} & \textbf{18.8} & \textbf{18.8} \\
& 1.25 & 22.0 & 11.0 & 14.2          & 12.6 & 11.5 & 11.0 & 10.7          & 11.4 & 11.1 & 10.5 & \textbf{10.3} & \textbf{9.8}  & \textbf{9.5}  \\
\hline
\end{tabular}
}\\[-3.5ex]
\subfloat[][$l_1$-attack~\label{tab:cifar10_l1}]{
\setlength\tabcolsep{6.0pt} \begin{tabular}{@{\hspace{4pt}}C{10mm}@{\hspace{4pt}} | @{\hspace{3pt}} C{6mm} @{\hspace{3pt}} | @{\hspace{5pt}} *{12}{r}}
\multicolumn{13}{c}{}\\
\diagbox[width=13mm,height=5mm]{\tiny model}{\tiny $l_1$} & $\epsilon$ & SparseFool & EAD & B\&B & PGD-1 & PGD-10 & PGD-100 & FAB-1 & FAB-10 & FAB-100 & Our-1 & Our-10 & Our-100 \\
\hline
\multirow{5}{*}{plain}
& 2  & 70.0 & 53.2 & 51.7 & 54.8 & 53.3 & 52.8 & 56.1 & 52.0 & 50.4 & 51.8 & \textbf{50.0} & \textbf{48.3} \\
& 4  & 54.4 & 23.1 & 22.5 & 25.8 & 24.6 & 23.8 & 31.3 & 25.0 & 22.7 & 23.2 & \textbf{20.5} & \textbf{19.0} \\
& 6  & 40.9 & 8.0  & 7.2  & 12.1 & 10.1 & 9.5  & 17.0 & 10.1 & 8.0  & 8.1  & \textbf{6.0}  & \textbf{5.0}  \\
& 8  & 30.0 & 2.1  & 1.9  & 6.7  & 5.5  & 4.5  & 7.7  & 3.3  & 2.5  & 2.1  & \textbf{1.3}  & \textbf{1.1}  \\
& 10 & 20.9 & 0.6  & 0.4  & 3.6  & 2.7  & 2.4  & 4.6  & 1.4  & 1.1  & 0.9  & 0.7           & \textbf{0.2}  \\
\hline
\multirow{5}{*}{$l_{\infty}$-at}
& 5     & 53.3 & 36.1 & 41.0 & 46.6 & 45.5 & 43.8 & 43.4 & 39.7 & 38.3 & 39.1 & \textbf{34.5} & \textbf{32.0} \\
& 8.75  & 38.1 & 18.3 & 22.9 & 30.3 & 28.6 & 27.6 & 26.0 & 22.2 & 20.3 & 21.4 & \textbf{15.6} & \textbf{13.8} \\
& 12.5  & 27.2 & 7.0  & 10.9 & 21.3 & 18.5 & 17.2 & 14.8 & 11.1 & 8.7  & 9.8  & \textbf{6.2}  & \textbf{5.4}  \\
& 16.25 & 19.2 & 2.3  & 4.6  & 15.1 & 12.9 & 11.8 & 7.1  & 4.7  & 3.5  & 4.9  & \textbf{2.3}  & \textbf{1.9}  \\
& 20    & 12.5 & 0.5  & 1.6  & 12.4 & 10.7 & 10.0 & 3.9  & 2.0  & 1.2  & 1.9  & 0.8           & \textbf{0.3}  \\
\hline
\multirow{5}{*}{$l_2$-at}
& 3  & 67.3 & 62.5 & 63.7 & 65.1 & 64.9 & 64.8 & 63.4 & 63.1 & 63.0 & 63.5 & \textbf{62.2} & \textbf{61.7} \\
& 6  & 58.3 & 44.5 & 47.0 & 50.4 & 49.3 & 48.7 & 49.0 & 47.2 & 45.8 & 47.3 & \textbf{42.3} & \textbf{41.0} \\
& 9  & 47.6 & 27.2 & 32.1 & 37.3 & 35.4 & 34.4 & 33.5 & 30.7 & 28.8 & 31.5 & \textbf{26.5} & \textbf{24.3} \\
& 12 & 37.5 & 16.7 & 20.9 & 30.7 & 27.9 & 26.5 & 23.7 & 20.2 & 17.6 & 20.2 & \textbf{15.5} & \textbf{14.2} \\
& 15 & 30.7 & 9.3  & 13.4 & 24.3 & 21.5 & 19.1 & 16.5 & 12.6 & 10.5 & 13.2 & 9.4           & \textbf{7.2} \\
\hline
\end{tabular}
}\\[-3.5ex]
\subfloat[][$l_0$-attack~\label{tab:cifar10_l0}]{
\setlength\tabcolsep{6pt} \begin{tabular}{@{\hspace{4pt}}C{10mm}@{\hspace{4pt}} | @{\hspace{3pt}} C{6mm} @{\hspace{3pt}} | @{\hspace{5pt}} *{11}{r}}
\multicolumn{11}{c}{}\\
\diagbox[width=13mm,height=5mm]{\tiny model}{\tiny $l_{\infty}$} & $\epsilon$ & SparseFool & JSMA & Pixel & B\&B & PGD-1 & PGD-10 & PGD-100 & CornerSearch & Our-1 & Our-10 & Our-100 \\
\hline
\multirow{5}{*}{plain}
& 1  & 81.3 & 88.6 & 71.3 & 83.1 & 82.2 & 75.6 & 72.2 & \textbf{56.5} & 72.8          & 69.3          & 67.2          \\
& 3  & 67.3 & 82.6 & 43.1 & 64.7 & 70.6 & 54.8 & 43.1 & 33.0          & 39.6          & \textbf{31.6} & \textbf{26.7} \\
& 5  & 52.8 & 75.7 & 35.8 & 46.8 & 57.5 & 36.1 & 23.5 & 27.7          & \textbf{17.5} & \textbf{10.6} & \textbf{7.1}  \\
& 8  & 33.3 & 65.2 & 36.2 & 25.5 & 37.0 & 17.8 & 9.9  & 15.1          & \textbf{4.1}  & \textbf{2.2}  & \textbf{1.1}  \\
& 12 & 16.7 & 53.3 & 33.2 & 8.1  & 19.8 & 8.3  & 4.5  & 4.9           & \textbf{0.5}  & \textbf{0.2}  & \textbf{0.0}  \\
\hline
\multirow{5}{*}{$l_{\infty}$-at}
& 1  & 70.3 & 80.3 & 68.2 & 76.3 & 74.5 & 72.1 & 69.4 & \textbf{54.6} & 70.2          & 67.6          & 66.0 \\
& 3  & 57.4 & 75.3 & 49.5 & 64.0 & 62.8 & 51.0 & 43.8 & 36.0          & 45.5          & 39.9          & \textbf{35.9} \\
& 5  & 46.4 & 68.7 & 46.2 & 49.5 & 51.7 & 38.3 & 32.7 & 32.9          & \textbf{30.0} & \textbf{21.7} & \textbf{18.6} \\
& 8  & 34.1 & 60.1 & 47.7 & 35.1 & 39.0 & 27.1 & 21.3 & 20.4          & \textbf{14.3} & \textbf{8.1}  & \textbf{6.1} \\
& 12 & 22.7 & 51.6 & 47.4 & 19.8 & 30.4 & 20.7 & 14.9 & 10.9          & \textbf{6.0}  & \textbf{3.0}  & \textbf{1.7} \\
\hline
\multirow{5}{*}{$l_2$-at}
& 1  & 72.8 & 79.6 & 71.2 & 77.0 & 75.0 & 73.2 & 71.2 & \textbf{58.0} & 71.9          & 70.3          & 69.5 \\
& 3  & 64.1 & 75.4 & 58.5 & 66.8 & 66.1 & 57.2 & 52.4 & \textbf{41.5} & 54.0          & 48.5          & 44.8 \\
& 5  & 55.2 & 71.4 & 56.0 & 56.4 & 55.0 & 42.9 & 37.3 & 38.8          & \textbf{35.0} & \textbf{29.5} & \textbf{25.4} \\
& 10 & 33.7 & 59.7 & 56.6 & 30.4 & 38.7 & 26.3 & 19.4 & 17.6          & \textbf{10.9} & \textbf{5.9}  & \textbf{4.2} \\
& 15 & 18.7 & 49.7 & 54.0 & 14.4 & 27.8 & 18.0 & 12.8 & 9.7           & \textbf{2.8}  & \textbf{1.5}  & \textbf{0.7} \\
\hline
\end{tabular}
}
\hfill
\end{table*}
\addtocounter{table}{1}

\begin{table*}[!bp]
\captionsetup[table]{position=t}
\captionsetup[subfloat]{captionskip=1pt}
\renewcommand*{\arraystretch}{1}
\caption{Comparison of $l_{\infty}$-, $l_2$-, $l_1$-, and $l_0$-attacks on a naturally trained, $l_{\infty}$-, and $l_2$- adversarially trained models on Restricted Imagenet.~\label{tab:imagenet_results}}
\addtocounter{table}{-1}
\vspace{-8mm}
\subfloat[][$l_{\infty}$-attack~\label{tab:imagenet_linf}]{
\setlength\tabcolsep{6pt} \begin{tabular}{@{\hspace{4pt}}C{10mm}@{\hspace{4pt}} | @{\hspace{3pt}} C{6mm} @{\hspace{3pt}} | @{\hspace{4pt}} *{10}{r}}
\multicolumn{12}{c}{}\\
\diagbox[width=13mm,height=5mm]{\tiny model}{\tiny $l_{\infty}$} & $\epsilon$ & DF & B\&B & DAA-1 & DAA-10 & PGD-1 & PGD-10 & FAB-1 & FAB-10 & Our-1 & Our-10 \\
\hline
\multirow{5}{*}{plain}
 & \sfrac{0.25}{255} & 81.0 & 76.6 & 76.5 & 76.5 & 76.6 & 76.5          & 77.1 & 77.1 & \textbf{76.5} & \textbf{76.4} \\
 & \sfrac{0.5}{255}  & 55.1 & 41.0 & 40.3 & 40.1 & 39.1 & 38.8          & 41.4 & 41.1 & 39.9          & \textbf{38.5} \\
 & \sfrac{0.75}{255} & 30.8 & 14.1 & 14.1 & 13.9 & 12.8 & \textbf{12.3} & 15.2 & 14.9 & 13.6          & 12.4          \\
 & \sfrac{1}{255}    & 14.0 & 3.9  & 4.1  & 3.9  & 3.6  & 3.5           & 4.0  & 3.9  & 3.9           & \textbf{3.0}  \\
 & \sfrac{1.25}{255} & 6.6  & 0.9  & 0.9  & 0.9  & 0.7  & 0.7           & 1.3  & 1.0  & \textbf{0.7}  & \textbf{0.6}  \\

\hline
\multirow{5}{*}{$l_{\infty}$-at}
& \sfrac{2}{255}  & 77.0 & 76.5 & \textbf{75.4} & \textbf{75.4} & \textbf{75.4} & \textbf{75.4} & 76.3 & 76.2 & \textbf{75.4} & \textbf{75.4} \\
& \sfrac{4}{255}  & 54.9 & 50.3 & \textbf{47.3} & \textbf{47.3} & \textbf{47.3} & \textbf{47.3} & 49.0 & 48.9 & 48.3          & 47.4          \\
& \sfrac{6}{255}  & 34.4 & 23.5 & 21.1          & 20.9          & \textbf{19.4} & \textbf{19.4} & 22.0 & 21.7 & 20.8          & 20.1          \\
& \sfrac{8}{255}  & 19.5 & 7.8  & 7.1           & 7.1           & 6.2           & 6.1           & 7.3  & 7.2  & 6.2           & \textbf{5.8}  \\
& \sfrac{10}{255} & 11.1 & 2.1  & 1.7           & 1.6           & 1.2           & \textbf{1.1}  & 1.8  & 1.6  & 1.3           & 1.3           \\
\hline
\multirow{5}{*}{$l_2$-at}
& \sfrac{2}{255}  & 75.4 & 74.6 & 73.3 & 73.3 & 73.3          & 73.3          & 74.4 & 74.3 & \textbf{73.2} & \textbf{73.1} \\
& \sfrac{4}{255}  & 49.1 & 43.6 & 39.2 & 39.2 & \textbf{39.1} & \textbf{39.1} & 40.8 & 40.5 & 42.0          & 41.2          \\
& \sfrac{6}{255}  & 27.5 & 16.9 & 14.9 & 14.8 & \textbf{13.0} & \textbf{13.0} & 14.3 & 13.9 & 14.4          & 13.8          \\
& \sfrac{8}{255}  & 12.6 & 5.5  & 4.4  & 4.3  & 3.6           & 3.6           & 4.0  & 3.9  & 3.9           & \textbf{3.5}  \\
& \sfrac{10}{255} & 5.9  & 1.6  & 0.8  & 0.7  & 0.7           & 0.7           & 1.3  & 1.3  & \textbf{0.7}  & \textbf{0.6}  \\
\hline
\end{tabular}
}\\[-3.5ex]
\subfloat[][$l_2$-attack~\label{tab:imagenet_l2}]{
\setlength\tabcolsep{6.0pt} \begin{tabular}{@{\hspace{4pt}}C{10mm}@{\hspace{4pt}} | @{\hspace{3pt}} C{6mm} @{\hspace{3pt}} | @{\hspace{4pt}} *{10}{r}}
\multicolumn{12}{c}{}\\
\diagbox[width=13mm,height=5mm]{\tiny model}{\tiny $l_2$} & $\epsilon$ & DF & C\&W & DDN & B\&B & PGD-1 & PGD-10 & FAB-1 & FAB-10 & Our-1 & Our-10 \\
\hline
\multirow{5}{*}{plain}
& 0.2 & 81.9 & 83.8 & 78.5 & 79.1 & 78.6 & 78.4 & 79.3 & 79.3 & \textbf{77.9} & \textbf{77.4} \\
& 0.4 & 57.9 & 70.4 & 40.9 & 44.9 & 42.5 & 42.1 & 44.4 & 44.3 & \textbf{40.0} & \textbf{39.3} \\
& 0.6 & 34.3 & 55.9 & 14.4 & 18.1 & 15.6 & 15.1 & 18.2 & 18.2 & 14.6          & \textbf{12.0} \\
& 0.8 & 18.0 & 35.3 & 4.5  & 6.8  & 4.6  & 4.6  & 6.2  & 5.9  & \textbf{4.0}  & \textbf{3.3}  \\
& 1   & 9.3  & 26.4 & 0.9  & 2.8  & 1.1  & 1.1  & 1.3  & 1.2  & \textbf{0.9}  & \textbf{0.8}  \\
\hline
\multirow{5}{*}{$l_{\infty}$-at}
 & 1 & 80.4 & 79.1 & 78.4 & 79.5 & 77.6 & 77.6 & 78.4 & 78.5 & \textbf{77.1} & \textbf{76.8} \\
 & 2 & 65.3 & 50.2 & 48.5 & 52.9 & 49.3 & 48.4 & 50.8 & 51.4 & \textbf{46.3} & \textbf{43.4} \\
 & 3 & 47.0 & 24.0 & 23.4 & 28.6 & 23.7 & 22.6 & 25.4 & 26.1 & \textbf{19.6} & \textbf{18.6} \\
 & 4 & 31.8 & 9.8  & 9.9  & 12.5 & 9.8  & 8.9  & 11.0 & 11.7 & \textbf{7.6}  & \textbf{6.2}  \\
 & 5 & 19.6 & 3.5  & 3.8  & 5.6  & 3.5  & 3.2  & 3.9  & 4.4  & \textbf{2.0}  & \textbf{1.9}  \\
\hline
\multirow{5}{*}{$l_2$-at}
& 2 & 74.7 & 72.3 & 72.2 & 74.4 & 72.0          & 72.0          & 73.0 & 73.0 & 72.2          & \textbf{71.7} \\
& 3 & 60.9 & 54.1 & 53.4 & 58.5 & 53.6          & 53.3          & 54.6 & 54.6 & 54.1          & \textbf{53.1} \\
& 4 & 46.4 & 36.1 & 33.3 & 39.3 & \textbf{33.0} & \textbf{33.0} & 36.5 & 36.2 & 34.9          & 34.2          \\
& 5 & 34.8 & 21.8 & 20.4 & 25.0 & 20.2          & 20.0          & 22.6 & 22.0 & \textbf{19.8} & \textbf{19.8} \\
& 6 & 24.7 & 11.1 & 10.6 & 14.5 & 10.4          & 10.2          & 11.9 & 11.8 & 10.3          & \textbf{9.7}  \\
\hline
\end{tabular}
}\\[-3.5ex]
\subfloat[][$l_1$-attack~\label{tab:imagenet_l1}]{
\setlength\tabcolsep{6.0pt} \begin{tabular}{@{\hspace{4pt}}C{10mm}@{\hspace{4pt}} | @{\hspace{3pt}} C{6mm} @{\hspace{3pt}} | @{\hspace{5pt}} *{9}{r}}
\multicolumn{11}{c}{}\\
\diagbox[width=13mm,height=5mm]{\tiny model}{\tiny $l_1$} & $\epsilon$ & SparseFool & EAD & B\&B & PGD-1 & PGD-10 & FAB-1 & FAB-10 & Our-1 & Our-10 \\
\hline
\multirow{5}{*}{plain}
& 5  & 89.7 & 78.2 & 80.9 & 90.4 & 90.4 & 83.0 & 80.0 & 80.6         & \textbf{76.2} \\
& 16 & 80.6 & 36.7 & 39.8 & 65.0 & 64.5 & 53.8 & 45.5 & 37.9         & \textbf{30.3} \\
& 27 & 70.6 & 13.2 & 17.9 & 30.6 & 29.7 & 34.4 & 25.6 & 14.2         & \textbf{9.0}  \\
& 38 & 61.6 & 5.2  & 5.9  & 12.2 & 10.9 & 22.2 & 13.7 & \textbf{4.5} & \textbf{2.2}  \\
& 49 & 52.9 & 2.2  & 2.1  & 4.8  & 3.7  & 15.4 & 7.7  & \textbf{1.5} & \textbf{0.9}  \\
\hline
\multirow{5}{*}{$l_{\infty}$-at}
& 15  & 79.8 & 66.6          & 71.4 & 90.9 & 88.7 & 72.8 & 69.6 & 67.2 & \textbf{63.6} \\
& 25  & 74.1 & \textbf{50.4} & 59.8 & 86.7 & 86.6 & 61.0 & 56.4 & 54.3 & 50.5          \\
& 40  & 64.5 & \textbf{34.0} & 44.1 & 82.6 & 82.6 & 48.6 & 42.8 & 39.8 & 35.8          \\
& 60  & 52.6 & \textbf{19.8} & 30.1 & 74.5 & 74.3 & 36.0 & 29.4 & 28.0 & 22.9          \\
& 100 & 39.2 & \textbf{7.4}  & 14.2 & 46.5 & 45.3 & 19.5 & 13.6 & 13.2 & 10.1          \\
\hline
\multirow{5}{*}{$l_2$-at}
& 50  & 86.2 & 79.0          & 81.3 & 87.4 & 87.4 & 80.8 & 80.5 & 79.1 & \textbf{78.6} \\
& 100 & 78.1 & 59.4          & 65.4 & 79.2 & 79.1 & 64.6 & 62.2 & 60.1 & \textbf{59.2} \\
& 150 & 69.8 & \textbf{41.5} & 48.3 & 65.1 & 64.4 & 48.6 & 45.7 & 42.7 & 41.7          \\
& 200 & 60.7 & 28.2          & 35.7 & 47.7 & 46.6 & 34.8 & 31.3 & 29.1 & \textbf{28.1} \\
& 250 & 52.2 & \textbf{19.0} & 25.0 & 30.4 & 29.1 & 26.6 & 22.8 & 20.6 & \textbf{19.0} \\
\hline
\end{tabular}
}\\[-3.5ex]
\subfloat[][$l_0$-attack~\label{tab:imagenet_l0}]{
\setlength\tabcolsep{6pt} \begin{tabular}{@{\hspace{4pt}}C{10mm}@{\hspace{4pt}} | @{\hspace{3pt}} C{6mm} @{\hspace{3pt}} | @{\hspace{5pt}} *{7}{r}}
\multicolumn{9}{c}{}\\
\diagbox[width=13mm,height=5mm]{\tiny model}{\tiny $l_{\infty}$} & $\epsilon$ & SparseFool & JSMA & B\&B & PGD-1 & PGD-10 & Our-1 & Our-10 \\
\hline
\multirow{5}{*}{plain}
& 10 & 83.0 & 91.4 & 79.7 & 88.0 & 83.4 & \textbf{69.7} & \textbf{62.9} \\
& 20 & 71.3 & 88.3 & 58.5 & 78.2 & 69.7 & \textbf{43.4} & \textbf{34.4} \\
& 30 & 57.9 & 85.0 & 40.8 & 66.6 & 53.8 & \textbf{25.2} & \textbf{17.8} \\
& 40 & 46.1 & 82.1 & 27.3 & 56.5 & 41.6 & \textbf{12.9} & \textbf{8.5}  \\
& 50 & 37.9 & 78.8 & 16.3 & 40.4 & 25.4 & \textbf{6.9}  & \textbf{4.6}  \\
\hline
\multirow{5}{*}{$l_{\infty}$-at}
& 10  & 78.9 & 88.0 & 79.5 & 85.0 & 81.0 & \textbf{70.2} & \textbf{65.4} \\
& 30  & 64.2 & 83.0 & 57.0 & 68.9 & 59.4 & \textbf{45.4} & \textbf{37.3} \\
& 50  & 51.4 & 78.8 & 40.7 & 51.2 & 40.6 & \textbf{28.8} & \textbf{21.6} \\
& 80  & 34.5 & 71.6 & 23.4 & 34.1 & 23.8 & \textbf{18.7} & \textbf{12.7} \\
& 100 & 26.9 & 68.1 & 17.7 & 25.8 & 16.2 & \textbf{14.3} & \textbf{10.3} \\
\hline
\multirow{5}{*}{$l_2$-at}
& 50  & 75.9 & 87.8 & 80.7 & 78.9 & 75.2 & \textbf{69.4} & \textbf{66.2} \\
& 100 & 52.6 & 83.5 & 65.6 & 62.0 & 52.6 & \textbf{42.4} & \textbf{39.7} \\
& 150 & 30.9 & 76.8 & 46.9 & 42.8 & 35.4 & \textbf{26.7} & \textbf{24.8} \\
& 200 & 16.5 & 69.6 & 35.1 & 32.3 & 24.4 & \textbf{16.5} & \textbf{15.8} \\
& 250 & 9.7  & 64.5 & 25.0 & 23.6 & 15.9 & 10.5          & \textbf{9.3} \\
\hline
\end{tabular}
}
\hfill
\end{table*}
\addtocounter{table}{1}
\twocolumn

\section{Analysis of the attacks}
\vspace*{-1pt}

In this section, we perform additional analysis of the proposed attack compared to PGD and FAB attacks. We illustrate how the attack's average robust norm and the attack's average robust accuracy changes as we increase the number of random restarts and the number of gradient queries, respectively.

\begin{figure*}[!b]
  \vspace*{-11pt}
  \setlength\tabcolsep{0pt} \renewcommand*{\arraystretch}{1.0}
  \caption{Evolution of the average robust norm for FAB and our attack on MNIST as we increase the number of restarts.~\label{fig:fab-vs-our-mnist}}
  \vspace{6pt}
  \begin{tabular}{C{172pt} C{172pt} C{172pt}}
      \scriptsize \cellcolor{gray!20}plain                                                              & \scriptsize \cellcolor{gray!20}$l_{\infty}$-AT                                                   & \scriptsize \cellcolor{gray!20}$l_2$-AT \\
      \resizebox{174pt}{122pt}{\includegraphics[width=\textwidth]{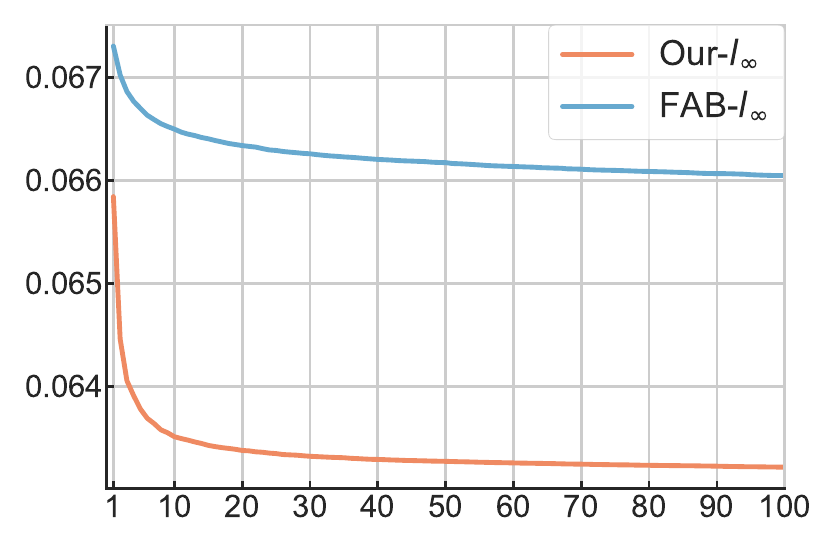}} & \resizebox{174pt}{122pt}{\includegraphics[width=\textwidth]{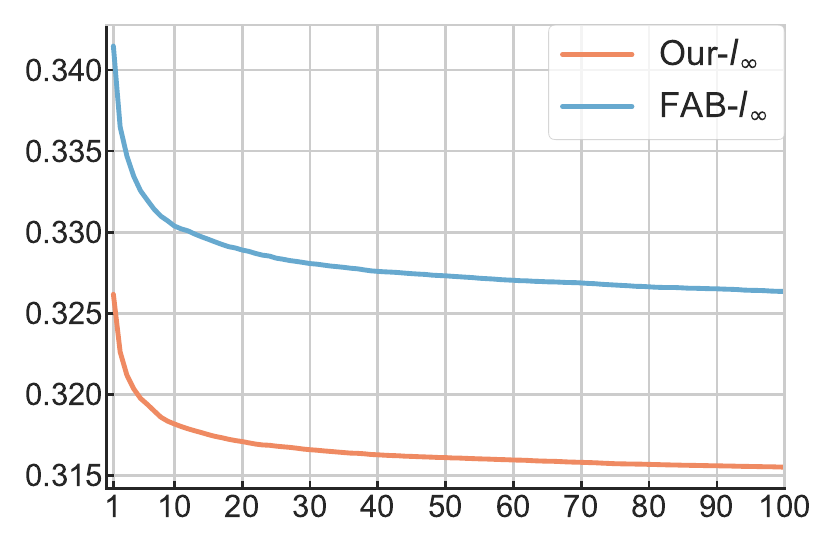}} & \resizebox{174pt}{122pt}{\includegraphics[width=\textwidth]{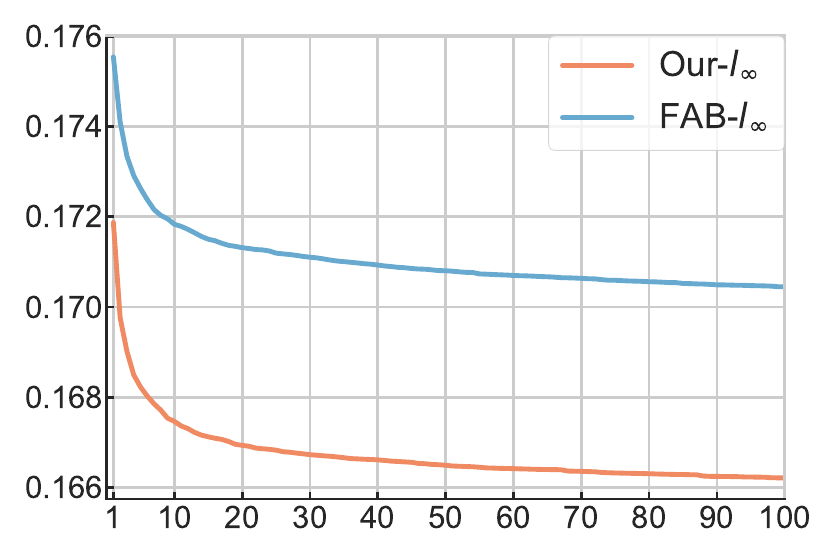}} \\[-1ex]
      \resizebox{174pt}{122pt}{\includegraphics[width=\textwidth]{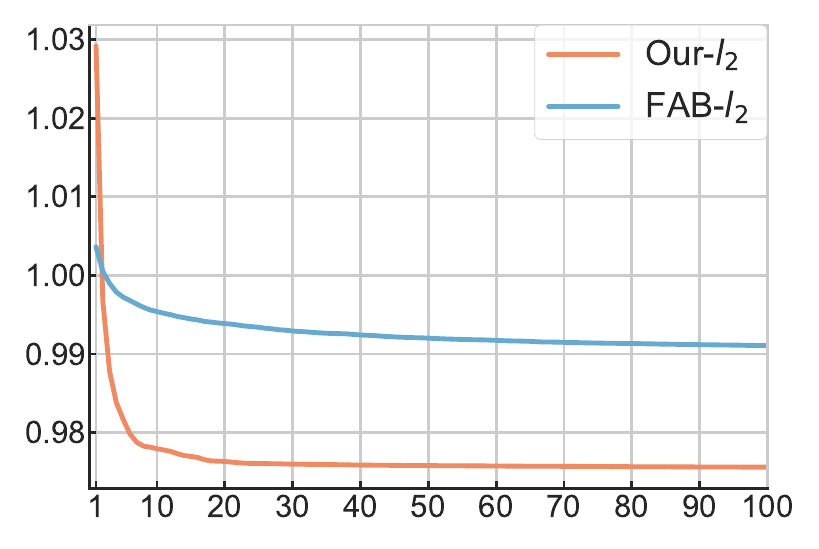}} & \resizebox{174pt}{122pt}{\includegraphics[width=\textwidth]{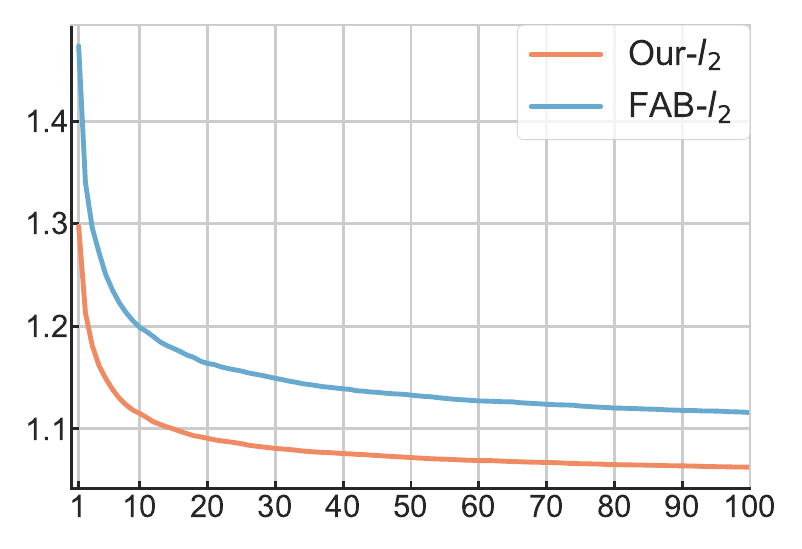}} & \resizebox{174pt}{122pt}{\includegraphics[width=\textwidth]{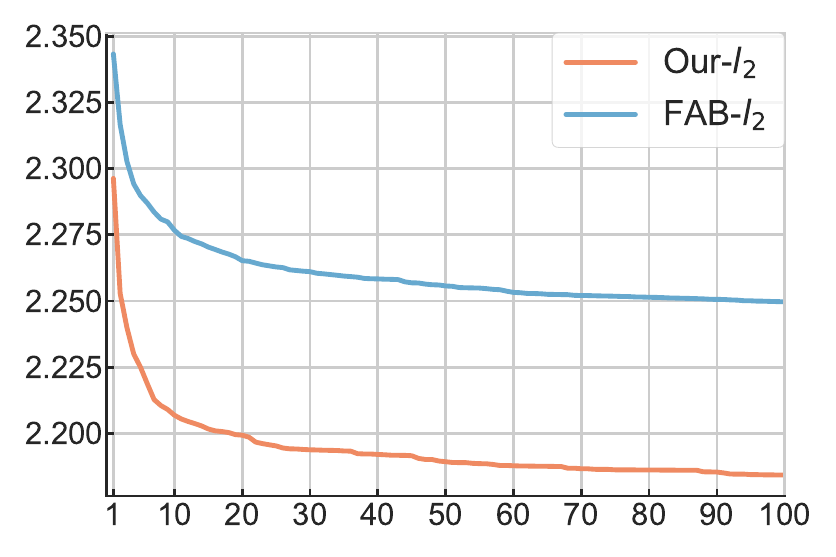}} \\[-1ex]
      \resizebox{174pt}{122pt}{\includegraphics[width=\textwidth]{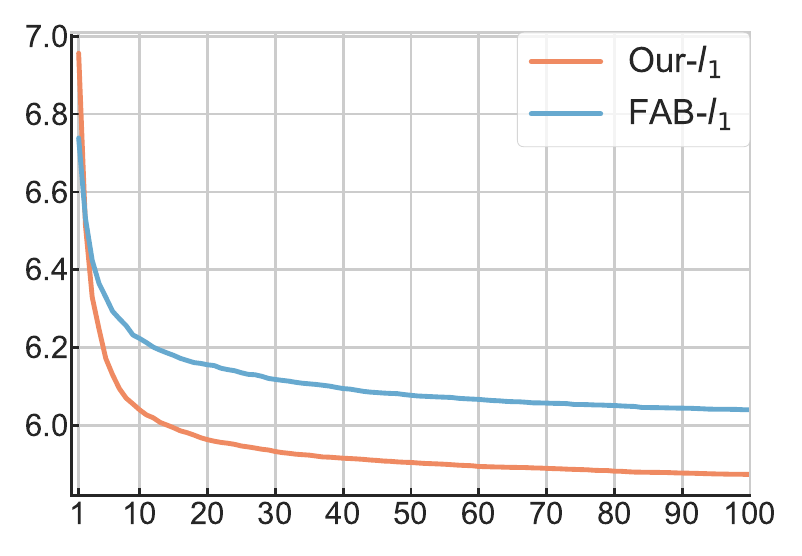}} & \resizebox{174pt}{122pt}{\includegraphics[width=\textwidth]{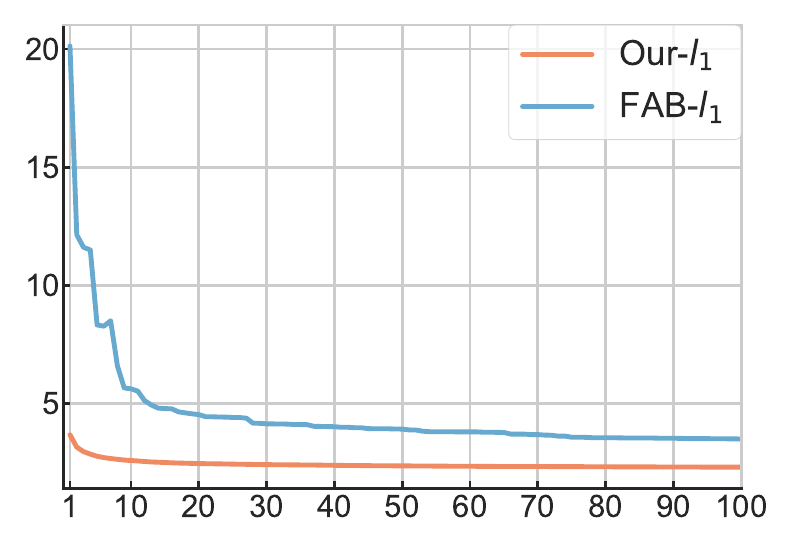}} & \resizebox{174pt}{122pt}{\includegraphics[width=\textwidth]{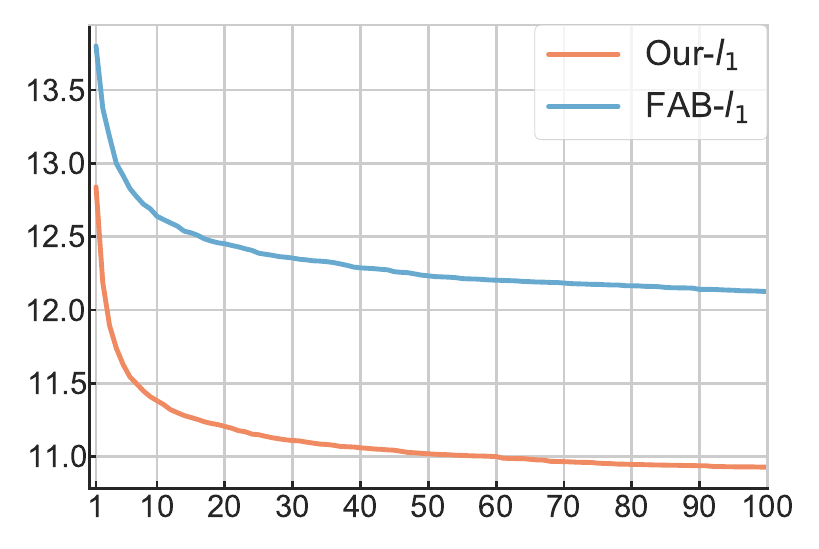}} \\[-1ex]
    \end{tabular}
\end{figure*}

\begin{figure*}
  \setlength\tabcolsep{0pt} \renewcommand*{\arraystretch}{1.0}
  \caption{Evolution of the average robust norm for FAB and our attack on CIFAR-10 as we increase the number of restarts.~\label{fig:fab-vs-our-cifar10}}
  \vspace{6pt}
  \begin{tabular}{C{172pt} C{172pt} C{172pt}}
    \scriptsize \cellcolor{gray!20}plain                                                              & \scriptsize \cellcolor{gray!20}$l_{\infty}$-AT                                                   & \scriptsize \cellcolor{gray!20}$l_2$-AT \\
    \resizebox{174pt}{122pt}{\includegraphics[width=\textwidth]{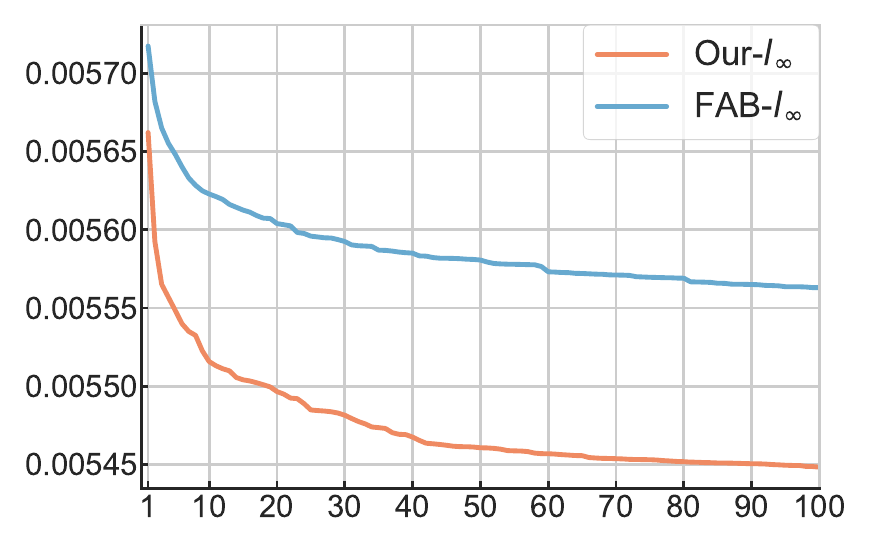}} & \resizebox{174pt}{122pt}{\includegraphics[width=\textwidth]{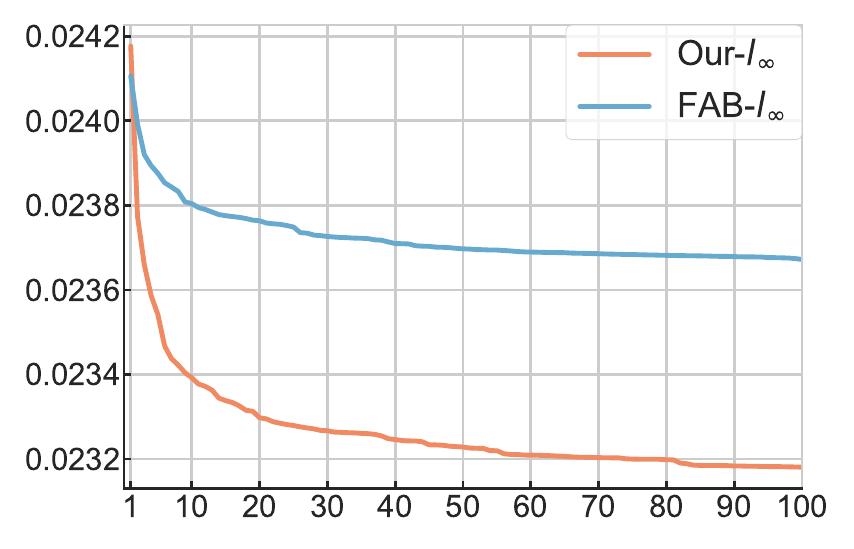}} & \resizebox{174pt}{122pt}{\includegraphics[width=\textwidth]{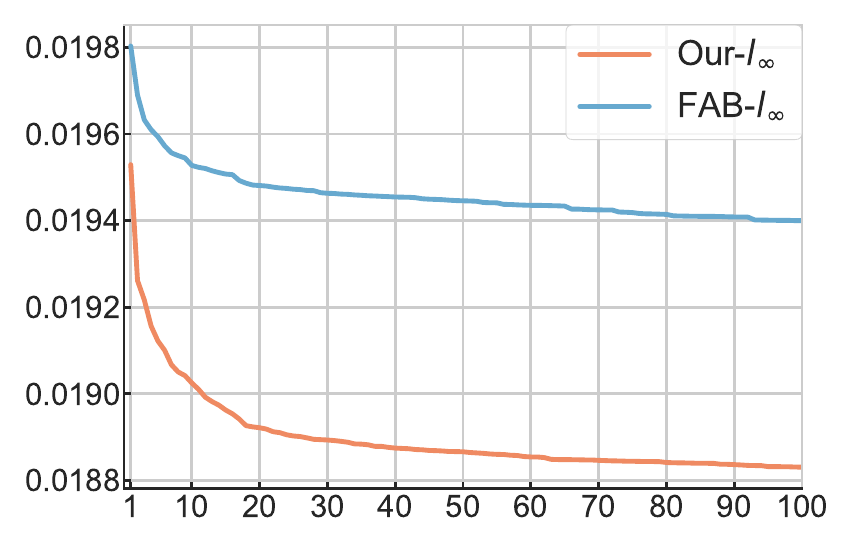}} \\[-1ex]
    \resizebox{174pt}{122pt}{\includegraphics[width=\textwidth]{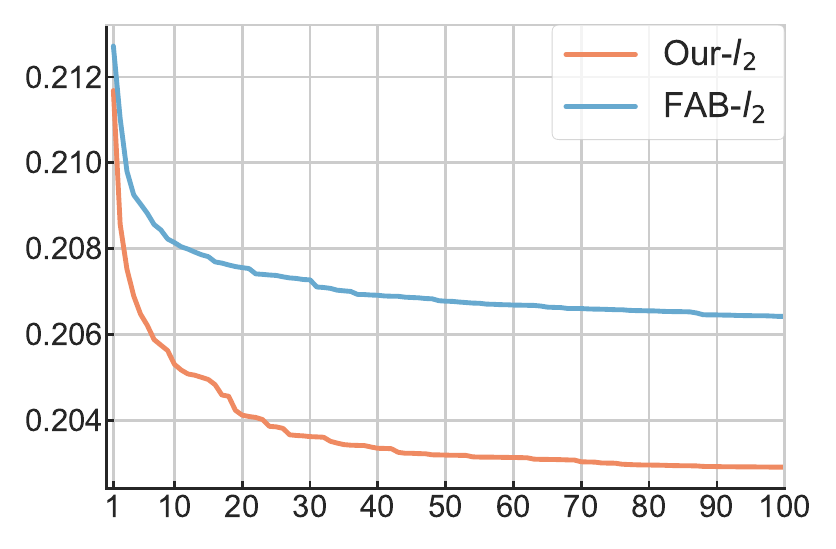}} & \resizebox{174pt}{122pt}{\includegraphics[width=\textwidth]{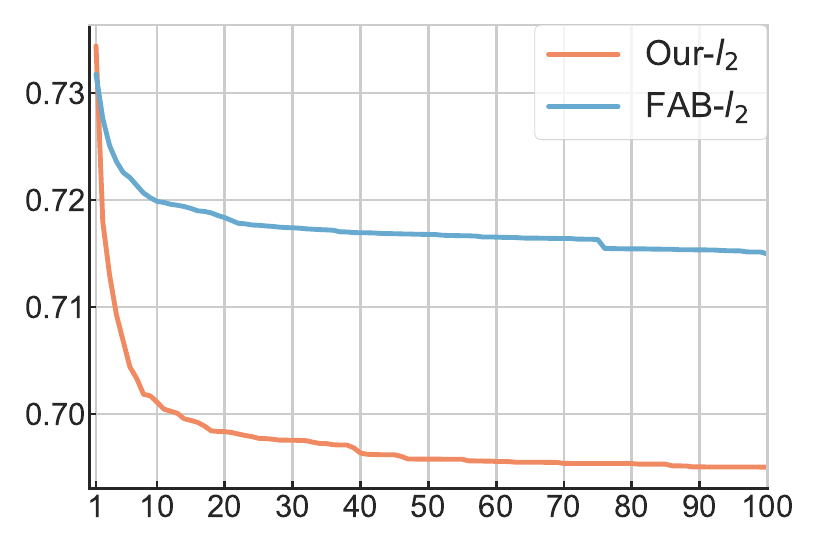}} & \resizebox{174pt}{122pt}{\includegraphics[width=\textwidth]{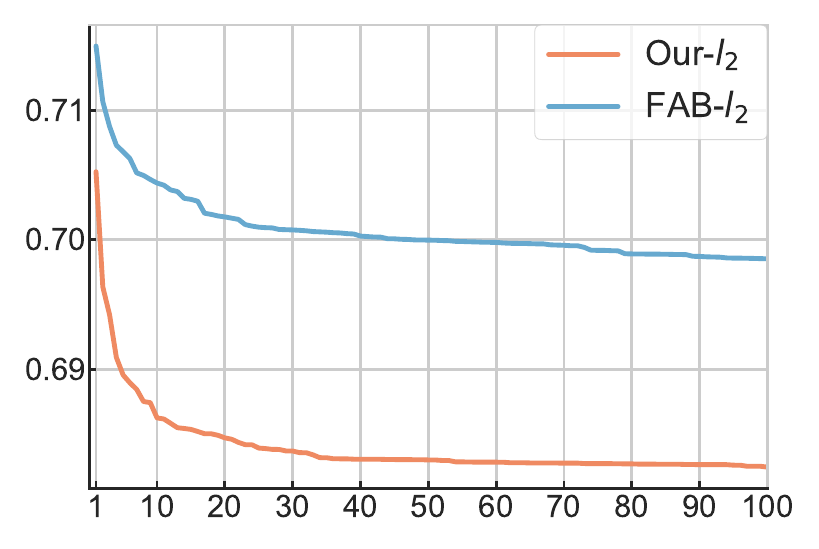}} \\[-1ex]
    \resizebox{174pt}{122pt}{\includegraphics[width=\textwidth]{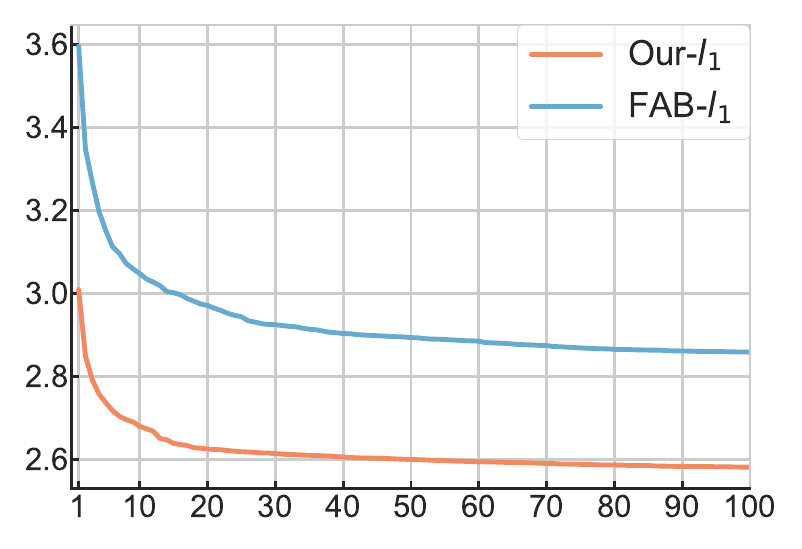}} & \resizebox{174pt}{122pt}{\includegraphics[width=\textwidth]{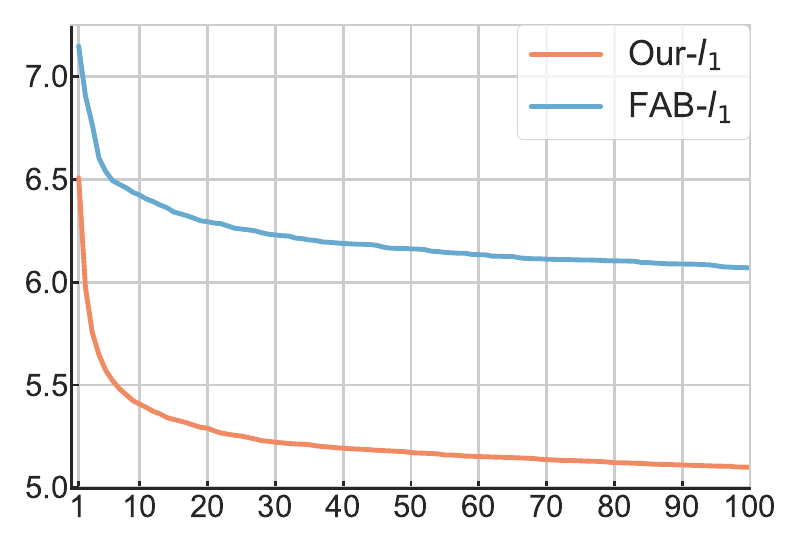}} & \resizebox{174pt}{122pt}{\includegraphics[width=\textwidth]{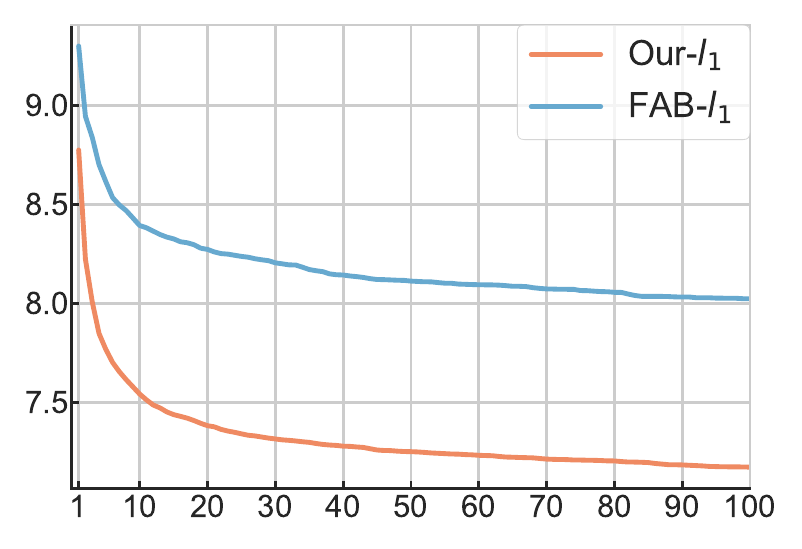}} \\[-1ex]
  \end{tabular}
\end{figure*}

First, we compare our attack with FAB attack, which has computational complexity similar to our attack~(see~Section VI.A of the main paper for details). We show the evolution of the average robust norm for our and FAB attacks in~\Cref{fig:fab-vs-our-mnist,fig:fab-vs-our-cifar10}. As we can see in~\cref{fig:fab-vs-our-mnist,fig:fab-vs-our-cifar10}, our attack with 1 random restart outperforms FAB attack with 1 random restart in 14 out 18 cases. FAB attack requires the computation of $k$ gradients in order to find the optimal target. The adversarial target for our attack depends on the initial random initialisation. Nonetheless, our $l_{\infty}$-norm attack with 1 random restart against naturally trained and $l_{\infty}$-AT models outperforms FAB attack with 100 random restarts on MNIST dataset. Our attack with 10 random restarts beats FAB attack with 100 random restarts for all models and norms on MNIST and {CIFAR-10} datasets. Our attack on average reduces the robust $l_{\infty}$-norm by 3.5\%/2.5\%, $l_2$-norm by 3.2\%/2.5\% and $l_1$-norm by 21.4\%/14.1\% on MNIST / CIFAR-10 datasets. The improvement over FAB is the most significant for $l_1$-norm.

\begin{figure*}
  \setlength\tabcolsep{0pt} \renewcommand*{\arraystretch}{1.0}
  \caption{Evolution of the average robust accuracy for PGD and our attack with $C=0.1$, $C=1$, $C=10$ on MNIST as we increase the number of gradient queries for each example (best viewed on-screen). We disable random initialisation for both attacks. We run 5 PGD attacks sequentially as we increase $\epsilon$. For PGD, we exploit the fact that the inputs non-robust at a threshold $\epsilon$ are non-robust for thresholds larger than $\epsilon$.~\label{fig:pgd-vs-our-mnist}}
  \vspace{6pt}
  \begin{tabular}{C{172pt} C{172pt} C{172pt}}
    \scriptsize \cellcolor{gray!20}plain                                                              & \scriptsize \cellcolor{gray!20}$l_{\infty}$-AT                                                   & \scriptsize \cellcolor{gray!20}$l_2$-AT \\
    \resizebox{180pt}{126pt}{\includegraphics[width=\textwidth]{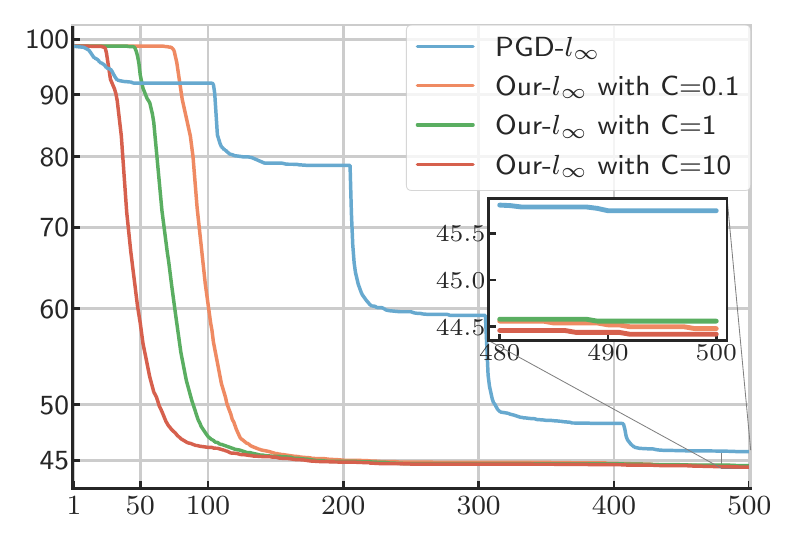}} & \resizebox{180pt}{126pt}{\includegraphics[width=\textwidth]{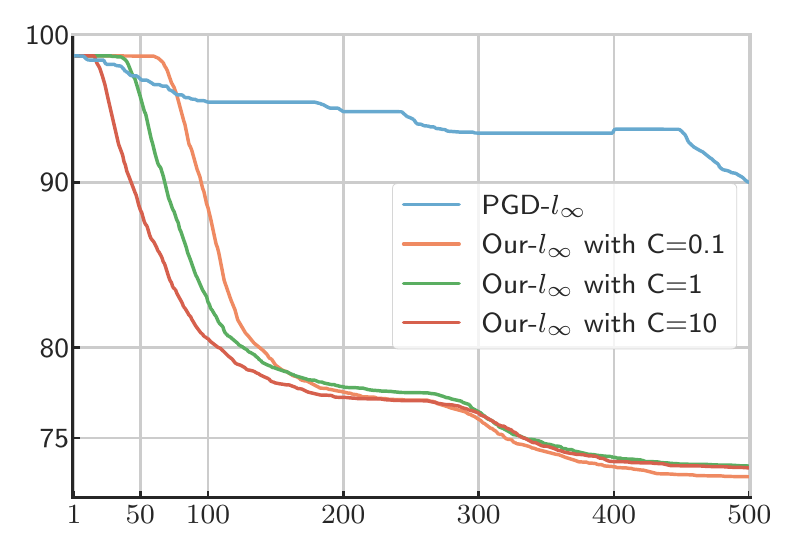}} & \resizebox{180pt}{126pt}{\includegraphics[width=\textwidth]{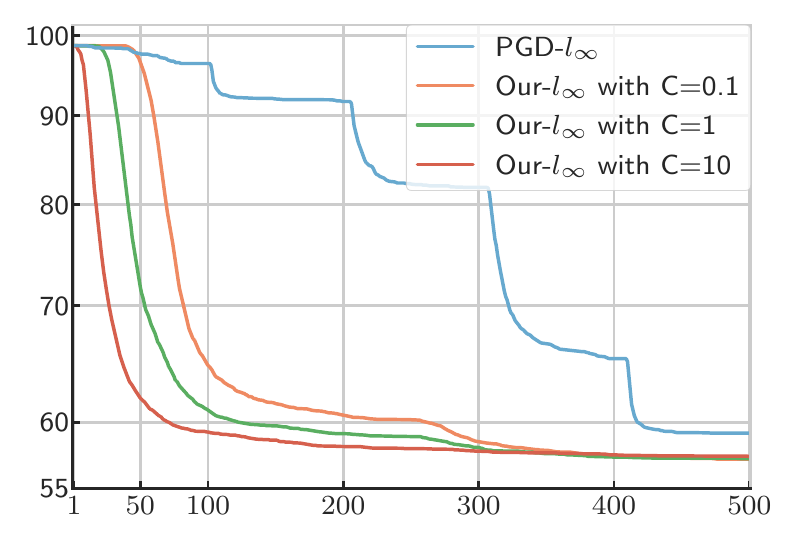}} \\[-1ex]
    \resizebox{180pt}{126pt}{\includegraphics[width=\textwidth]{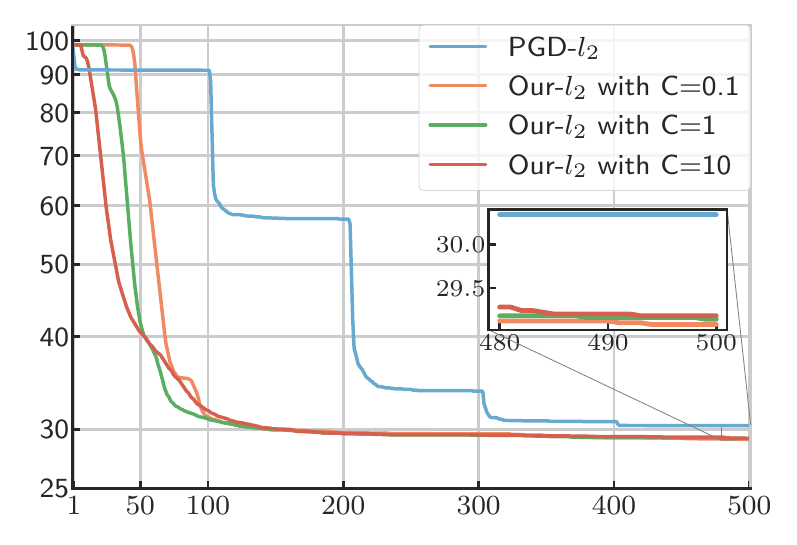}} & \resizebox{180pt}{126pt}{\includegraphics[width=\textwidth]{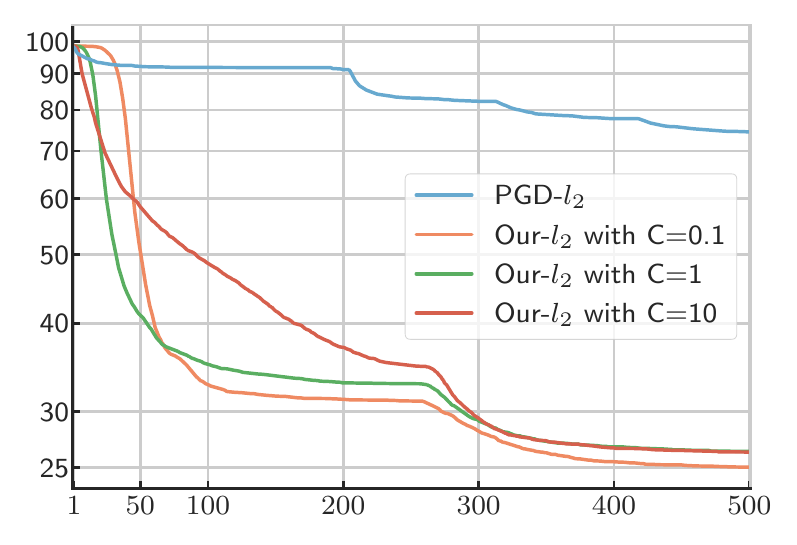}} & \resizebox{180pt}{126pt}{\includegraphics[width=\textwidth]{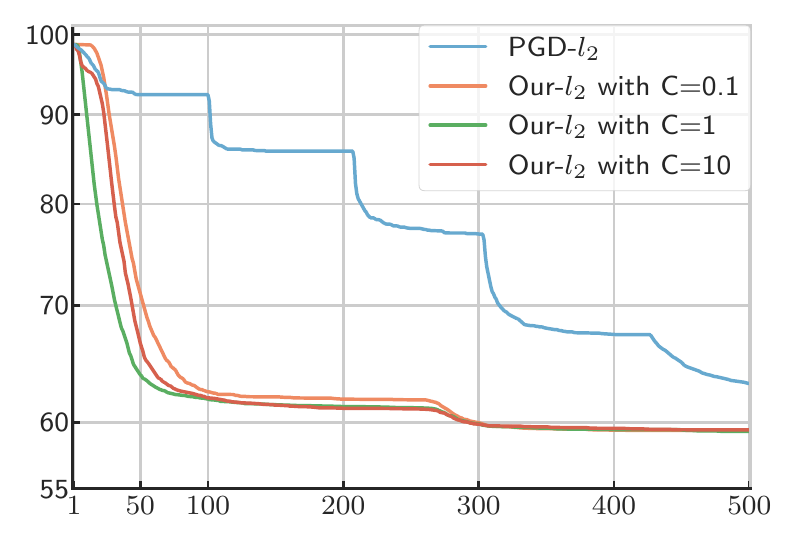}} \\[-1ex]
    \resizebox{180pt}{126pt}{\includegraphics[width=\textwidth]{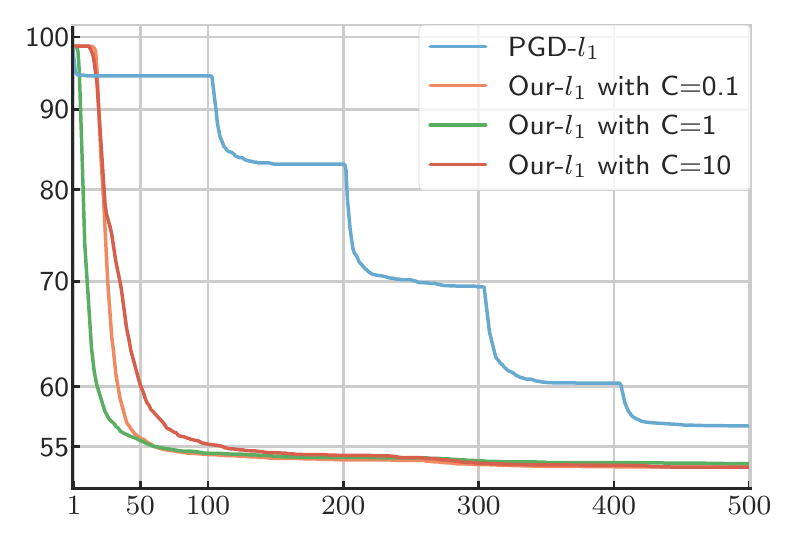}} & \resizebox{180pt}{126pt}{\includegraphics[width=\textwidth]{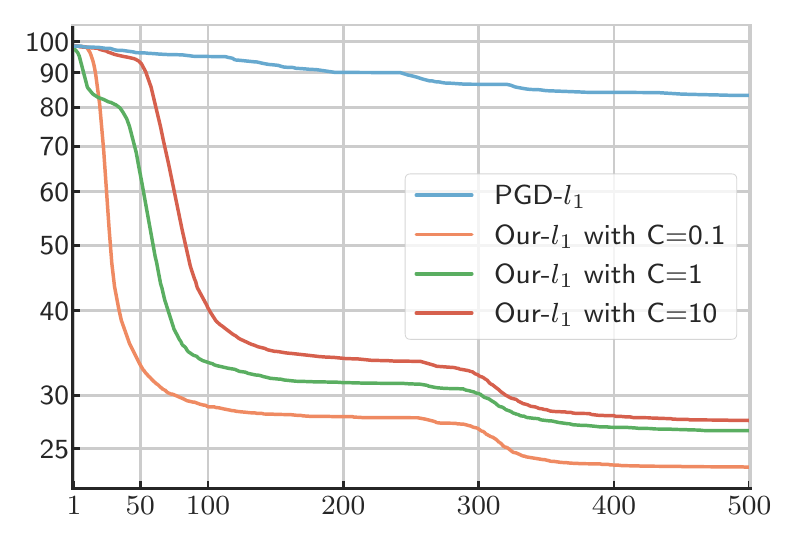}} & \resizebox{180pt}{126pt}{\includegraphics[width=\textwidth]{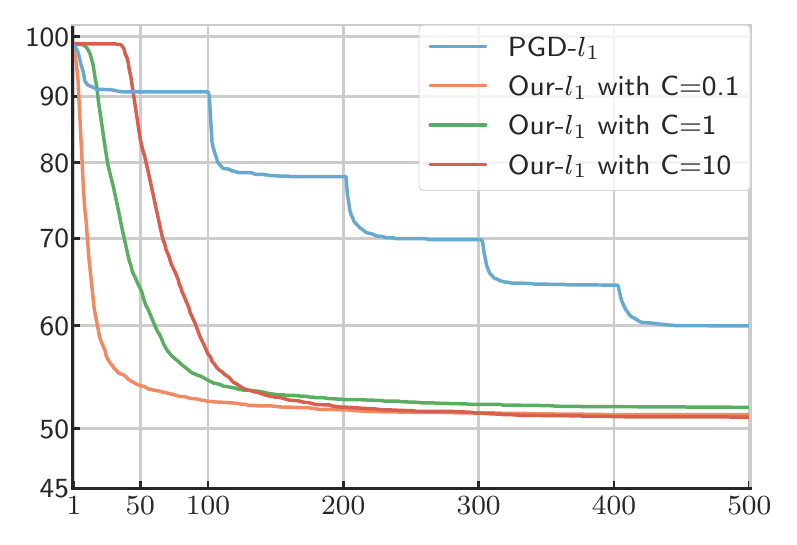}} \\[-1ex]
    \resizebox{180pt}{126pt}{\includegraphics[width=\textwidth]{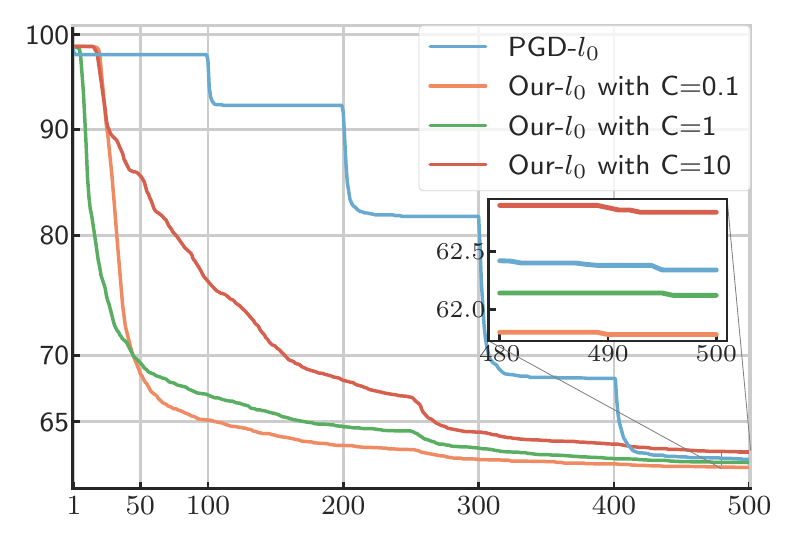}} & \resizebox{180pt}{126pt}{\includegraphics[width=\textwidth]{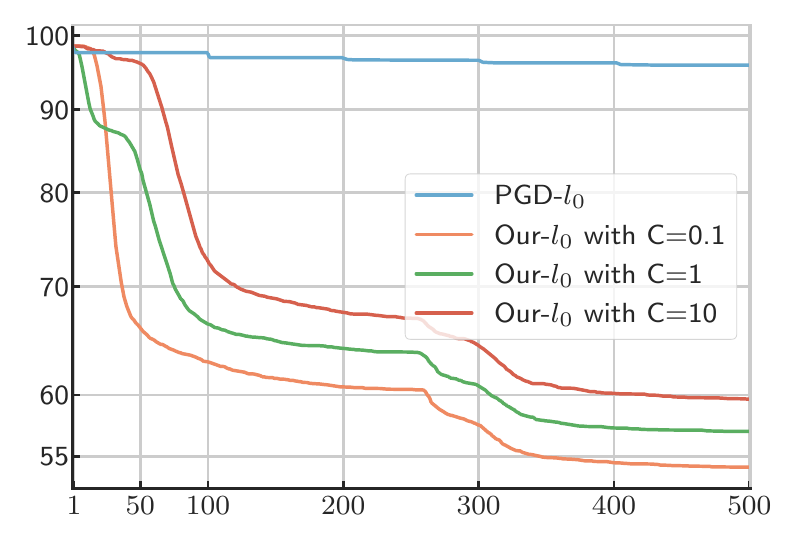}} & \resizebox{180pt}{126pt}{\includegraphics[width=\textwidth]{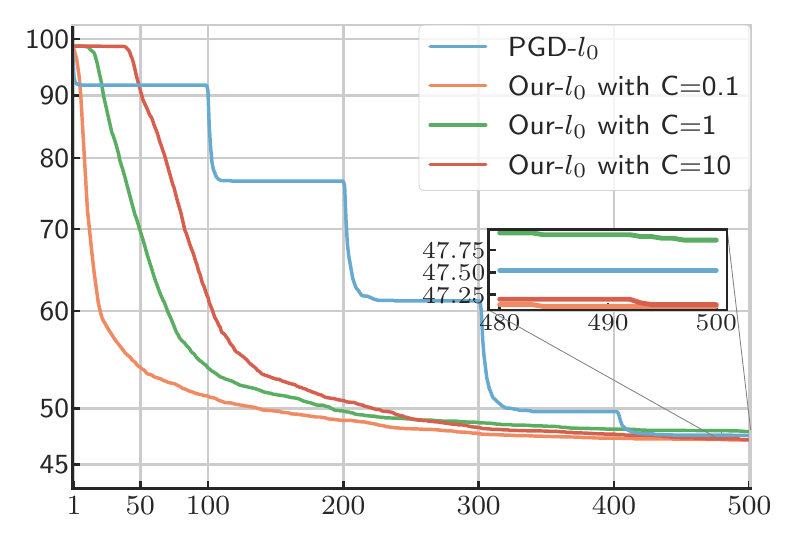}}
  \end{tabular}
\end{figure*}

\begin{figure*}[!h]
  \setlength\tabcolsep{0pt} \renewcommand*{\arraystretch}{1.0}
  \caption{Evolution of the average robust accuracy for PGD and our attack with $C=0.1$, $C=1$, $C=10$ on CIFAR-10 as we increase the number of gradient queries for each example (best viewed on-screen). We disable random initialisation for both attacks. We run 5 PGD attacks sequentially as we increase $\epsilon$. For PGD, we exploit the fact that the inputs non-robust at a threshold $\epsilon$ are non-robust for thresholds larger than $\epsilon$.~\label{fig:pgd-vs-our-cifar10}}
  \vspace{6pt}
  \begin{tabular}{C{172pt} C{172pt} C{172pt}}
    \scriptsize \cellcolor{gray!20}plain                                                              & \scriptsize \cellcolor{gray!20}$l_{\infty}$-AT                                                   & \scriptsize \cellcolor{gray!20}$l_2$-AT \\
    \resizebox{180pt}{126pt}{\includegraphics[width=\textwidth]{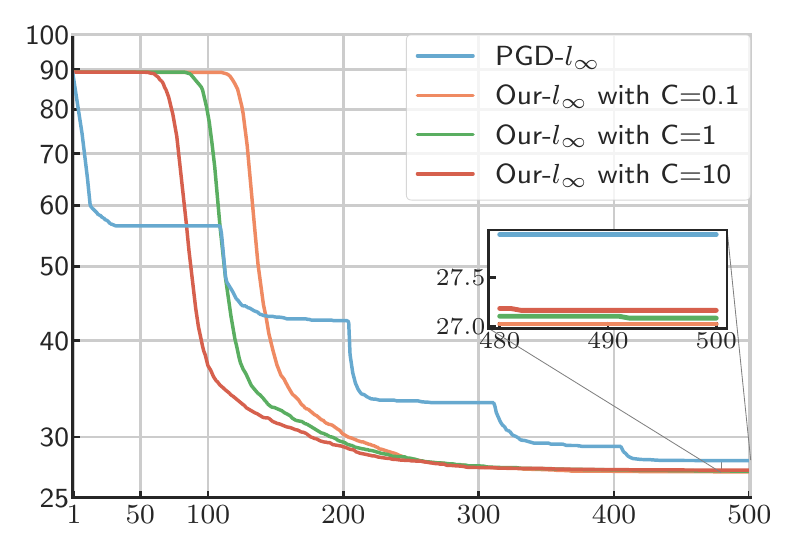}} & \resizebox{180pt}{126pt}{\includegraphics[width=\textwidth]{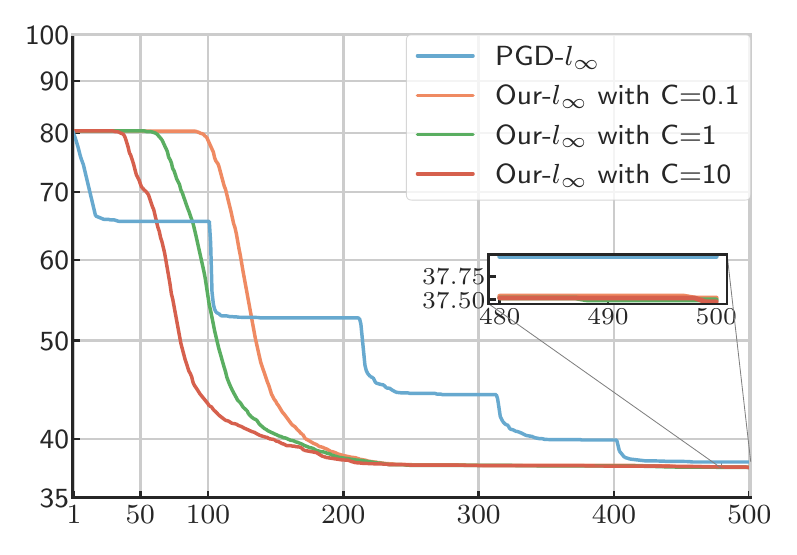}} & \resizebox{180pt}{126pt}{\includegraphics[width=\textwidth]{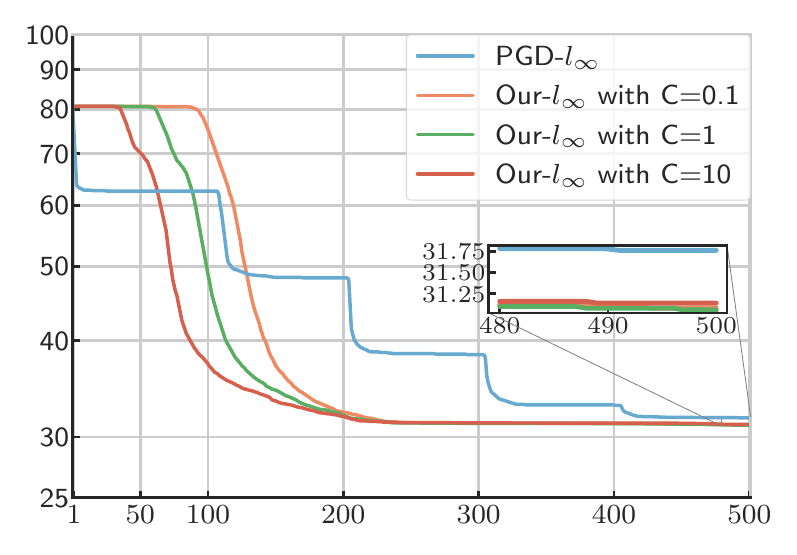}} \\[-1ex]
    \resizebox{180pt}{126pt}{\includegraphics[width=\textwidth]{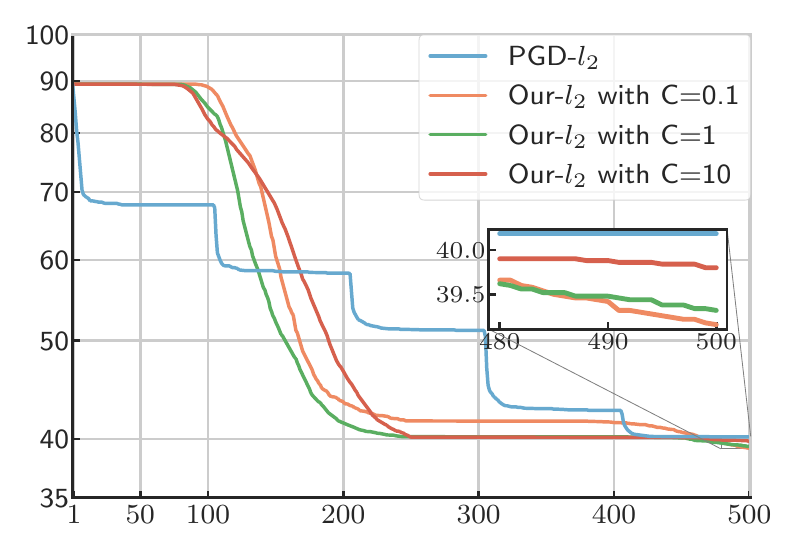}} & \resizebox{180pt}{126pt}{\includegraphics[width=\textwidth]{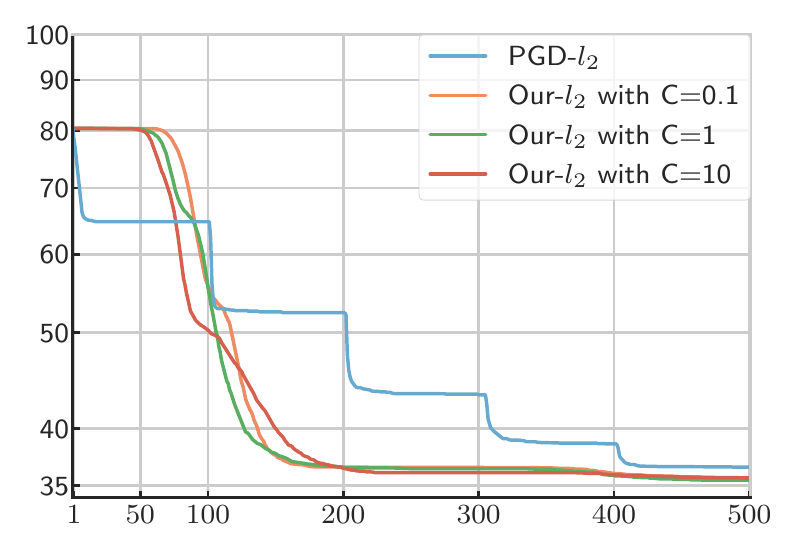}} & \resizebox{180pt}{126pt}{\includegraphics[width=\textwidth]{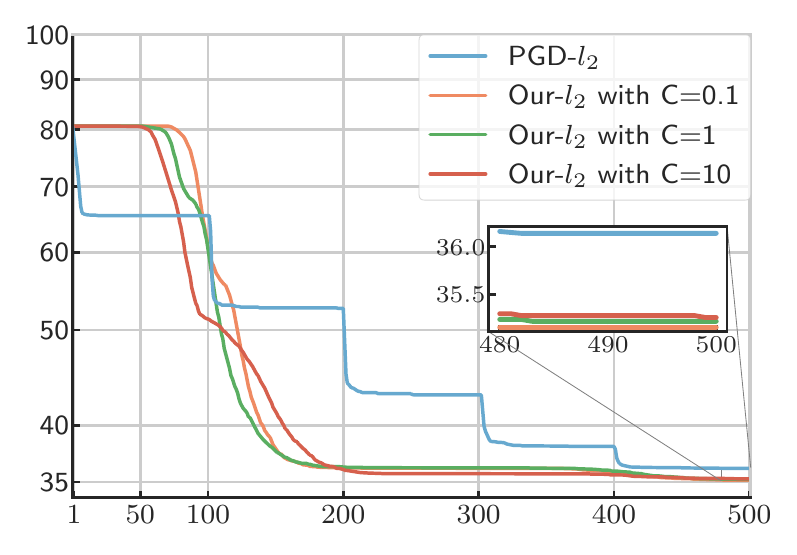}} \\[-1ex]
    \resizebox{180pt}{126pt}{\includegraphics[width=\textwidth]{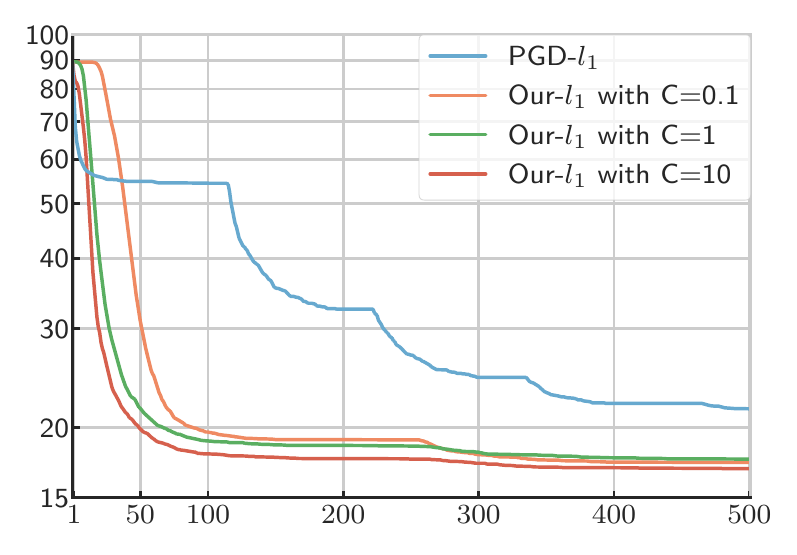}} & \resizebox{180pt}{126pt}{\includegraphics[width=\textwidth]{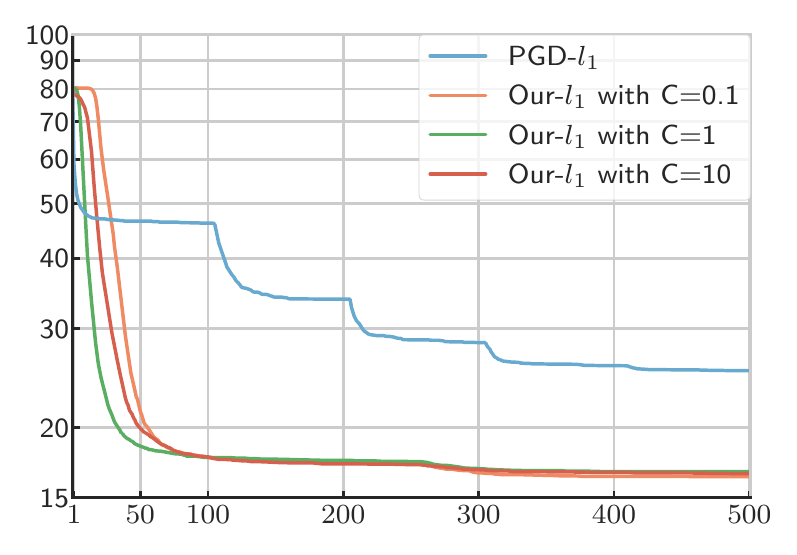}} & \resizebox{180pt}{126pt}{\includegraphics[width=\textwidth]{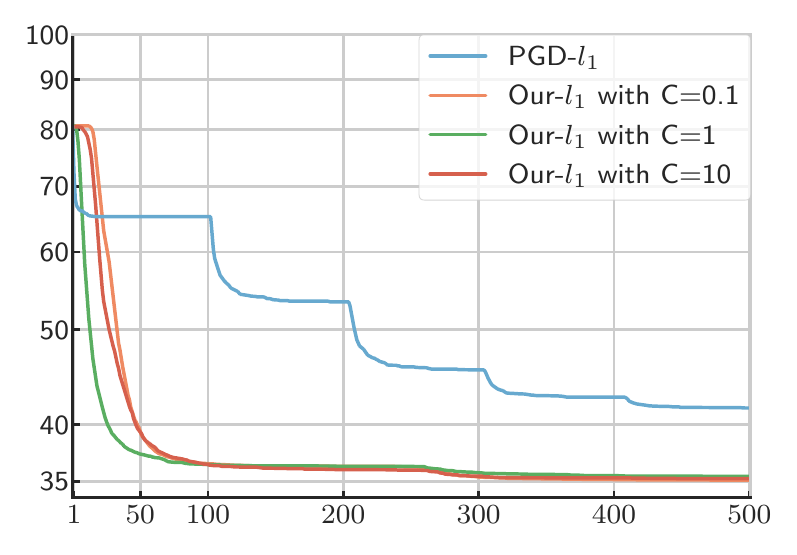}} \\[-1ex]
    \resizebox{180pt}{126pt}{\includegraphics[width=\textwidth]{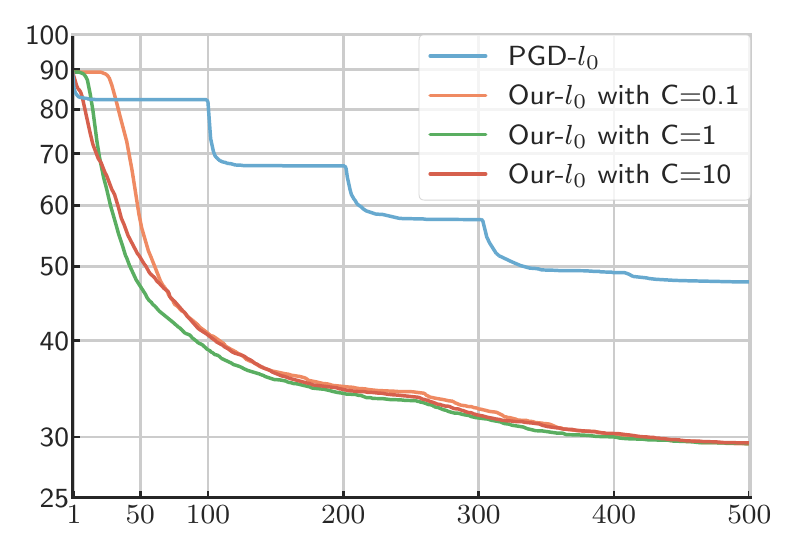}} & \resizebox{180pt}{126pt}{\includegraphics[width=\textwidth]{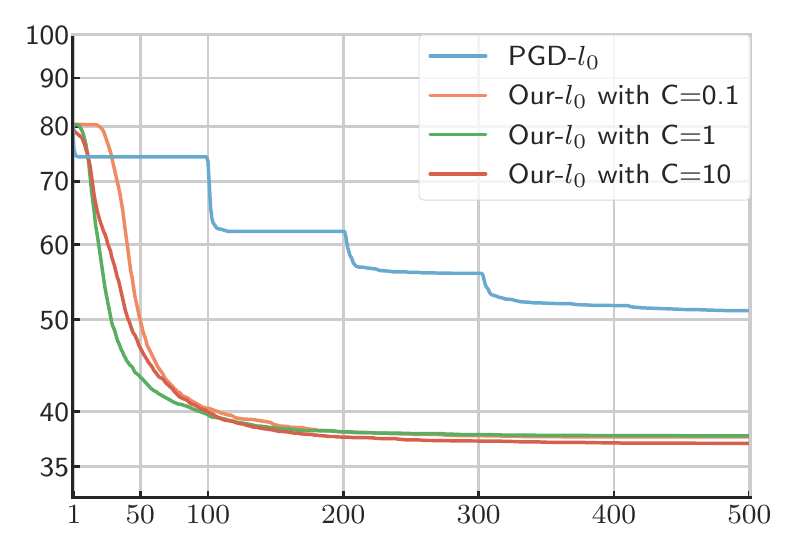}} & \resizebox{180pt}{126pt}{\includegraphics[width=\textwidth]{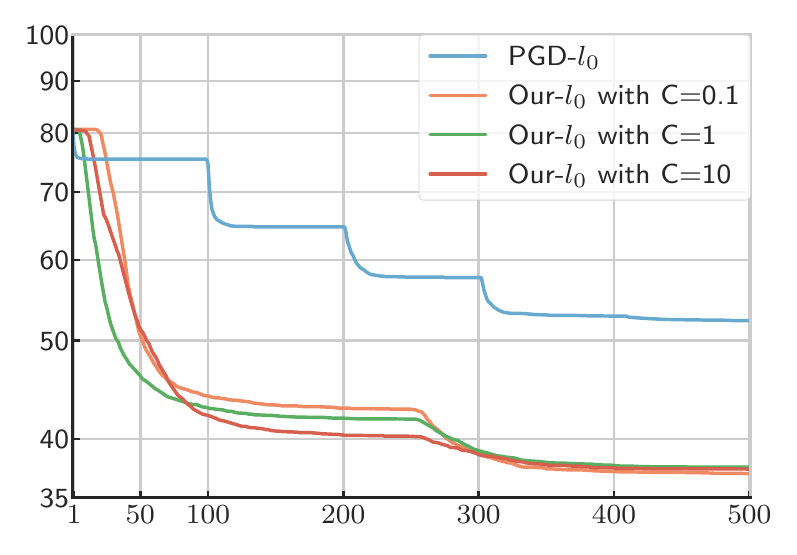}}
  \end{tabular}
\end{figure*}

In the next experiment, we compare PGD and our attack without restarts as we increase the number of model's gradient queries in~\Cref{fig:pgd-vs-our-mnist,fig:pgd-vs-our-cifar10} on MNIST and CIFAR-10 datasets, respectively. PGD attack allows to quickly estimate the robust accuracy at the specific threshold. To estimate the robust accuracy at 5 thresholds, we sequentially run 5 PGD attacks with an equal computational budget as we increase $\epsilon$, so the total number of gradient queries on each clean image is the same for both attacks. We disable random initialisation for both attacks to reduce the variation due to the random starting point. For PGD, we also exploit the fact that the inputs non-robust at a threshold $\epsilon$ are non-robust for thresholds larger than $\epsilon$. As we can see in~\cref{fig:pgd-vs-our-mnist,fig:pgd-vs-our-cifar10}, our attack outperforms PGD attack for all norms and models on MNIST and {CIFAR-10} datasets. Our attack has a ``slow start` because it optimises primal and dual variables simultaneously. We can improve the convergence speed of our attack by changing the initial value $C$ of the dual variable. Overall, our attack always outperforms PGD attack on MNIST and CIFAR-10 datasets given a sufficient computational budget.

\end{document}